\documentclass[11pt]{article}

\usepackage[final]{acl}

\usepackage{times}
\usepackage{latexsym}

\usepackage[T1]{fontenc}

\usepackage[utf8]{inputenc}

\usepackage{microtype}

\usepackage{inconsolata}

\usepackage{graphicx}

\usepackage{xcolor}
\usepackage{tcolorbox}
\usepackage{booktabs}
\usepackage{subcaption}
\usepackage[textsize=tiny]{todonotes}
\usepackage{amsfonts}
\usepackage{amsmath}
\usepackage{xspace}

\usepackage{CJKutf8}
\usepackage{newunicodechar}
\usepackage{amssymb}
\usepackage{fancyvrb}
\newunicodechar{✓}{\checkmark}
\newunicodechar{✗}{$\times$}
\newunicodechar{→}{$\rightarrow$}
\newunicodechar{é}{\'e}
\newunicodechar{à}{\`a}
\definecolor{red}{HTML}{C1121F}
\definecolor{blue}{HTML}{1D4ED8}
\definecolor{plotly}{HTML}{344967}
\newcommand{\red}[1]{\textcolor{red}{#1}}
\newcommand{\blue}[1]{\textcolor{blue}{#1}}

\newcommand{\wlt}{\textsc{WLT}\xspace}
\usepackage[capitalise]{cleveref}
\usepackage{calc}  %

\title{Copy First, Translate Later: Interpreting Translation Dynamics\\in Multilingual Pretraining}

\author{
 \textbf{Felicia Körner\textsuperscript{*,1,2}},
 \textbf{Maria Matveev\textsuperscript{*,2,3}},
 \textbf{Florian Eichin\textsuperscript{*,1,2}},
\\
 \textbf{Gitta Kutyniok\textsuperscript{$\dagger$,2,3,4,5}},
 \textbf{Barbara Plank\textsuperscript{$\dagger$,1,2}},
 \textbf{Michael A. Hedderich\textsuperscript{$\dagger$,1,2}}
\\
 \textsuperscript{1}MaiNLP, Center for Information and Language Processing, LMU Munich, Germany,\\
 \textsuperscript{2}Munich Center for Machine Learning (MCML),\\
 \textsuperscript{3}Department of Mathematics, LMU Munich, Germany,\\
  \textsuperscript{4}Department of Physics and Technology, University of Tromsø, Norway,\\
 \textsuperscript{5}DLR-German Aerospace Center, Germany\\
 \small{
 \textsuperscript{*}Equal contribution. \textsuperscript{$\dagger$}Shared senior authorship.
   \textbf{Correspondence:} \href{mailto:f.koerner@lmu.de}{f.koerner@lmu.de}
 }
}

\begin{document}
\maketitle
\begin{abstract}
Large language models exhibit impressive cross-lingual capabilities. However, \emph{how} cross-lingual generalization emerges has only been analyzed through isolated factors and at sparse points during training, limiting our understanding\textemdash particularly of the early phases of learning. To study the early trajectory of linguistic and translation capabilities, we pretrain a multilingual 1.7B model on nine diverse languages, capturing checkpoints at a much finer granularity. We use word-level translation as a testbed, introducing a novel dataset to trace how translation develops over training through behavioral analyses, model-component analysis, and parameter-based ablations. We find that the model quickly acquires basic linguistic capabilities in parallel with token-level copying, while translation develops in two distinct phases: an initial phase dominated by copying and surface-level similarities, and a second phase in which generalizing translation mechanisms develop and copying is refined. Together, these findings provide a fine-grained view of how cross-lingual generalization develops during multilingual pretraining.\!\footnote{To facilitate future work, we release our novel word-level translation dataset, checkpoints, and
code (\url{https://github.com/mainlp/copy-first-translate-later}).}
\end{abstract}

\section{Introduction}
Cross-lingual transfer is a key property of multilingual large language models, allowing high-resource languages to benefit lower-resourced ones \cite{wu-dredze-2019-beto,patil-etal-2022-overlap}. Surprisingly, cross-lingual transfer arises even without explicit cross-lingual supervision, such as parallel training data or cross-lingual training objectives \cite{blevins-zettlemoyer-2022-language}.
Prior work has identified contributing factors for this phenomenon, such as vocabulary overlap \cite{kallini-etal-2025-false}, incidental parallel data \cite{briakou-etal-2023-searching}, and shared intermediate representation spaces \cite{wendler-etal-2024-llamas}. However, how the \emph{mechanisms} behind cross-lingual transfer emerge is poorly understood.
\begin{figure}[t]
    \centering
    \includegraphics[width=\linewidth]{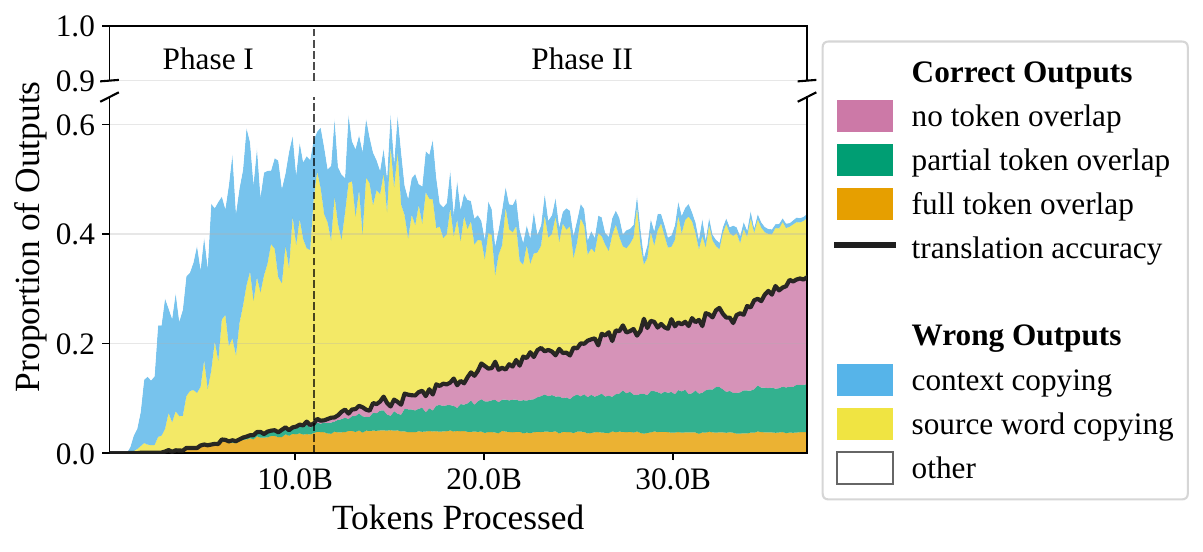}
    \caption{\textbf{Training dynamics of word-level translation over pretraining.} The stacked areas decompose outputs into correct translations and error types; the black line tracks overall translation accuracy. Early training (Phase I) is characterized by frequent copying, whereas translation develops in later training (Phase~II).
    }
    \label{fig:copying_error_mode}
\end{figure}

We argue that studying the final model alone cannot explain why mechanisms underlying cross-lingual transfer arise and how they interact. Instead, we trace their emergence and interplay over pretraining. To this end, we pretrain a multilingual $1.7$B model on a mix of data from nine typologically diverse languages, and capture a particularly high checkpoint density. We follow prior work in treating word-level translation (\wlt) as a testbed for cross-lingual transfer \cite{wendler-etal-2024-llamas}, reasoning that this translation capability is a precursor to more complex cross-lingual generalization. We construct a novel \wlt benchmark to go beyond simple words studied in prior work. %

Our evaluation shows that in the very early stages of training, word-level translation is enabled by token\footnote{We use ``token'' to refer to a subword unit produced by a tokenizer, historically also referred to as ``subtoken''.} overlap, while at the same time the model frequently erroneously copies tokens. We identify two phases of learning: Phase~I, in which the model rapidly acquires linguistic knowledge and a useful, but overeager copying mechanism, and Phase~II, in which the model develops translation aided by token overlap, and, more slowly, the ability to translate \emph{without} token overlap (see \cref{fig:copying_error_mode}). We hypothesize these phases to reflect the evolution and interaction of two mechanisms: token-level copying \cite{feucht2025the}, and a generalizing mechanism implementing shared representation spaces \cite{wendler-etal-2024-llamas}.

We study the training dynamics of these mechanisms, combining data attribution, model component analysis, and parameter-based ablation, to offer a unified view. Our analysis sheds light on how the copying mechanism emerges and refines, how the model develops the generalizing mechanisms which enable translation \emph{without} overlap, and how it learns to prefer translation over copying.

Our contributions are as follows:
\begin{itemize}
\item A release of pretraining checkpoints of a multilingual $1.7$B model,\!\footnote{\url{https://huggingface.co/mainlp/copy-first-translate-later}} more fine-grained (every $185$M tokens) and capturing earlier training than any other publicly available multilingual checkpoints.
\item A novel, carefully curated word-level translation dataset\footnote{\url{https://huggingface.co/datasets/mainlp/word-level-translations}} across 72 language pairs, stratified over PoS tag and English word frequency, with language-pair specific valid translations.
\item A behavioral analysis of word-level translation and basic linguistic capabilities over early multilingual pretraining, identifying two main phases of learning.
\item A multi-method interpretability study attributing this behavior to model components, uncovering the emergence and interaction of two central mechanisms: a useful, albeit overeager copying mechanism, which is refined over training, and generalizing mechanisms which enables translation without overlap.
\end{itemize}

\section{Related Work}
A rich body of work investigates cross-lingual transfer, identifying beneficial factors such as morphological, syntactical, and lexical similarity \cite{philippy-etal-2023-towards, hammerl-etal-2024-understanding}. A beneficial factor relevant to our work is \emph{token overlap} \cite{wu-dredze-2019-beto,patil-etal-2022-overlap,limisiewicz-etal-2023-tokenization,bagheri-nezhad-etal-2025-beyond}. \citet{kallini-etal-2025-false} study the function of token overlap, training bilingual models with varying levels of overlap between the monolingual corpora. However, we focus on the related mechanism of token-level copying at inference, as studied by \citet{feucht2025the}, who identify token induction heads, which develop early in language models, and concept induction heads, with the latter critical for word-level translation.

Token-level copying does not account for translation of word pairs \emph{without} shared tokens, e.g., across scripts. In such cases, a different mechanism is at play. \citet{binbinliu2026token} identify token \emph{alignment} heads, which map source tokens to target tokens during translation even without overlap. \citet{wendler-etal-2024-llamas} suggest that such translation is mediated by shared internal representations, referred to as a shared ``concept space''. \citet{dumas-etal-2025-separating} provide causal evidence that concept and language are encoded independently in latent representations.  We study the emergence of the mechanism implementing shared representations.

In contrast to previous work that studies trained models, we argue for studying training dynamics to understand the emergence of unsupervised cross-lingual transfer. \citet{voita-etal-2021-language} is similar to our work in spirit, finding that \wlt emerges early in neural machine translation (NMT). However, NMT relies on parallel data and a cross-lingual training objective, whereas we study the emergence of cross-lingual ability \emph{without} such explicit cross-lingual signals. Studies of large language models are hampered by the limited availability of multilingual checkpoints, which, if released at all, are typically sparse and provide limited information about pretraining data \cite{apertus2025apertusdemocratizingopencompliant}. \citet{riemenschneider-frank-2025-cross} and \citet{zeng-etal-2025-converging} focus primarily on BLOOM \cite{DBLP:journals/corr/abs-2211-05100}, while \citet{korner-etal-2026-meanings} focus on EuroLLM $1.7$B \cite{martins2025eurollm9btechnicalreport}. All three focus on the development of cross-lingual shared neurons or internal representations.

By contrast, we pretrain our own model, enabling fine-grained checkpoint density and control over pretraining data. More importantly, we combine these lines of research, studying both token-level copying and generalizing mechanisms and how they develop and interact in early pretraining.

\section{Model Pretraining}\label{sec:pretraining}
To form the basis of our study, we pretrain a LLaMA-style causal decoder-only language model \cite{touvron2023llamaopenefficientfoundation} with $1.7$B parameters from scratch on nine languages: English, French, German, Indonesian, Italian, Portuguese, Spanish, Mandarin Chinese, and Japanese, covering three different script families. This selection provides naturally varying levels of token overlap between language pairs, which we confirm in \cref{sec:jsd}. English training data is drawn from FineWeb-HQ, the remaining languages from FineWeb2-HQ \cite{messmer2025enhancing_fineweb}.

Each batch consists of a multilingual mixture of sequences sampled according to probabilities of $0.5$ for English, and $0.0625$ for each of the remaining languages. Importantly, we train without a cross-lingual objective and do not include \emph{targeted} cross-lingual training signals such as parallel data. Instead, we randomly sample from each language's partition at each step to mirror realistic training and mitigate any order-based effects. Note that this nonetheless results in small amounts of \emph{incidental} parallel data, which we analyze in \cref{sec:parallel}.

We capture model checkpoints every $185$M tokens, providing $200$ snapshots across training for our fine-grained analysis. We share our code,\!\footnote{\url{https://github.com/mainlp/copy-first-translate-later}} and report training details in \cref{app:details_pretraining}.

\section{Evaluation}\label{sec:eval}
We use two main tasks to track our model's monolingual and cross-lingual ability: linguistic acceptability and word-level translation.
\subsection{Linguistic Acceptability}
The first task evaluates linguistic capabilities using BLiMP-style minimal pairs \cite{warstadt-etal-2020-blimp-benchmark}. For English, German, French, Spanish, Italian, and Portuguese, we use the respective subsets of MultiBLiMP \cite{jumelet2025multiblimp10massivelymultilingual}, which focuses on subject-verb agreement. Japanese, Chinese, and Indonesian are not included in MultiBLiMP. We use JBLiMP \cite{someya-oseki-2023-jblimp} for Japanese and ZhoBLiMP \cite{DBLP:journals/corr/abs-2411-06096} for Chinese, subsampling phenomena for comparability with MultiBLiMP. For Indonesian, no equivalent benchmark designed for log-probability scoring is available to the best of our knowledge. See \cref{app:blimp-phenomena} for details, including subsampling criteria.
\newlength{\figtwoaw}
\newlength{\figtwobw}
\newlength{\figtwocw}
\settowidth{\figtwoaw}{\includegraphics[height=3.2cm]{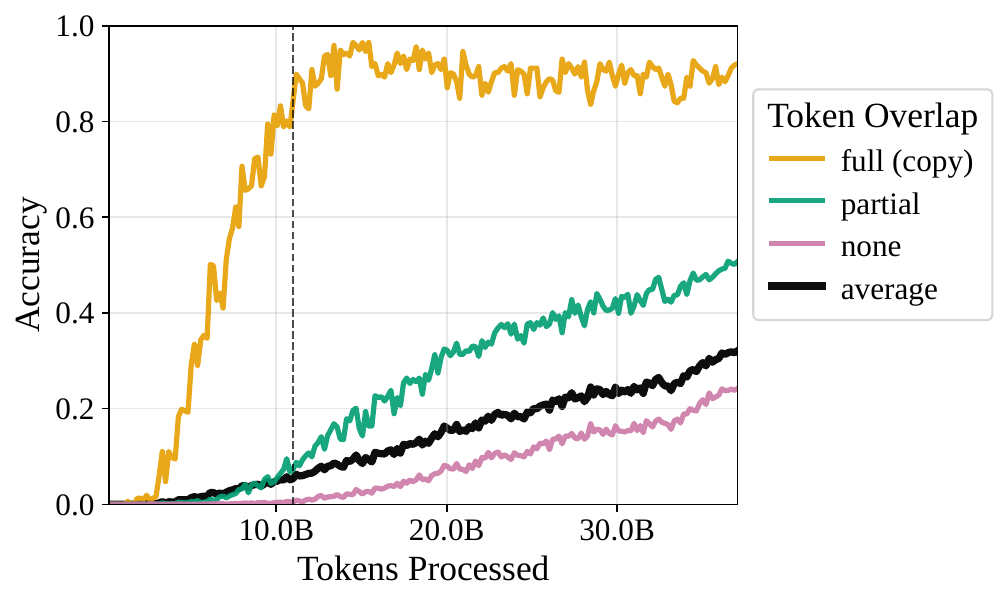}}
\settowidth{\figtwobw}{\includegraphics[height=3.2cm]{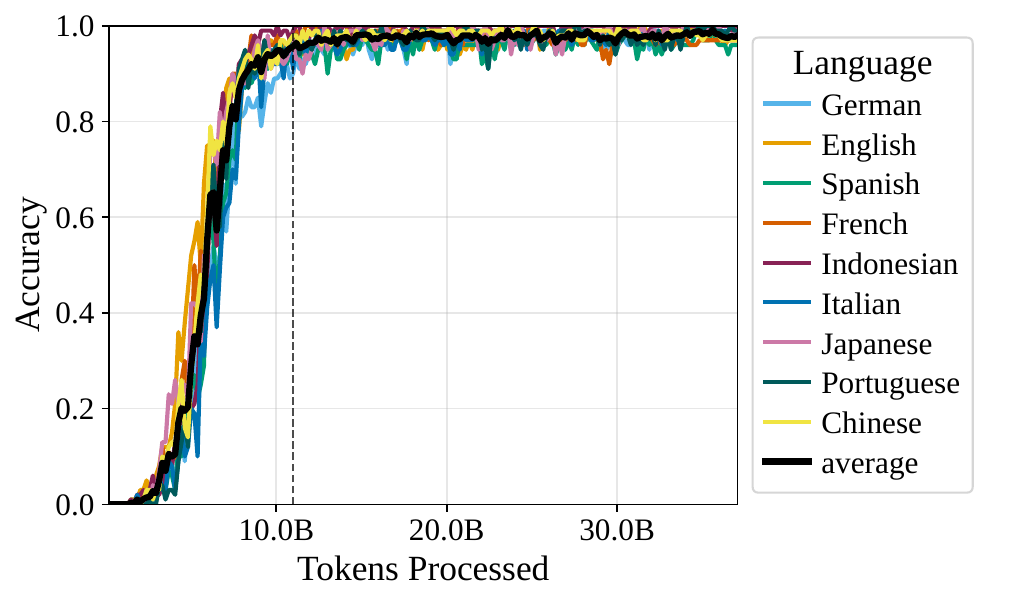}}
\settowidth{\figtwocw}{\includegraphics[height=3.2cm]{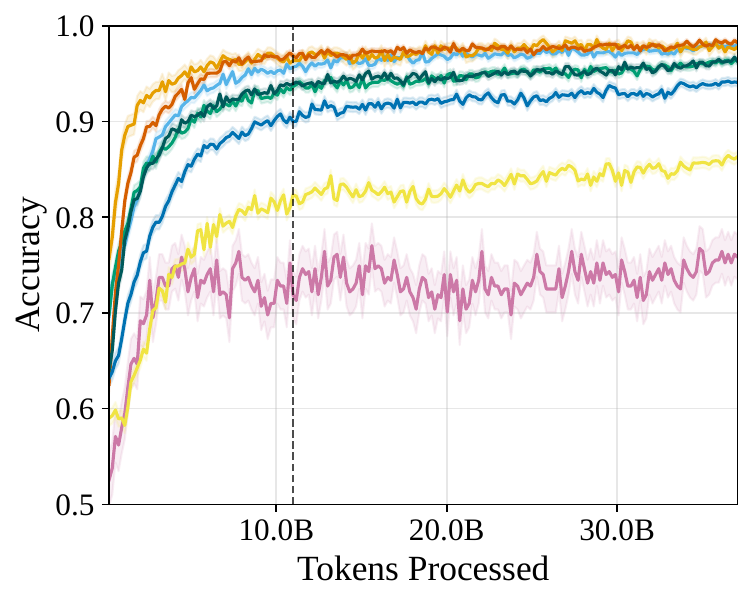}}

\begin{figure*}[ht]
    \centering
    \begin{subfigure}[t]{\figtwoaw}
        \centering
        \includegraphics[height=3.2cm]{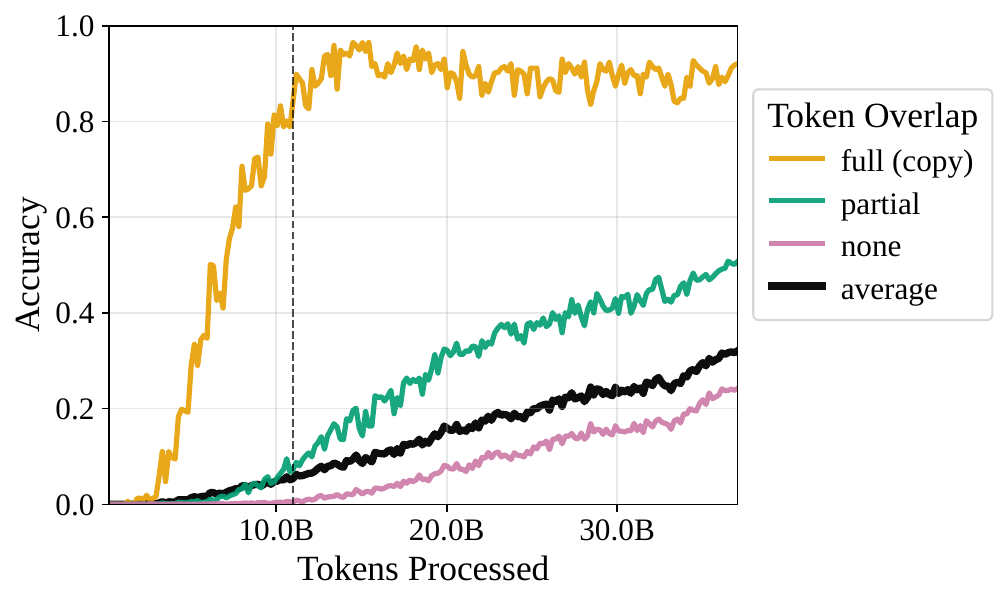}
        \caption{\wlt accuracy per token overlap bucket and average over all language pairs.}
        \label{fig:accuracy_by_bucket}
    \end{subfigure}%
    \hspace{0.005\linewidth}%
    \begin{subfigure}[t]{\figtwobw}
        \centering
        \includegraphics[height=3.2cm]{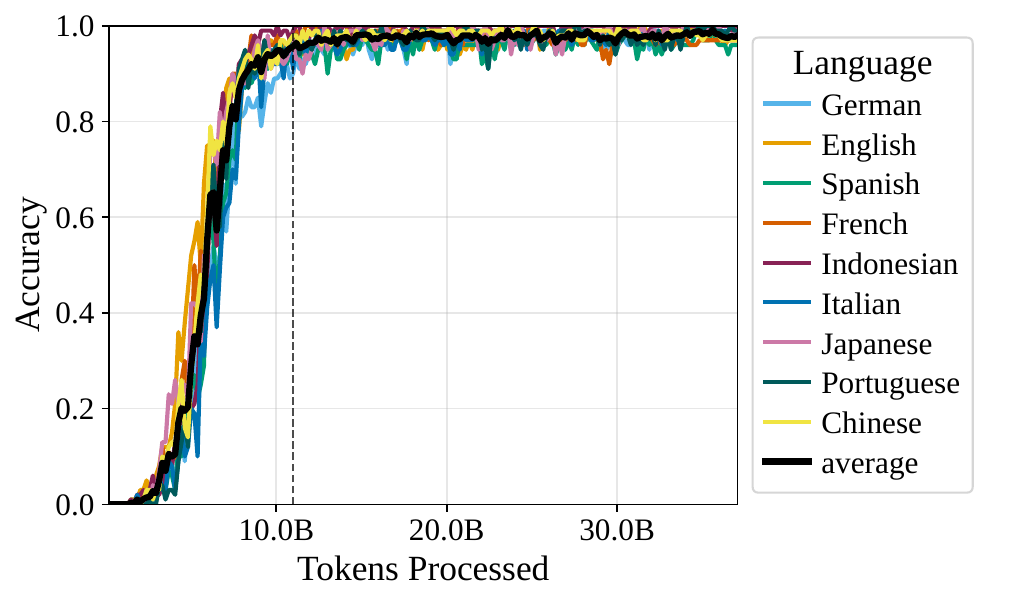}
        \caption{Repetition accuracy per language.}
        \label{fig:repetition_success_rate}
    \end{subfigure}%
    \hspace{0.005\linewidth}%
    \begin{subfigure}[t]{\figtwocw}
        \centering
        \includegraphics[height=3.2cm]{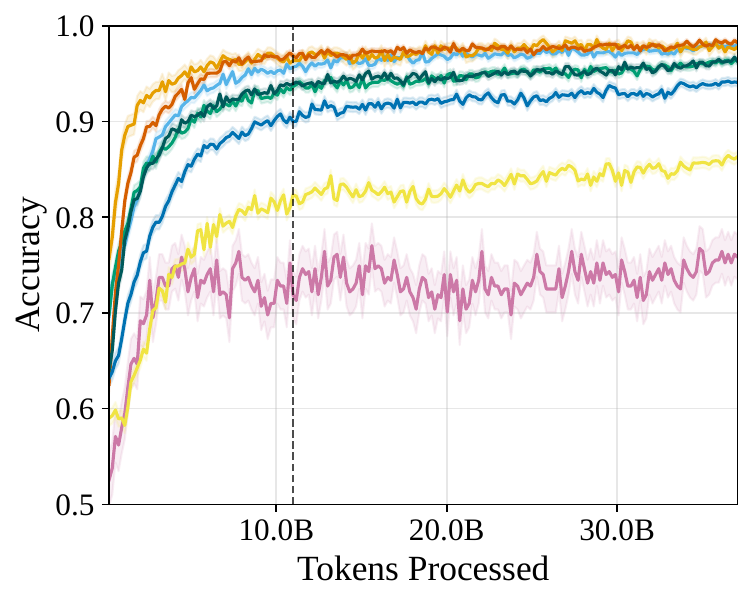}
        \caption{MultiBLiMP, JBLiMP and ZhoBLiMP accuracy.}
        \label{fig:blimp-evolution}
    \end{subfigure}
    \caption{Evolution of model performance across diverse linguistic tasks over training.}
    \label{fig:combined_metrics}
\end{figure*}

\subsection{Word-Level Translation}\label{sec:data-wlt}
The second task is word-level translation (\wlt), taken to study cross-lingual transfer. Prior interpretability studies of cross-lingual generalizing mechanisms have largely focused on frequent, simple nouns, partly due to methodological constraints such as reliance on unique first tokens \cite{wendler-etal-2024-llamas,dumas-etal-2025-separating}. Since we aim to study natural language use, we construct a novel dataset, balanced across word class and frequency ranks, and allowing for inflection. To do so, we curate language-pair specific valid translations for each source word, allowing polysemous words and words belonging to multiple word classes, e.g., ``dance'' may be translated to both the verb ``tanzen'' and the noun ``Tanz'' in German.

Our concepts are drawn from ChiKhaPo, a multilingual lexical benchmark \cite{Chang2025ChiKhaPoAL}.  We use ``concept'' to describe an abstract idea that can be expressed by different words. We randomly select 125 concepts, balanced across English frequency (as determined by fastText,\!\footnote{\href{https://fasttext.cc/docs/en/crawl-vectors.html}{cc.en.300.bin}}
\citealp{grave-etal-2018-learning}); we discard any word that is not in the top 20k most frequent. Additionally, we ensure coverage of English word classes (as determined by WordNet\footnote{\url{https://wordnet.princeton.edu/}} via NLTK, \citealp{Bird2006NLTKTN}), selecting 40 nouns, 40 verbs, 25 adjectives, and 20 adverbs.

We split concepts into 25 few-shot and 100 query examples. Importantly, in the query set, we \emph{allow} for identical cognates, such as ``ala'' (meaning ``wing'' in both Spanish and Italian) to reflect natural language use\textemdash these make up $4.4\%$ of word pairs across language pairs (see Appendix \cref{fig_app:copy_stat}). However, we require that few-shot examples are unique across languages to prime the model to translate instead of copy. Given this constraint, we secondarily maintain coverage of PoS and frequency bins in the few-shot set on a best-effort basis.

The resulting multi-way parallel dataset includes 125 concepts aligned across our nine languages, with language-pair specific valid translations for the 100 query concepts. We detail our dataset construction process in \cref{app:details_wlt}.

\paragraph{Prompt Construction}\label{sec:prompt}
We follow \citet{wendler-etal-2024-llamas} for prompt construction. Given $\ell_{src}/\ell_{tgt}$ as the source and target language and $w_{src}/w_{tgt}$ as the source and target word for the same concept, each prompt consists of five examples, formulated as $\ell_{src}$: ``$w_{src}$'' - $\ell_{tgt}$: ``$w_{tgt}$'', followed by a query ending in $\ell_{src}$: $w_{src}$ - $\ell_{tgt}$: ``. For example, for the source word ``dance'':
\begin{tcolorbox}[colback=gray!5,colframe=gray!40,boxsep=2pt,left=4pt,right=4pt]
\texttt{English:~``field''~- Deutsch:~``Feld''} \\
\textit{\texttt{$[$...four more examples...$]$}}\\
\texttt{English:~``dance''~- Deutsch:~``}
\end{tcolorbox}
$\ell_{src}/\ell_{tgt}$ are expressed in their own language, in this case \emph{Deutsch} for $\ell_{tgt}$ German. We also evaluate a variant of this task, referred to as \emph{repetition}, where source and target language are equivalent and the word simply has to be copied.

We measure case-insensitive mean accuracy, counting a word as correctly translated if it matches one of the synonyms. For the repetition task, we count a word as correctly repeated if it matches the source word (see \cref{app:wlt-exp} for further details).

\section{Model Learning Phases}\label{sec:learning-phases}
\wlt accuracy improves gradually, reaching a final mean accuracy of $32.1\%$ (\cref{fig:accuracy_by_bucket}, see \cref{app:wlt-eval} for language-specific results).
Error analysis shows that copying is the dominant error mode, peaking at $53.2\%$ of errors attributable to erroneous copying across all language pairs (\cref{fig:copying_error_mode}).
We also observe cases where token overlap misleads the model. For example, when asked to translate ``\textbf{irr}tümlich'' (mistaken) from German to French, the model outputs ``\textbf{irr}éparable'' (irreparable) instead of ``erroné'', copying the initial token ``\textbf{irr}'' and continuing in the target language with the wrong meaning.
However, token-level copying can also be a useful strategy: e.g., ``incorrectamente'' (meaning: incorrectly) is a valid translation from Spanish to Portuguese. We refer to this as \emph{valid copying}; our \wlt dataset contains $4.4\%$ such instances (see Appendix \cref{fig_app:copy_stat} for statistics on token overlap in the \wlt dataset).

We group word pairs into token-overlap buckets in \cref{fig:accuracy_by_bucket}, which shows that word pairs with full overlap are learned first, and those with \emph{partial} overlap are learned earlier and more quickly than those without. Word pairs without overlap must necessarily be translated via a different mechanism than token-level copying\textemdash we hypothesize that this is the generalizing mechanisms implementing shared concept spaces \cite{wendler-etal-2024-llamas}.

The model must therefore develop and balance these two mechanisms: how and \emph{whether} to copy tokens, and how to translate without token overlap. We identify two phases characterizing this learning process in our model, where we set the phase boundary at the qualitative milestone when words without token overlap are first translated.

\paragraph{Phase I -- Development of Copy Mechanism}\label{sec:early_training}
We first observe a rapid increase in performance on both the repetition task (\cref{fig:repetition_success_rate}) and linguistic acceptability (\cref{fig:blimp-evolution}). By $9.5$B processed tokens, the model reaches $94.1\%$ average accuracy for repetition and surpasses $90\%$ for each MultiBLiMP subset. Accuracy climbs similarly for ZhoBLiMP and JBLiMP, which test more phenomena, though the latter is slightly noisier, likely due to its smaller size (\cref{app:blimp-phenomena}). In parallel, the loss declines steeply (Appendix \cref{fig:loss_curve}), generalizing \citet{chen2025suddendropslosssyntax}'s findings in an English-only encoder model to the multilingual setting.

In \wlt, copying is increasingly the dominant error mode, plateauing between $7-15$B tokens processed at up to $56.3\%$ of model \emph{outputs} (\cref{fig:copying_error_mode}). We observe two main variants of copying: \emph{context copying}, where the model copies a few-shot example, and \emph{source word copying}, where it copies the source word. We consider context copying less \emph{targeted}, as source word copying indicates that the model attends to the right part of the prompt.

During Phase I, copying becomes more targeted. As shown in \cref{fig:copying_error_mode}, the rate of context copying rises to $33.6\%$ of outputs until $6.7$B tokens processed, and then falls for the rest of training. The rate of source word copying begins to rise later, overtaking context copying after $7.5$B tokens. A similar pattern appears in the repetition task, where context copying is even more prevalent, but also decays more quickly over training (see Appendix \cref{fig:app_copy_rates_full}).

\paragraph{Phase II -- Transition to Translation}
At around $11$B tokens processed, we observe first successful predictions in \wlt for word pairs \emph{without} token overlap (\cref{fig:accuracy_by_bucket}), indicating the development of generalizing mechanisms. Meanwhile, erroneous copying begins a steady decline that continues for the rest of training, suggesting the model's diminishing reliance on copying (\cref{fig:copying_error_mode}).

\section{Interpretability Methods}\label{sec:interp}
To understand how token-level copying and generalizing mechanisms develop and interact over our model's learning phases we study the training dynamics of its internal components. We apply two complementary interpretability methods over time: logit lens \cite{nostalgebraist2020logitlens} to track how prediction candidates compete over layers, and ExPLAIND \cite{eichin2025explaindunifyingmodeldata} to identify what model components and data drive these predictions.

\subsection{Logit Lens: Layer-Wise Analysis}\label{sec:logit lens}
We use the logit lens \cite{nostalgebraist2020logitlens} to analyze the representation of concepts in our nine languages at different layers and checkpoints. This method projects intermediate hidden states into the vocabulary space by passing the last token's activation from each layer through the model's final normalization, unembedding, and log softmax. For multi-token words, we utilize teacher forcing: we sequentially append the ground-truth tokens of a candidate word to the context and extract the intermediate log probability for each successive token. We normalize the sum of these log probabilities by length to compare candidates.

\paragraph{Multilingual Analysis}
We apply logit lens to each concept (\cref{sec:data-wlt}). To account for lexical variation we evaluate all synonyms and select the highest-scoring one at the final layer. Although each prompt specifies a single target translation direction (e.g., English to German), we repeat the process for all nine languages, which yields the concept's score for each language. This lets us track layer-wise scores for a concept across the model and compare relative language-(in)dependent trends.
\paragraph{Copy vs.\ Translate Analysis}
To analyze how target (translation) and source (copy) candidates compete over layers, we define the translation-over-copy margin:
$m_L=\log p_L(w_\text{tgt})-\log p_L(w_\text{src})$
where $w_\text{tgt}$ is the translation candidate,  $w_\text{src}$ the copy candidate and $\log p_L(w)$ the (normalized) log probability of a word at layer $L$, as well as its layer-wise difference
$$\Delta m_{L\to L+1}=m_{L+1}-m_{L}.$$
A \blue{negative} $\Delta m_{L\to L+1}$ indicates a layer transition where the source word is preferred over the target word\textemdash we refer to such transitions as \emph{\blue{copy-promoting}}. Conversely, a \red{positive} $\Delta m_{L\to L+1}$ indicates that the transition is \emph{\red{translation-promoting}}.
\begin{figure*}
\begin{subfigure}[t]{.49\linewidth}
\centering
    \includegraphics[width=\linewidth]{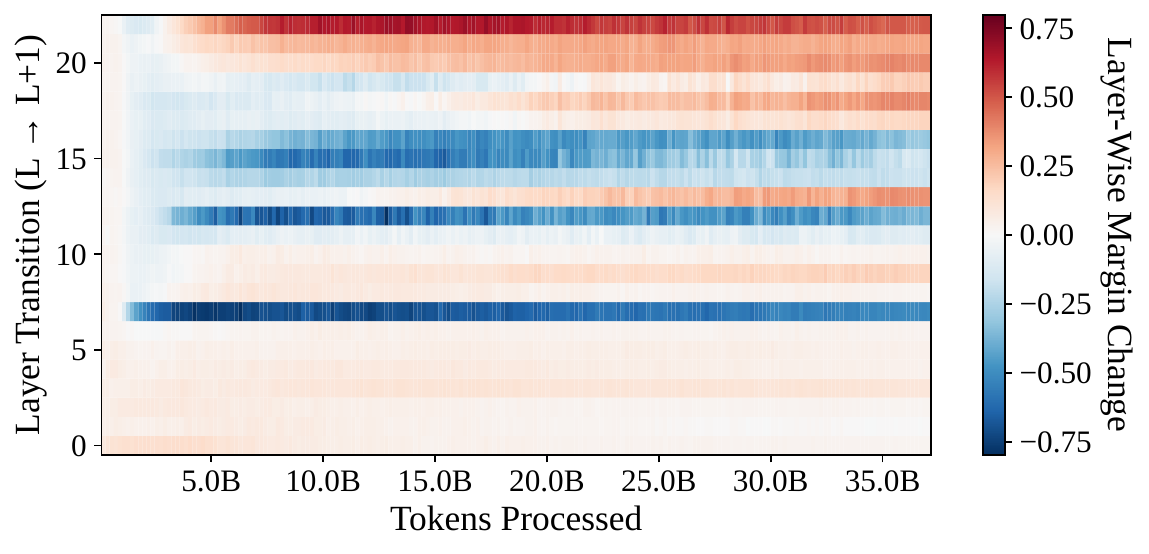}
    \caption{\textbf{Evolution of the translation-over-copy margin}. \blue{Blue} and \red{red} indicate \blue{copy-} and \red{translation-}promoting layer transitions, respectively. Prominent \blue{copy-promoting} layers appear as blue stripes, while \red{translation} accumulates a small relative gain (light red area) across most layers, increasing towards model output and as training progresses.}
\label{fig:margin_gradient_ll}
\end{subfigure}
\hfill
\begin{subfigure}[t]{.49\linewidth}
    \centering
    \includegraphics[width=\linewidth]{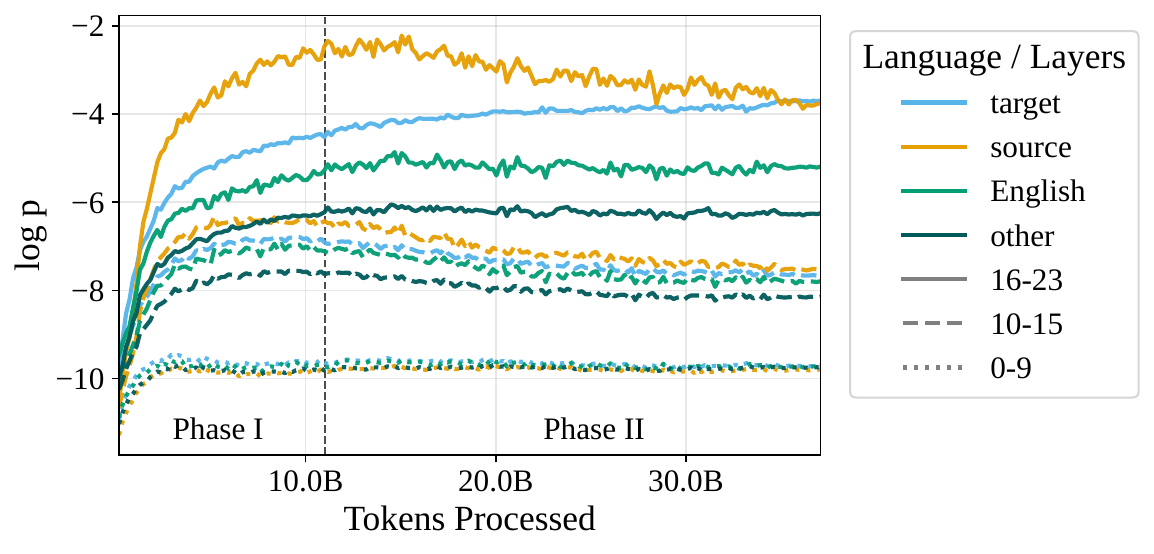}
    \caption{\textbf{Evolution of different model blocks across training steps.} The vertical line marks the transition to Phase II, where the copying mechanism gets suppressed by the evolving translation capacity. We provide more fine-grained layer-wise evolutions in Appendix \cref{fig:app_over_layers_steps}.}
        \label{fig:main-ll-over-time}
\end{subfigure}
\caption{\textbf{Multilingual logit lens analysis over training}. We aggregate results across all $72$ language pairs.} 
\label{fig:main_logit_lens}
\end{figure*}

\subsection{ExPLAIND: Attribution over Model Components and Data}\label{sec:methodology_explaind}
We use ExPLAIND \cite{eichin2025explaindunifyingmodeldata} to identify relevant model parameters and quantify the influence of different types of training data on the two mechanisms. We use the framework to decompose the loss trajectory $L(\theta_s, x)$ of an unseen sample $x$ onto model parameters $\theta_s^{(i)}$
at step $s$ %
and training samples $x_k$, %
computing influence scores $\phi(s, \theta_s^{(i)}, x_k, x) \in \mathbb{R}$ which add up to $L(\theta_s, x)$.

We subsample the training steps $s$ to every 500th step, i.e., each one of our checkpoints. Since we are interested in global rather than local influences, we accumulate influence scores over collections of samples for predictions and training data and over parameters to the layer level (following \citealp{eichin2025explaindunifyingmodeldata}). Thus, for a step $s$ the influence through parameters $\Theta_s$ of a training data partition $X$ onto a set of predictions $B$ is:
\begin{equation*}
    \Psi(s, \Theta_s, X, B) = \sum_{\hat{x} \in B} \sum_{x \in X}\sum_{\theta \in \Theta_s} \phi(s, \theta, x, \hat{x}).
\end{equation*}
Negative influences imply a decrease in loss, i.e., the model is more likely to make the prediction $x$.

\paragraph{Identifying\ Copy-Promoting \&\ Copy-Suppressing Parameters}
To isolate parameters involved in the competition between copy and translate we study the influence of simulated training data on erroneously copied and successfully translated predictions. We hypothesize that explicit parallel data as well as target word occurrences encourage translation. We construct such data, $X_{parallel}$ (see \cref{app:explaind_train_data}) and combine it with training batches of our checkpoints to simulate training data.

We sample a set of erroneously copied predictions ($B_{copy}$), and corresponding correctly translated predictions ($B_{translate}$). Sets are aligned across source words: for each $w_{src}$, $B_{copy}$ includes $w_{src}\rightarrow w_{src}$, and $B_{translate}$ includes $w_{src}\rightarrow w_{tgt}$ (see \cref{sec:explaind_test_data}).

We define
\begin{equation*}
\begin{split}
    D_{copy}(s, \Theta) := & \Psi(s, \Theta, X_{parallel}, B_{copy})\\
    & - \Psi(s, \Theta, X_{parallel}, B_{translate}).
\end{split}
\end{equation*}
and consider the most negative parameter groups in terms of $D_{copy}$ during Phase I as \emph{\blue{copy-promoting}}, and the most positive during Phase II as \emph{\red{copy-suppressing}}. To identify sparse groups of parameters, we rank parameters by the corresponding influences and only include parameters with at least $50\%$ of the first-ranked parameter's influence.

\section{Model Component Analysis}
In the following sections, we use ExPLAIND and the logit lens to examine the mechanisms behind the behavior observed in \cref{sec:learning-phases}: the copying mechanism (\cref{sec:copy-heuristic}) and generalizing mechanisms (\cref{sec:generalizing}). We identify relevant model components, and then study their role by performing ablations: parameter scaling and layer swapping. Parameter scaling gives insight into the role of specific parameter groups, whereas layer swapping helps isolate the development progress of blocks of layers; we detail the methodology below. Combining interpretability methods with causal ablations allows us to understand which model components are responsible and how they develop even before corresponding behavioral changes.

\subsection{Copy Mechanism}\label{sec:copy-heuristic}
The logit lens margin plot identifies several \blue{copy-promoting} layer transitions: $7\to8$, $12\to13$, and more weakly $15\to17$ (\cref{fig:margin_gradient_ll}). Throughout Phase II, the layer-wise dynamics of copying change little; the \blue{copy-promoting} layer transitions established during Phase I persist. However, in Phase II the target candidate rises steadily while the source candidate (i.e., copy) is progressively downranked, indicating an increasing preference for translation (\cref{fig:main-ll-over-time}), also reflected by the increasingly red area in the margin plot (\cref{fig:margin_gradient_ll}). While the logit lens offers insights into the preferred prediction at a particular layer, it does not reveal which components influence these. To understand how parameters in the model drive the observed changes, we turn to ExPLAIND.

\begin{figure}[t]
    \centering
    \includegraphics[trim=0 40 0 73,clip,width=\linewidth]{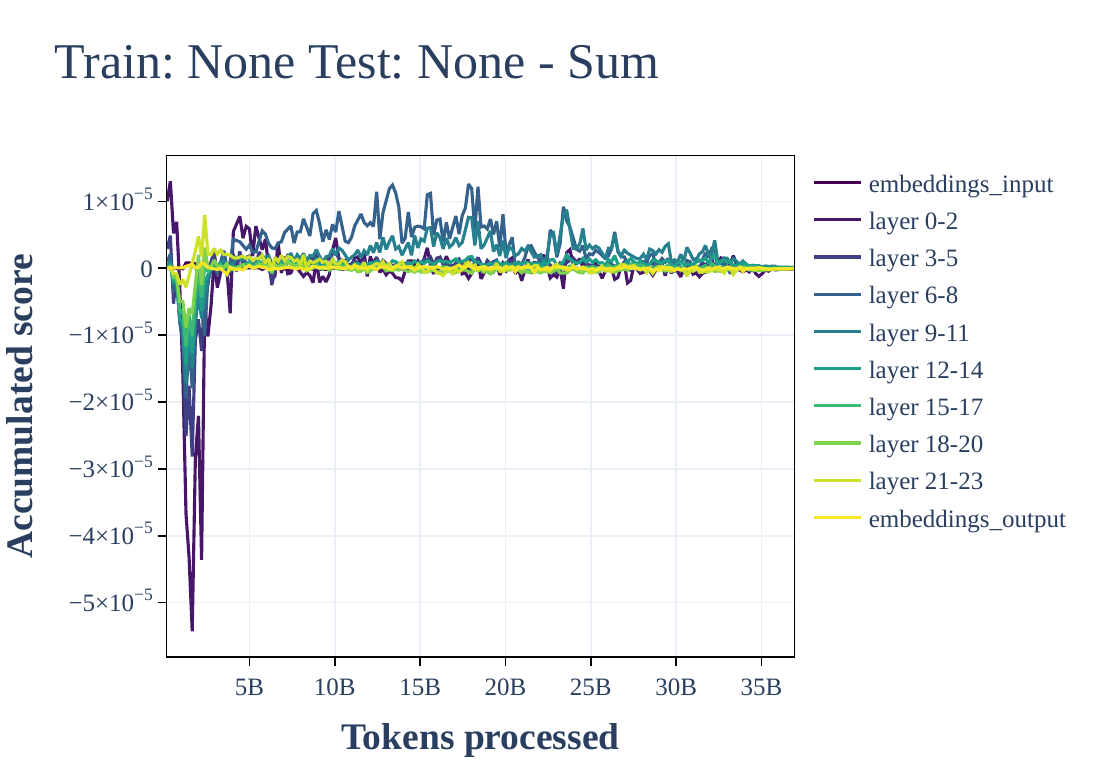}
    \textbf{\scriptsize \color{plotly} \hspace{-22pt} Tokens Processed}
    
    \caption{We simulate the effect of training on parallel data containing the concepts of challenging instances and plot the difference in influence on copy behavior $D_{copy}(s, \Theta^{(i, i+2)})$. A negative score indicates \blue{copy promotion} and a positive score indicates \red{copy suppression}.
    We study the effect of actual data in \cref{app:explaind_actual}.
    }
    \label{fig:parallel_on_copy_explaind}
\end{figure}

We use ExPLAIND to identify \blue{copy-promoting} parameters and \red{copy-suppressing} parameters (\cref{sec:methodology_explaind}). \cref{fig:parallel_on_copy_explaind} shows influence scores aggregated over groups of three consecutive layers. A negative score indicates \blue{copy promotion}, and a positive score indicates \red{copy suppression}.

\paragraph{Ablation: Parameter Scaling}
We investigate the role of each group of parameters by performing two causal ablations:
1) we scale parameters by $2$ to \emph{excite} their influence on predictions, 2) we scale them by $0.5$ to \emph{inhibit} their influence. We then compare the ablated checkpoint's copy and translation behavior to that of the corresponding original checkpoint. To keep our interventions sparse, we intervene on only the parameters which have at least $50\%$ of the first-ranked parameter's influence (listed in \cref{sec:copy_ablation_additional_details}).

\paragraph{Copy-Promoting Parameters}
ExPLAIND identifies the attention value projections in layers $0$--$2$ as the top contributors of \blue{copy-promoting} influence in Phase I. \emph{Exciting} their influence makes copy behavior dominant over all checkpoints; source word and context copying are almost tripled in the final checkpoint. \emph{Inhibiting} them reduces source word copying with only a minor impact on translation performance ($<4$ percentage points), primarily affecting instances where token copying produces correct translations; it has little effect on context copying (see Appendix \cref{fig:copy_ablations_prom}). \textbf{This indicates that the parameters in layers $\mathbf{0}$--$\mathbf{2}$, identified as copy-promoting via ExPLAIND, control \emph{whether} to copy.}

\paragraph{Copy-Suppressing Parameters}
ExPLAIND identifies parameters in layers $6$--$9$ as the top contributors of \red{copy-suppressing} influence in Phase II (see positive scores in \cref{fig:parallel_on_copy_explaind}). These parameters play a more complicated role than the parameters in layers $0$--$2$, aligning less cleanly with copy suppression. Both exciting and inhibiting their influence results in higher context copying throughout training, indicating that the parameters are sensitive to \emph{where} to copy from. The effect on source word copying shifts over training: in the first half of training both ablations reduce source word copying. However, from around $20$B tokens processed, exciting the parameters results in \emph{higher} source word copying than the original checkpoints, whereas inhibiting them has little effect. We hypothesize that in later training, the parameters also contribute to the decision of whether to copy, but are not necessary for this decision (hence, source word copying does not decrease when they are inhibited). \textbf{Together, this indicates that the parameters in layers $\mathbf{6}$--$\mathbf{9}$, identified as copy-suppressing via ExPLAIND, primarily control \emph{where} to copy from.}

Overall, the parameter groups in these lower layers ($0$--$2$ and $6$--$9$) appear to drive much of the copy behavior, controlling whether to copy and where to copy from. We hypothesize that the increasing \red{translation signal} seen in \cref{fig:margin_gradient_ll} and the shift in ablation behavior for parameters in $6$--$9$ reflect the development of generalizing mechanisms, which we investigate in the following section.

\begin{figure}
    \centering
        \centering
        \begin{minipage}[t]{.52\linewidth}
            \includegraphics[trim=0 0 0 0.2cm,clip,height=4.6cm]{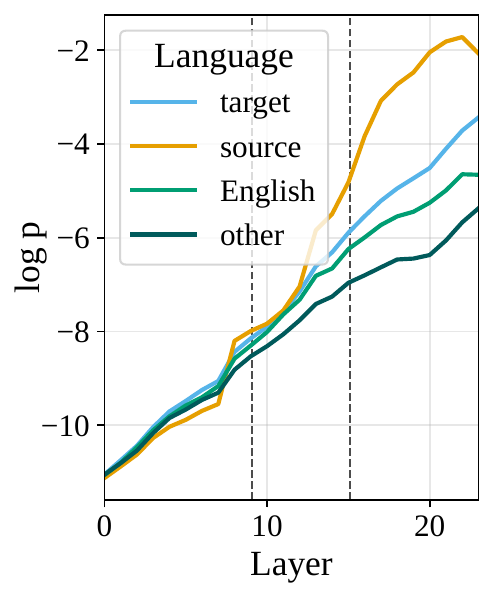}
            \subcaption{$11$B Tokens}\label{fig:early}
        \end{minipage}
        \hfill
        \begin{minipage}[t]{.46\linewidth}
            \includegraphics[trim=1.6cm 0 0 0.2cm,,clip,height=4.6cm]{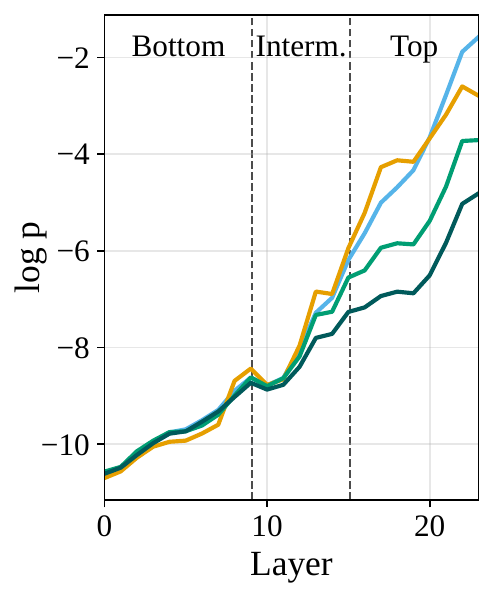}
            \subcaption{$37.5$B Tokens}\label{fig:final}
        \end{minipage}
        \caption{\textbf{Evolution across layers.} We trace concept log probability for different candidates for our proposed layer blocks (bottom, intermediate, and top) at the phase transition (\cref{fig:early}) and the final checkpoint (\cref{fig:final}).}
        \label{fig:main_ll-language_splitting}
\end{figure}

\subsection{Generalizing Mechanisms}\label{sec:translation}
To identify relevant parameters for the generalizing mechanisms, we turn to prior work, which casts multilingual processing as a three-stage process: first, the \emph{bottom} layers map from the input language to an \emph{intermediate} shared space, second, inputs are processed in the intermediate space, and third, the \emph{top} layers map to the output language \cite{DBLP:conf/nips/0006ZCKB24}. Our logit lens analysis of the final checkpoint supports this\textemdash candidates across languages only separate in the top block, as the correct output language is chosen (\cref{fig:main_ll-language_splitting}). However, it does not reveal how the generalizing mechanisms in the intermediate layers develop.

\paragraph{Ablation: Layer Swapping}
We use targeted layer-swapping experiments to isolate how blocks of layers responsible for each stage develop. We fix blocks to their final-checkpoint state and pair them with complementary, unfixed blocks from other checkpoints. Importantly, we combine frozen blocks from different points in training (following \citealp{DBLP:journals/corr/abs-2410-01335}), i.e., we do not conduct any further training. This approach gives us a snapshot of the unfixed blocks' state at each checkpoint. 

Given the three-stage hypothesis, we group the model's layers into \textit{bottom} (layers $0$--$9$), \textit{intermediate} ($10$--$15$), and \textit{top} ($16$--$23$) blocks. We split based on observable model behavior: The bottom block includes copy-promoting layers (\cref{sec:copy-heuristic}), and the upper block begins where the logit lens shows other language candidates are downranked relative to source and target (indicated in \cref{fig:main_ll-language_splitting}). We expect that the final bottom and top block allow mapping in and out of the generalizing mechanisms in the final intermediate block.

\begin{figure}
    \centering
    \includegraphics[trim=0 0 50 60,clip,height=5cm]{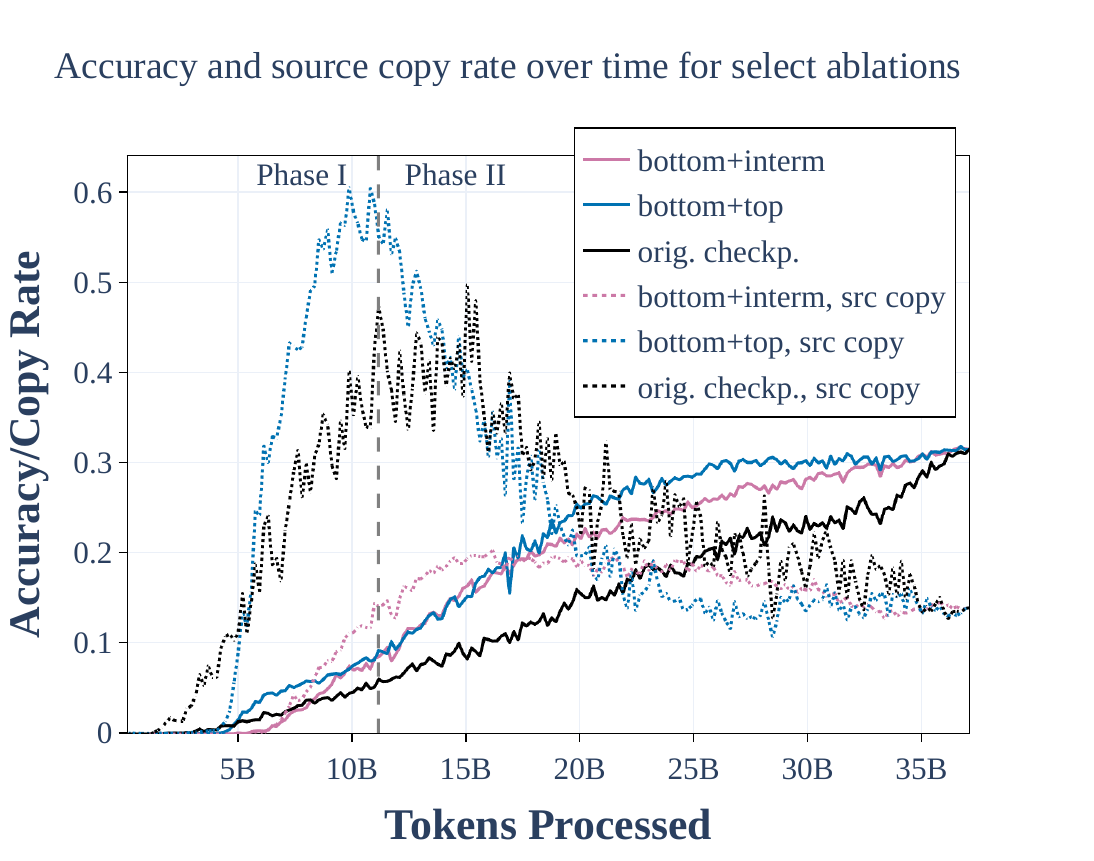}
    \caption{\textbf{Layer swapping over time.} We swap layer blocks of the final checkpoint into earlier checkpoints and compare \wlt accuracy of the resulting model to the (unchanged) original checkpoint.
    }\label{fig:main_layer_swapping}
\end{figure}
\paragraph{Emergence of Generalizing Mechanisms}\label{sec:generalizing}
We fix the bottom+top blocks to study the development of the generalizing mechanisms in the intermediate block. We expect that once the generalizing mechanisms are formed, the model can make use of them with the final bottom+top blocks. The ablation shows a striking trajectory: Over all checkpoints after only $10$B tokens, the ablation results in an improvement of up to $30\%$ of the final performance of the original model (\cref{fig:main_layer_swapping}). After a steep increase following the phase transition, the curve flattens, reaching $90\%$ of the final performance at roughly the midway point ($18.8$B tokens), and $98\%$ by $25$B tokens. \textbf{This indicates that the generalizing mechanisms are largely formed by $\mathbf{18.8}$B tokens, with smaller refinements until $\mathbf{25}$B.}

\paragraph{Interplay between Mechanisms}\label{sec:interplay}
In \cref{sec:copy-heuristic}, we identified parameters in the \emph{bottom} block as driving copy behavior. To understand the role of the remaining blocks in the copy vs.\ translate decision, we fix the bottom+intermediate blocks and compare to the bottom+top setting.
Fixing the bottom+intermediate blocks performs on par with the bottom+top block setting in the first half of training, roughly until the generalizing mechanisms are mostly formed, at which point it performs slightly worse. However, fixing the bottom+top blocks results in higher peaks of copying around the phase transition than the original, whereas fixing the bottom+intermediate blocks \emph{does not} produce a copying peak (\cref{fig:main_layer_swapping}). \textbf{This suggests that bottom+intermediate blocks \emph{together} suppress copying.} Furthermore, any combination with the final bottom block leads to some improvement in \wlt performance over the unchanged checkpoint, whereas fixing the final intermediate block alone does not (Appendix \cref{fig:appendix_swapping}).

The logit lens further supports the role of the bottom+intermediate blocks in the copy decision. We observe that correct translation candidates with partial or even full token overlap with the source word are downranked in the upper layers; such changes on the activation level of upper layers are driven by the layers below. Such candidates with overlap are discouraged even though the model ultimately correctly prefers them (see Appendix \cref{fig:app_over_time_token_groups}). Notably, we do not observe this downranking for correct candidates \emph{without} token overlap. \textbf{Together, this supports a mechanism that discourages token copying in the bottom and intermediate layers.}

\section{Discussion}
We study the development of two mechanisms that appear central to word-level translation (\wlt): token-level copying \cite{feucht2025the}, and generalizing mechanisms implementing shared concept spaces \cite{wendler-etal-2024-llamas}. Through behavioral analysis of our model, we identify two learning phases characterizing the evolution and interplay of these mechanisms: Phase I, in which token-level copying is learned, Phase II, in which generalizing mechanisms develop and copying is deprioritized (\cref{sec:learning-phases}). We further investigate the mechanisms through two complementary interpretability methods, identifying relevant model components and ablating them (\cref{sec:interp}). 

In \cref{sec:copy-heuristic}, we find that the copying mechanism is driven by parameters in the model's lower layers, where \textbf{parameters in layers $\mathbf{0}$--$\mathbf{2}$ decide \emph{whether} to copy, and parameters in layers $\mathbf{6}$--$\mathbf{9}$ learn \emph{where} to copy from.} However, our ablations suggest that the role of parameters in layers $6$--$9$ evolves over training, shifting around $20$B tokens processed, such that exciting them encourages source word copying beyond the original rate while inhibiting them has little effect\textemdash we hypothesize this relates to the interplay between the two mechanisms, discussed below.

Surprisingly, in \cref{sec:generalizing} we find that the \textbf{generalizing mechanisms enabling translation without token overlap are largely developed by $\mathbf{20}$B tokens processed, \emph{before} this is reflected in the model output.} Through our layer-swapping experiments, we discover that the final bottom and top layers enable translation on par with the final model via the generalizing mechanism in the intermediate layers at this point. This aligns with prior work which finds that shared spaces develop even before the model has learned to successfully use them to translate \cite{korner-etal-2026-meanings}, though our high checkpoint density allows us to observe this even earlier in training. Furthermore, this point coincides with the shift in ablation behavior for the parameters in layers $6$--$9$. Once the generalizing mechanisms have formed, these parameters appear to promote the source word as a fallback for the correct translation.

\emph{How do these two mechanisms interplay?} We find that the final bottom and intermediate layers \emph{together} deprioritize erroneous copying (\cref{sec:interplay}), while the bottom alone does not (\cref{sec:copy-heuristic}). Furthermore, we observe through the logit lens that valid translations with partial or full token overlap are increasingly downranked in the upper layers during Phase II, even though the model ultimately prefers these outputs. Importantly, we do not observe this downranking for translations without token overlap (see Appendix \cref{fig:app_over_time_token_groups}). Taken together, \textbf{we hypothesize that the generalizing mechanisms help suppress erroneous copying, in combination with the bottom layers.}

Finally, we consider possible future work enabled by our insights. In \cref{sec:copy-heuristic}, we identify sparse sets of parameters which control copying. An increasing body of work (and our paper) studies the importance of this early capability, identifying it as one of the first to develop \cite{liu2026what} and important for in-context learning \cite{sahin2026incontext}. Our simple parameter ablations show potential for controlling copying in a more sophisticated manner. Furthermore, mature copying parameters could be patched into earlier training to accelerate its emergence, following  \citet{baherwani2026emergentcapabilitiesariserandomly}.

\section{Conclusion}
We study the emergence of cross-lingual transfer, using the task of word-level translation as a testbed. We argue that studying the final model alone does not allow insights into how the underlying mechanisms develop and interact. To this end, we pretrain a multilingual $1.7$B model on nine diverse languages, and construct a word-level translation dataset. We identify two learning phases characterized by the development of two central mechanisms: token-level copying, and generalizing mechanisms that enable translation for word pairs without shared tokens. Using complementary interpretability approaches to analyze how these mechanisms emerge and interact, we find that token-level copying is a strong early strategy, learned in parallel with basic linguistic capabilities. As training progresses, generalizing mechanisms develop, coinciding with a decline in copying as an error mode, and more selective token-level copying. Together, these findings shed light on how incidental word-level translation, an early cross-lingual ability, develops in multilingual language models.

\section*{Limitations}
Our \wlt benchmark is designed as a controlled testbed for the emergence of cross-lingual transfer in early training, though it captures only one part of the broader phenomenon and leaves open how these dynamics extend to more complex tasks, such as phrase-level translation or reasoning-heavy tasks. Furthermore, its evaluation depends on curated synonym sets for each concept, which are not exhaustive and may miss valid translations, which can underestimate translation accuracy and bias comparisons toward better-covered languages.

We assess translation through in-context prompting since the model is not fine-tuned for translation; however, this choice may affect the model behavior. While our few-shot prompt follows established practice in multilingual interpretability for word-level translation for base models \cite{wendler-etal-2024-llamas}, it may introduce prompt-specific effects on copying behavior. For example, because none of the presented examples contains \emph{valid copying}, it may bias the model away from copying.

Although our studied model ($1.7$B parameters) is sufficiently large to exhibit non-trivial multilingual behavior, our analysis is restricted to this setting. The training run has not converged, so our conclusions concern the dynamics observed during early pretraining and may not fully extend to fully-trained multilingual models. However, we argue that early training is the most relevant window for the emergence of these mechanisms. Indeed, prior work has found related generalizing mechanisms to be mostly in place by the first checkpoints of publicly available multilingual models \cite{korner-etal-2026-meanings}. Furthermore, prior work has identified both token- and concept-level copying heads in fully trained models \cite{feucht2025the}. These heads relate to our mechanisms, as concept-level copying heads were found to enable translation, hence, there is evidence that our identified mechanisms remain relevant in converged models.

While our behavioral and interpretability analyses offer converging evidence for our model's learning phases, we stress that the boundary between these is behavioral\textemdash given a different \wlt dataset, or differing pretraining data, this boundary may appear at a different point in pretraining. However, we study the mechanisms that develop in each of these phases, and do not claim that the position of this boundary is universal. We randomly sample pretraining samples from each of our language partitions to mitigate effects of training data ordering.

We further acknowledge specific technical limitations associated with the methods used for our component-level analysis. For logit lens, teacher forcing introduces a length- and tokenization-dependent bias: multi-token words are scored via multiple conditioned steps, even with length-normalization. In addition, absolute logit-lens scores are affected by script-dependent biases in the output embedding. Another known limitation of the logit lens is that intermediate residual-stream states are not necessarily calibrated to be compatible with the final unembedding. %
We also note that while ExPLAIND offers novel perspectives from which to study model behavior, its explanations do not inform about the causal effect that changes in the model or data would have on the model behavior. We address some of these objections by performing different model interventions with layer swapping and activation scaling over time. While they support our hypotheses, we do not perform them exhaustively over all possible settings due to their computational cost. This might occlude other interesting dynamics.

In our layer-swapping experiments, we group layers into bottom, intermediate, and top blocks to parallel existing literature \cite{DBLP:conf/nips/0006ZCKB24}, despite our logit lens analysis suggesting them to be less distinct than in previous work. We hypothesize that this apparent two-stage process is inherent to translation, which simply maps representations into and out of a shared space, whereas tasks demanding complex cross-lingual reasoning (e.g., math problems) might exhibit more pronounced intermediate processing. Indeed, our layer-swapping experiments suggest that the bottom and intermediate blocks develop concurrently rather than sequentially. Furthermore, we note that no single correct boundary exists, as optimal splits vary with the task at hand \cite{bandarkar-peng-2025-unreasonable}.
In addition, while we argue for the emergence of shared representational spaces, our methodology does not directly examine the underlying activations (cf. \citealp{dumas-etal-2025-separating}). However, our layer-swapping experiments still provide convincing evidence for their existence.

\section*{Ethical Considerations}
Our work focuses on interpretability, which contributes to the broader goal of algorithmic transparency. However, there is an inherent ethical risk of over-reliance on such insights. Despite these risks, the goal of improving cross-lingual transfer is a significant positive consequence. Through a better understanding of cross-lingual transfer, our work can inform future work on the inclusion of low-resource languages in global AI development and help bridge the digital divide.

We disclose that AI tools were used during this research for support with code generation, ideation, writing and experimental suggestions. To ensure the integrity of the work, all AI-generated outputs were rigorously verified by the authors, and we confirm that we take full responsibility for the final content and results presented.

\section*{Acknowledgments}
FK and BP are supported by the ERC Consolidator Grant DIALECT 101043235. MM and GK are supported by the DAAD program Konrad Zuse Schools of Excellence in Artificial Intelligence, sponsored by the German Federal Ministry of Research, Technology and Space.
This work was partly completed at the EuroCC AI Hackathon, part of the Open Hackathons program. We are grateful to our hackathon mentors Benjamin Geißler and Severine Habert. We acknowledge computational resources provided by the Leibniz Supercomputing Centre (LRZ). We are also grateful to Ryan Soh-Eun Shim for fruitful discussions throughout the project, as well as Domenico De Cristofaro and Xinpeng Wang for their feedback on previous drafts. Finally, we thank FAST LTA for providing additional storage, and Thomas Schäfer and Karl Ischebeck for technical support. 

\section*{Contribution Statement}
FE conceived the research idea, which was further developed with FK and MM. All three performed experiments and analyzed results. MM led training code development; all three developed code for analysis. FK curated data for evaluation, with the exception of data used for ExPLAIND, curated by FE. Paper writing was led by FK, with MM and FE contributing, and MH and BP assisting in improving the draft. GK, MH, and BP supervised the project and reviewed the paper.

\bibliography{custom_new}
\appendix
\crefalias{section}{appendix}
\crefalias{subsection}{appendix}
\clearpage

\section{Additional Details on Pretraining}
\label{app:details_pretraining}
The \textbf{model} is a decoder-only transformer \cite{vaswaniAttentionAllYou2017} following the LLaMA architecture \cite{touvron2023llamaopenefficientfoundation}, with 24 decoder layers, a hidden size of 2048, and standard LLaMA components such as multi-head attention, RMSNorm and rotary positional embeddings (RoPE) with a high base. We provide an overview of all parameters of the model in \cref{tab:config_model}. 

We use the Tekken (Mistral–Nemo) \textbf{tokenizer} \cite{mistral_nemo_2024}, a state-of-the-art multilingual tokenizer \cite{apertus2025apertusdemocratizingopencompliant} that provides broad script coverage.

We train the model using the widely adopted AdamW \textbf{optimizer }\cite{loshchilov2018decoupled} with a peak learning rate of $1\times 10^{-4}$. The learning rate schedule follows a \textit{wsd-decay} scheme \cite{DBLP:journals/corr/abs-2404-06395}, which first increases the learning rate linearly, plateaus for most of the training, and then decreases proportionally to $1/\sqrt{t}$. We apply gradient clipping at~1.0 for stability under mixed-precision training. These settings follow established practice \cite{martins2025eurollm9btechnicalreport}. We report the exact parameters in \cref{tab:configuration_optimizer}.

We draw our pretraining data from the FineWeb2-HQ and FineWeb-HQ \textbf{datasets} \cite{messmer2025enhancing_fineweb}. The mixture covers nine languages spanning multiple scripts (Latin, Han, and Japanese). Sampling probabilities are listed in \cref{tab:languages_overview}, with English accounting for $50\%$ of all samples and the remaining eight languages drawn uniformly. We do not exhaust any of the monolingual datasets during training and every data point is seen at most once.

We \textbf{implement} the training in \texttt{Python} using \texttt{PyTorch}. The model implementation uses the \texttt{LlamaForCausalLM} module from the HuggingFace \texttt{transformers} library.\footnote{\url{https://huggingface.co/docs/transformers/main/model_doc/llama}} Distributed training across eight NVIDIA A100 GPUs is handled via the \texttt{accelerate} library from HuggingFace. In total, the run processes approximately $37$B tokens and converges to a final training loss of~$2.58$, consistent with expectations for models of comparable scale and architecture \cite{allal2025_the_smol_training_playbook_the_secrets_to_building_world_class_llms}. Pretraining was completed in $281.4$ hours.

\begin{table}[!t]
\centering
\caption{Model and training configuration.}
\label{tab:full_config}
\begin{minipage}[t]{\linewidth}%
\vspace{0pt}
\centering
\subcaption{\textbf{Languages used}, together with their
ISO2 code, script and sampling probability $p$ during
training.}
\begin{tabular}{llll}
\toprule
\textbf{Language} & \textbf{ISO2} & \textbf{Script} & $p$ \\
\midrule
English          & en & Latin    & $0.50$   \\
Mandarin Chinese & zh & Han      & $0.0625$ \\
German           & de & Latin    & $0.0625$ \\
Spanish          & es & Latin    & $0.0625$ \\
Japanese         & ja & Japanese & $0.0625$ \\
French           & fr & Latin    & $0.0625$ \\
Italian          & it & Latin    & $0.0625$ \\
Portuguese       & pt & Latin    & $0.0625$ \\
Indonesian       & id & Latin    & $0.0625$ \\
\bottomrule
\end{tabular}
\label{tab:languages_overview}
\vspace{1em}
\subcaption{\textbf{Optimizer configuration}}
\begin{tabular}{ll}
\toprule
Optimizer      & AdamW \\
Learning rate  & $1 \times 10^{-4}$ \\
Betas          & $(0.9, 0.999)$ \\
Epsilon        & $1 \times 10^{-8}$ \\
Weight decay   & $0.01$ \\
Warmup ratio   & $0.10$ \\
Scheduler      & warmup-stable-decay \\
Min.\ LR ratio & $0.10$ \\
\bottomrule
\end{tabular}
\label{tab:configuration_optimizer}
\end{minipage}
\begin{minipage}[t]{\linewidth}%
\vspace{1em}

\centering
\subcaption{\textbf{Model configuration}}
\begin{tabular}{ll}
\toprule
Architecture       & Llama-like \\
Hidden size        & $2048$ \\
Intermediate size  & $5632$ \\
Layers             & $24$ \\
Attention heads    & $16$ \\
KV heads           & $8$ \\
Max sequence length & $4096$ \\
RoPE $\theta$      & $500{,}000$ \\
RoPE scaling       & dynamic (factor $1.0$) \\
Tokenizer          & Mistral-Nemo-Base-2407 \\
Precision          & bfloat16 \\
Attention          & FlashAttention2 \\
Collate style      & Packing \\
Parameters         & $1{,}669{,}433{,}344$ \\
\bottomrule
\end{tabular}\label{tab:config_model}
\vspace{1em}
\subcaption{\textbf{Training run details}}
\begin{tabular}{ll}
\toprule
Total steps      & $100{,}000$ \\
Effective batch size & $512$ \\
Gradient accumulation & $32$ \\
Gradient clipping & $1.0$ \\
Precision        & bfloat16 \\
\bottomrule
\end{tabular}
\label{tab:configuration_run}
\end{minipage}
\end{table}

\begin{figure}[t]
    \centering
    \includegraphics[width=\linewidth]{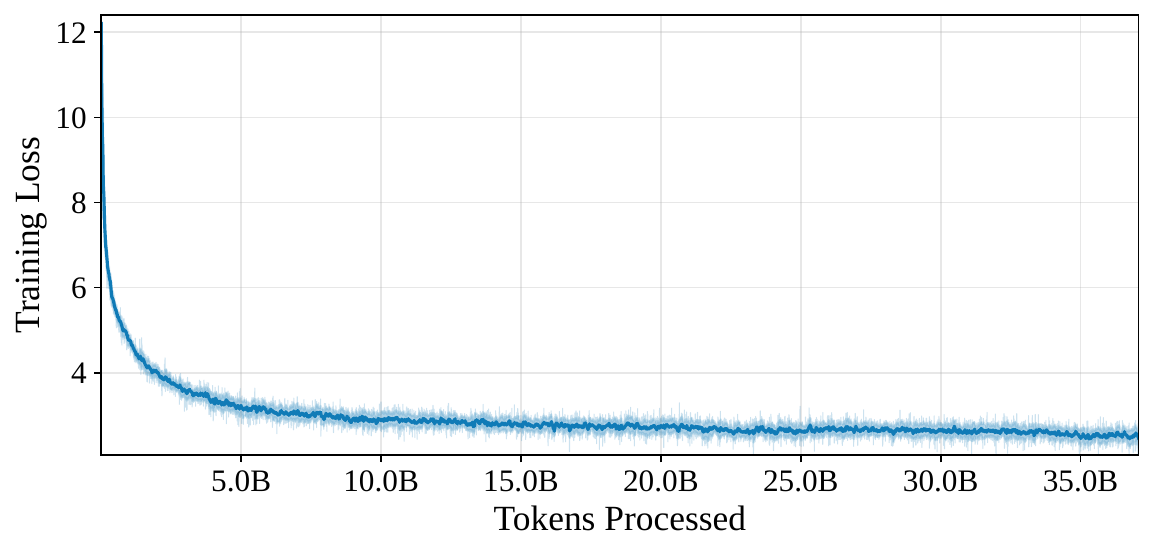}
    \caption{We report the (smoothed) loss-curve of the training. The model consumes around $37$B tokens, reaching a final loss of around $2.58$. The learning rate decay at the end of training (last $10\%$) is visible by a slight drop in the loss curve.}
    \label{fig:loss_curve}
\end{figure}

\begin{figure}[t]
    \centering
    \includegraphics[width=\linewidth]{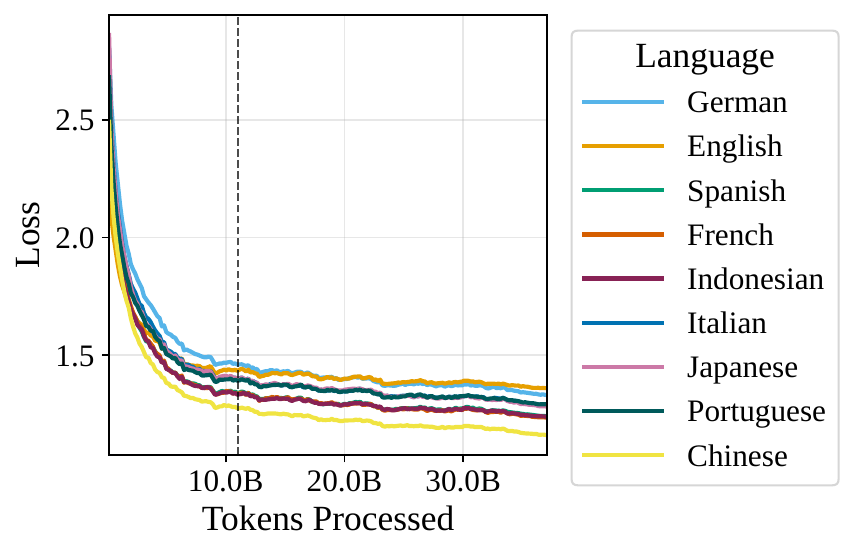}
    \caption{We report loss on a fixed set of 512 sequences per language, sampled from unseen (i.e., unconsumed) partitions of our training data. This breakdown indicates consistent progress across all languages.}
    \label{fig:loss_curve_multi}
\end{figure}

We summarize the model architecture, optimizer settings, and training configuration in \cref{tab:full_config}, as well as the loss curve in \cref{fig:loss_curve} and \cref{fig:loss_curve_multi}. 


\section{Additional Details on BLiMP}\label{app:blimp-phenomena}
MultiBLiMP \cite{jumelet2025multiblimp10massivelymultilingual}, used for the English, French, Spanish, Portuguese, Italian, and German, focuses on subject-verb-agreement, a local morphosyntactic phenomenon. As MultiBLiMP does not cover Japanese and Chinese, we turn to JBLiMP \cite{someya-oseki-2023-jblimp} and ZhoBLiMP \cite{DBLP:journals/corr/abs-2411-06096}, respectively. However, these benchmarks cover a wide range of phenomena, including ones that test long-range dependencies. By contrast, MultiBLiMP is fairly simple: the MultiBLiMP authors report that relatively small Goldfish models (125M parameters, \citealp{chang2026goldfishmonolinguallanguagemodels}) achieve a mean score of 95.8\% across European languages. Therefore, we subsample both ZhoBLiMP and JBLiMP phenomena.

For ZhoBLiMP, we restrict to phenomena classified by the benchmark authors as ``easy'': any phenomenon for which their Pythia models achieve above 85\% accuracy. This leaves: verb phrase, topicalization, control/raising, question, BA construction, relativization, and 55 fine-grained phenomena under these categories, for which we subsample 30 examples each, to roughly match the MultiBLiMP partitions in size.

For JBLiMP, we exclude phenomena requiring long-range dependencies and found by the benchmark authors to be particularly challenging for language models \cite{someya-oseki-2023-jblimp}, achieving scores as low as 17\% in their evaluation (chance is 50\% for BLiMP-style pairs). This leaves: argument structure, morphology, nominal structure, and ellipsis, for which models evaluated by the benchmark authors achieve at least 85\% accuracy. This leaves us with 219 samples, notably smaller than the other benchmarks.

While LINDSEA \cite{DBLP:journals/corr/abs-2309-06085} includes linguistic minimal pairs for Indonesian, these are not designed for log-probability scoring; the authors note that some pairs require judgment of semantic plausibility rather than grammaticality to select the correct sentence. Furthermore, the benchmark is not validated for log-probability scoring: The authors evaluate using prompted LLMs rather than log-probabilities. Therefore, we choose not to include it. Unfortunately, to our knowledge no other alternative exists for validating linguistic capability in Indonesian in a comparable manner. As an approximation of basic linguistic capability, including Indonesian, we report per-language loss on unseen data in Appendix \cref{fig:loss_curve_multi}.

We use the LM Evaluation Harness \cite{eval-harness} to compute accuracy based on length-normalized log-probability scoring.
\section{Word-Level Translation Dataset Construction}\label{app:details_wlt}
\begin{figure*}[p]
    \centering
    \begin{subfigure}[t]{.49\linewidth}
        \includegraphics[width=\linewidth]{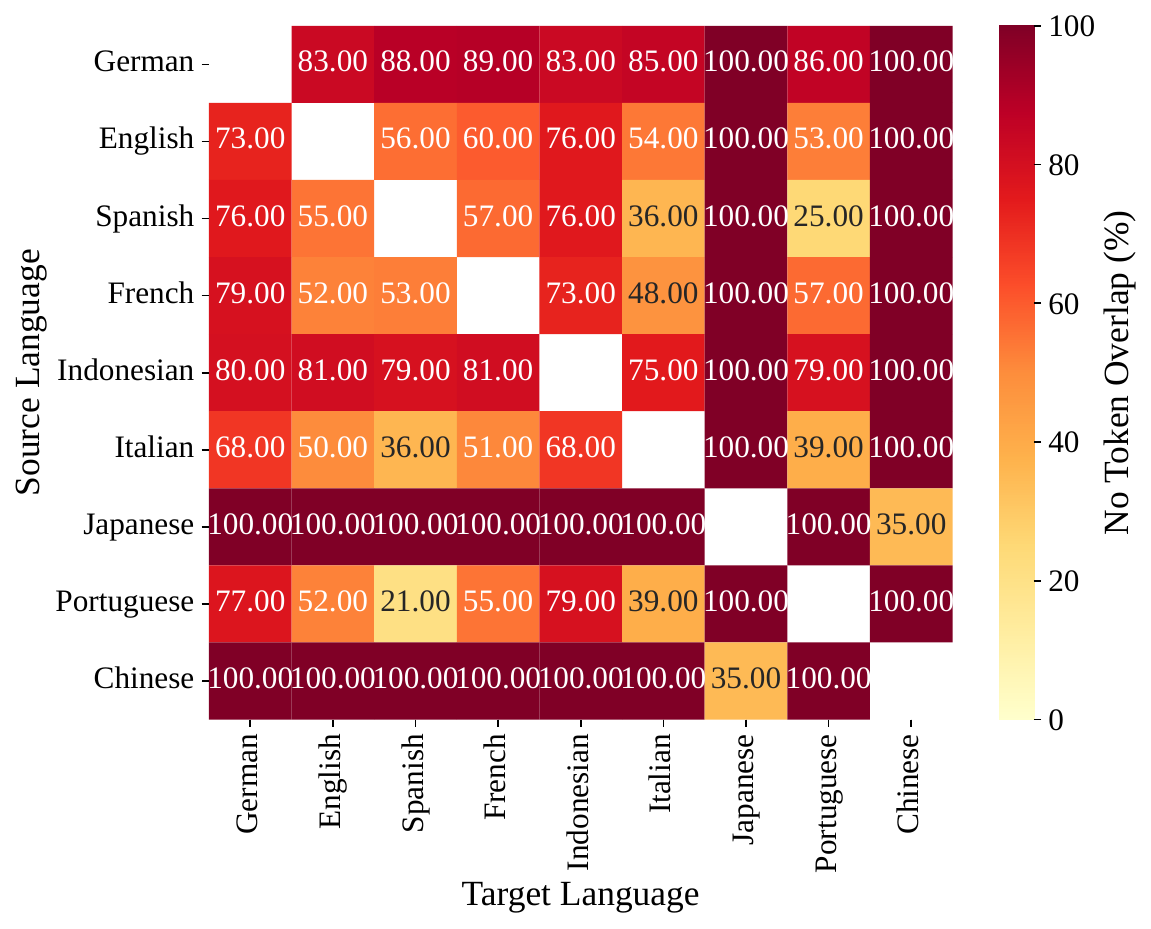} 
        \caption{No token overlap.}
        \label{fig:app_copy_stat_no}
    \end{subfigure}\hfill
        \begin{subfigure}[t]{.49\linewidth}
        \includegraphics[width=\linewidth]{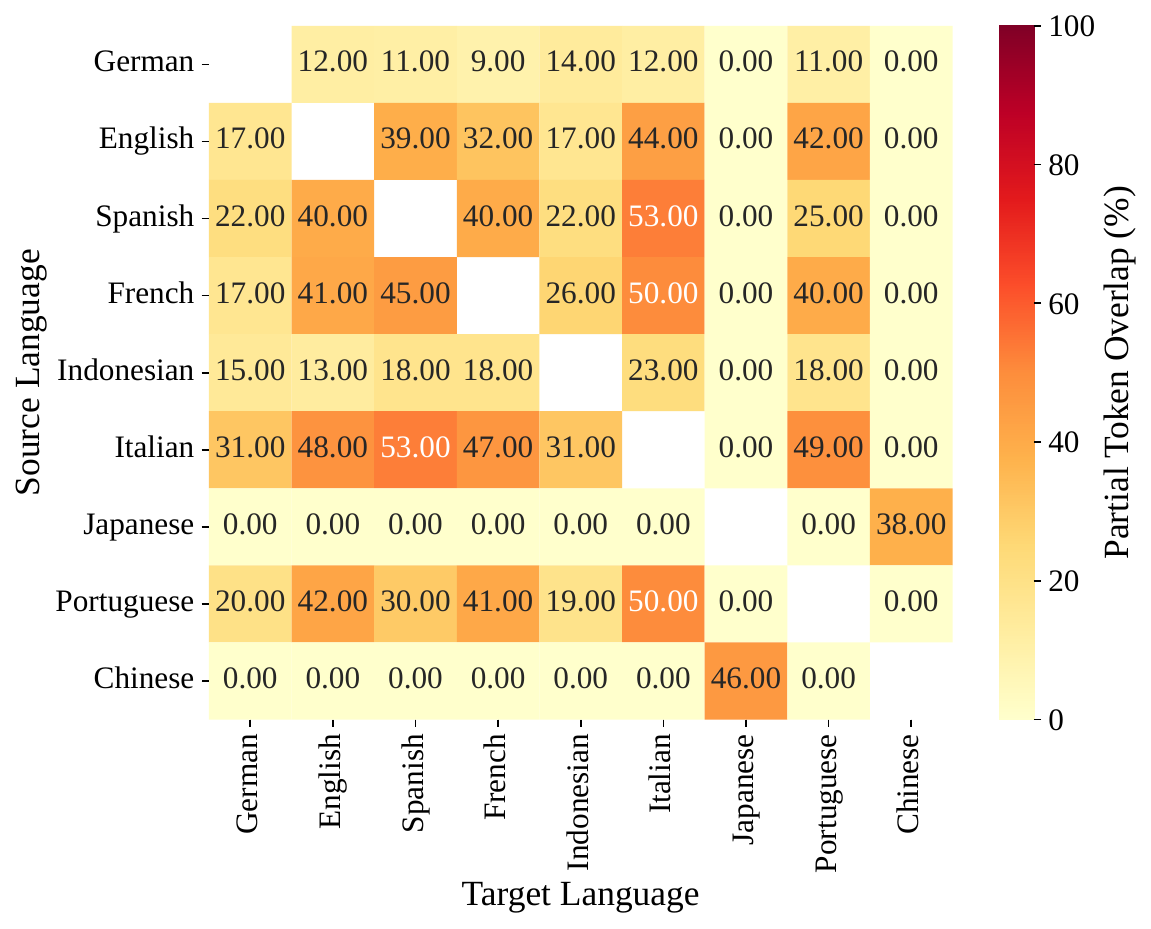}
        \caption{Partial token overlap. }
        \label{fig:app_copy_stat_partial}
    \end{subfigure}
      \begin{subfigure}[t]{.49\linewidth}
        \includegraphics[width=\linewidth]{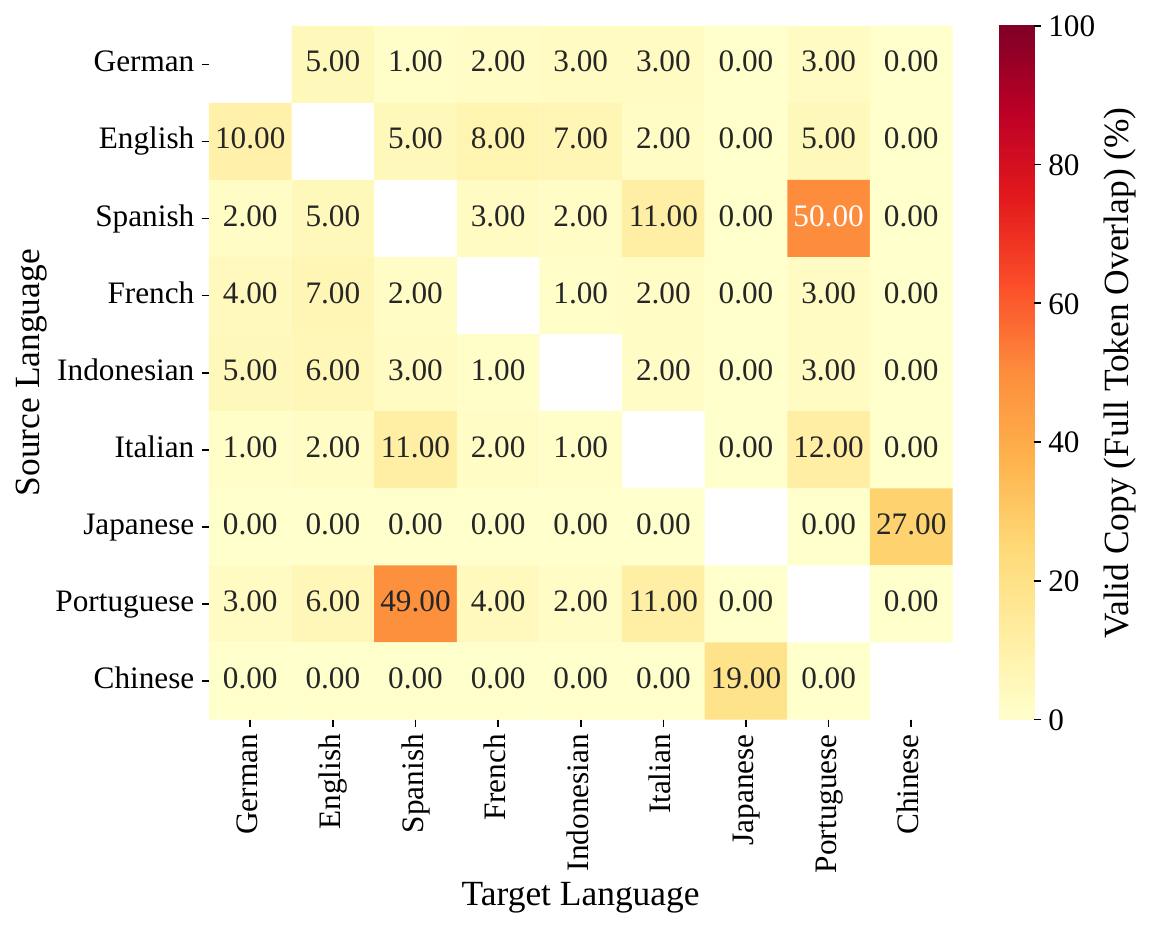} 
        \caption{Valid copy (full token overlap). }
        \label{fig:app_copy_stat_fiull}
    \end{subfigure}
    
    \caption{For each language pair, we report the fraction of words per token overlap bucket in our \wlt data. For example, all word pairs ($100\%$) between Chinese/Japanese and Latin-script languages fall into the "no overlap" bin. Even within Latin-script languages, the test data exhibits heterogeneous levels of token overlap. Due to the varying sets of valid translations, this breakdown is not symmetric. We count a concept as having partial token overlap when at least one word in the set of valid translations has such partial overlap. We further count the partial-overlap words by excluding valid copies; in other words, the token overlap bins are disjoint.}
    \label{fig_app:copy_stat}
\end{figure*}

Here we provide additional details on the construction of the word-level translation (\wlt) dataset, which comprises 100 query concepts, expressed in nine languages with valid translations for all 72 directed language pairs accounting for polysemy. Furthermore, we also construct 25 \emph{few-shot concepts} with translations in nine languages, which do not account for polysemy (i.e., we do not collect synonyms for these). In this study, we focus on natural language use. To this end, we allow for inflection and words that belong to multiple word classes, which may otherwise be excluded from \wlt tasks to avoid ambiguity in the target translation. When stratifying over word classes, we use a simple heuristic to assign a primary PoS tag, first querying WordNet\footnote{https://wordnet.princeton.edu/ via NLTK, \citealp{Bird2006NLTKTN}} for each word's syntactic categories. A word was considered a member of a PoS class if WordNet contained at least one synset for that word under the corresponding category (noun, verb, adjective, or adverb; adjective satellites were merged with adjectives). When a word belonged to multiple categories, we applied a priority ordering of noun > verb > adjective > adverb. An additional rule handled gerunds (ending in ``-ing''): if the stem had verbal synsets, these were assigned the verb class. We override the heuristic only once, for ``complex'', where it assigned ``adjective'', an unintuitive label. We show the final distribution over PoS tag sets in \cref{tab:pos}, and curate a set of valid translations to reflect this PoS tag ambiguity. We further report the token overlap in our test data, broken down per language pair in \cref{fig_app:copy_stat} which confirms that language pairs have heterogeneous levels of token overlap, even within same-script languages.

As described in \cref{sec:data-wlt}, constructing a set of valid translations for \wlt requires careful consideration. We source our concepts from ChiKhaPo \cite{Chang2025ChiKhaPoAL}, but find that the synsets are in some cases too broad for our purposes, including words that would only be considered a valid translation in very specific contexts. In other cases, they are not broad enough, for example not including gendered variants. Therefore, we come up with a multi-source voting procedure to filter and introduce candidates. For each source word, language-pair specific candidate translations were collected from four evidence streams:
\begin{itemize}
    \item Machine translation outputs given the source word from multiple MT systems: \texttt{Aya-Expanse-32B} \cite{dang2024ayaexpansecombiningresearch}, \texttt{translategemma-27B-it} \cite{gemmatranslate2026}, madlad400-10b-mt \cite{kudugunta2023madlad400}, mbart-large-50-many-to-many-mmt \cite{DBLP:journals/corr/abs-2008-00401}, and \texttt{nllb-200-3.3B} \cite{nllbteam2022languageleftbehindscaling}.
    \item Lexicon entries: ChiKhaPo for non-English targets and Merriam-Webster\footnote{https://www.merriam-webster.com/} synonyms and thesaurus entries for English targets.
    \item DeepL\footnote{https://www.deepl.com/} translations of the source word, and DeepL sense-based translations of the English word (using Merriam-Webster entries), filtered by back-translation match.
    \item Back-translation verification, where candidates translating back to the source word received an additional vote.
\end{itemize}

Candidates were required to be supported by at least one MT system; lexicon-only candidates (ChiKhaPo, Merriam-Webster) contributed fractional votes but could not admit a candidate alone. The candidate sets were subsequently filtered using an LLM judge (\texttt{claude-sonnet-4-5-20250929}). For each source word and every target language (72 language pairs total), the model was prompted to discard candidates with mismatching part of speech, number, or meaning, while retaining morphological variants (e.g.\ gendered forms) when the source word is under-specified. We show the system prompt in \cref{fig:filter-prompt}, and refer readers to our research code for full details. We verify that the LLM does not introduce new words, mitigating the risk of hallucinations. Finally, we manually check the resulting candidates.

\begin{table}[t]
\centering
\caption{Word-level translation dataset distribution by English primary PoS and PoS tag set.}
\begin{tabular}{lrr}
\toprule
primary PoS & count & \textbf{\%} \\
\midrule
noun & 40 & 32\% \\
verb & 40 & 32\% \\
adj & 25 & 20\% \\
adv & 20 & 16\% \\
\midrule
\multicolumn{3}{l}{PoS tag sets} \\
\midrule
noun & 33 & 26\% \\
noun, verb & 22 & 18\% \\
adv & 19 & 15\% \\
verb & 18 & 14\% \\
adj & 12 & 10\% \\
adj, verb & 9 & 7\% \\
adj, noun, verb & 6 & 5\% \\
adj, noun & 5 & 4\% \\
adj, adv, noun & 1 & 1\% \\
\bottomrule
\end{tabular}
\label{tab:pos}
\end{table}

\begin{figure*}
\small
\begin{Verbatim}
<examples>
<example>
<source_word>schnell (de → it)</source_word>
<candidates>veloce, trovatore, velocemente, velocità, schnell</candidates>
<reasoning>
- veloce      ✓ (fast - adjective)
- trovatore   ✗ (troubadour - wrong)
- velocemente ✓ (quickly - adverb valid)
- velocità    ✗ (speed - noun, schnell not noun)
- schnell     ✗ (source copy)
</reasoning>
<output>veloce, velocemente</output>
</example>
<example>
<source_word>imprimir (pt → en)</source_word>
<candidates>print, printer, printing, printout</candidates>
<reasoning>
- print    ✓ (verb)
- printer  ✗ (machine - object noun, imprimir is verb)
- printing ✓ (verb form)
- printout ✗ (result noun, imprimir is verb)
</reasoning>
<output>print, printing</output>
</example>
\end{Verbatim}
\begin{CJK}{UTF8}{gbsn}
\begin{Verbatim}
<example>
<source_word>幸运 (zh → es)</source_word>
<candidates>afortunado, afortunada, suerte</candidates>
<reasoning>
- afortunado ✓ (lucky - masculine)
- afortunada ✓ (lucky - feminine, 幸运 ungendered)
- suerte     ✗ (luck - noun, 幸运 is adjective)
</reasoning>
<output>afortunado, afortunada</output>
</example>
\end{Verbatim}
\end{CJK}
\begin{Verbatim}
<example>
<source_word>matériaux (fr → en)</source_word>
<candidates>material, materials</candidates>
<reasoning>
- material  ✗ (singular, matériaux is clearly plural)
- materials ✓ (plural matches)
</reasoning>
<output>materials</output>
</example>
\end{Verbatim}
\begin{CJK}{UTF8}{min}
\begin{Verbatim}
<example>
<source_word>登る (ja → id)</source_word>
<candidates>mendaki, pendaki, pendakian</candidates>
<reasoning>
- mendaki   ✓ (to climb - verb)
- pendaki   ✗ (climber - agent noun, 登る is verb)
- pendakian ✗ (climb/climbing - result noun, 登る is verb)
</reasoning>
<output>mendaki</output>
</example>
</examples>
\end{Verbatim}
\end{CJK}
\begin{Verbatim}
**Keep:** Valid sense + matching word class + matching number and gender (unless ambiguous)
**Discard:** Wrong meaning, POS mismatch (agent/object/place nouns from verbs), number mismatch when
unambiguous, source copy

Format:
<reasoning>brief per candidate</reasoning>
<output>valid1, valid2</output>
\end{Verbatim}
\caption{System prompt for LLM-based candidate filtering (Claude Sonnet,
\texttt{claude-sonnet-4-5-20250929}, temperature~0).}
\label{fig:filter-prompt}
\end{figure*}

We acknowledge that this pipeline is an approximation of gold human judgment. We expect errors to be predominantly false negatives, i.e., valid translations excluded by conservative filtering, which would lead to slight underestimates of model accuracy rather than inflated scores. 

A risk when constructing a benchmark is that it is too difficult for a model to achieve decent scores. Therefore, we verify that the benchmark is tractable by evaluating EuroLLM $1.7$B \cite{martins2025eurollm9btechnicalreport}, a similarly sized multilingual model. EuroLLM $1.7$B achieves an accuracy of $61.1\%$, verifying that the benchmark is tractable. While this is significantly higher than our final model's performance, this comparison is limited due to the significant difference in pretraining data size: EuroLLM $1.7$B is trained on $4$T tokens, compared to our $37$B.

\section{Additional Details on Word-Level Translation Evaluation}\label{app:wlt-eval}

\subsection{Experimental Setup}\label{app:wlt-exp}
We greedily decode model outputs, allowing up to 16 tokens (corresponding to the longest valid translation in our \wlt dataset). If a '' is found, we split on this, otherwise, we use the full decoded output. A word is counted as correctly translated if its lowercased form matches one of the lowercased valid translations. For the repetition task, a word is counted as correctly repeated if its lowercased form matches the lowercased source word.

We count an output as \emph{source word copying} if the lowercased generated word matches the lowercased source word. Source word copying can be valid, as is the case for the word ``ala'' (wing in both Spanish and Italian). %
We count an output as \emph{context copying}, when it matches the target word of any of the few-shot examples in the prompt, modulo casing. 

\subsection{Evaluation}
\label{app:training_metrics}

We complement the metrics reported in \cref{fig:copying_error_mode,fig:combined_metrics}, with additional statistics about \wlt performance, alongside further noteworthy observations from our analysis.

\begin{figure}[t]%
\centering 
\begin{subfigure}[t]{\linewidth} 
\includegraphics[width=\linewidth]{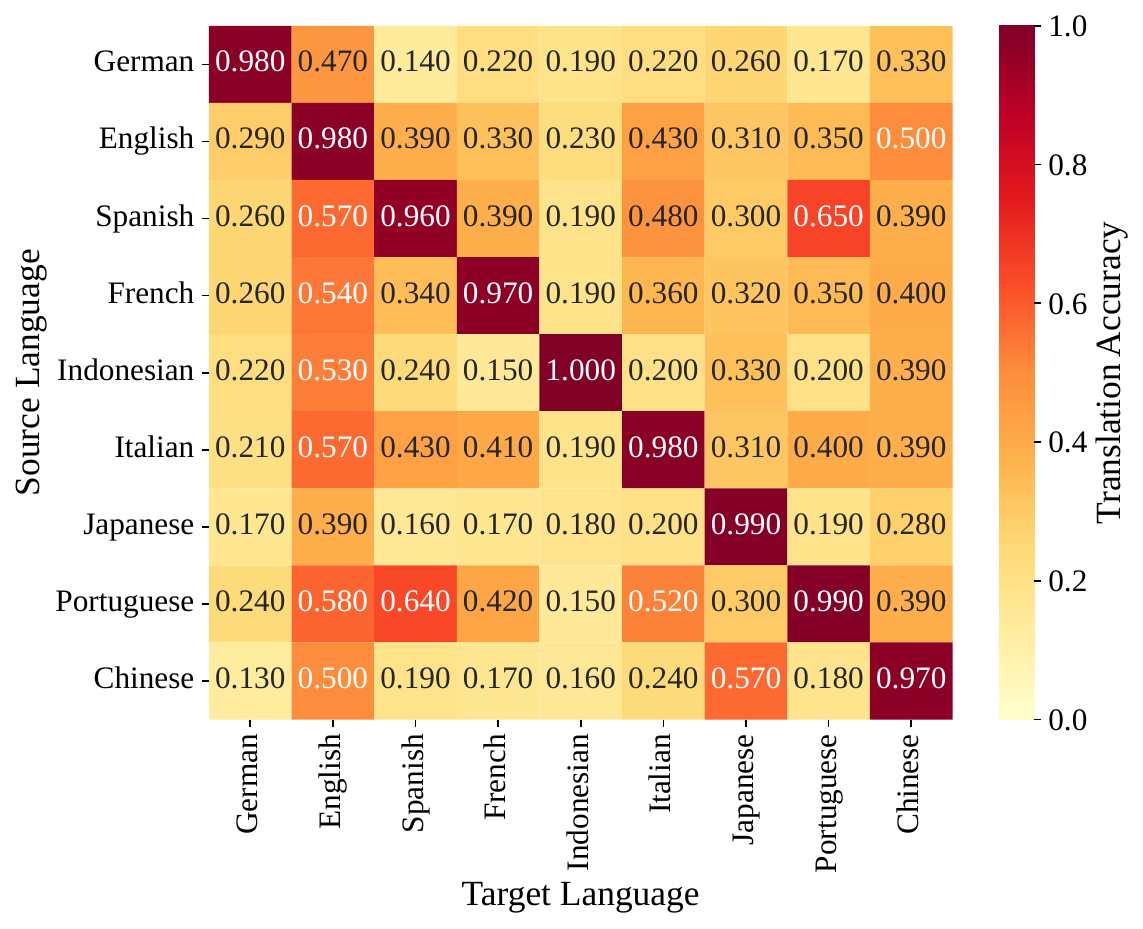}
\caption{For all language pairs at the last checkpoint, we report the translation and repetition accuracies on the \wlt dataset. The overall average is $32.1\%$. \cref{fig:translation_accuracy_from} shows the accuracy evolution over time.}
\label{fig:final_translation_accuracies}
\end{subfigure}

\begin{subfigure}[t]{\linewidth}
\includegraphics[width=\linewidth]{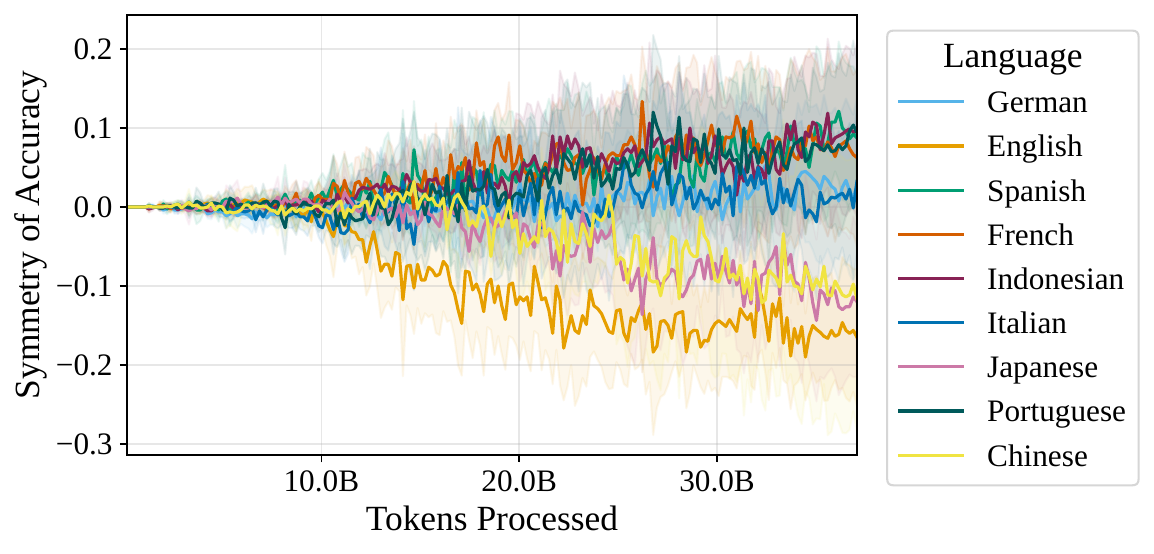}
\caption{Signed mean accuracy difference for \wlt throughout training. Negative values indicate higher accuracy when translating \emph{into} a given language. We observe that asymmetry develops around the start of Phase II, and strengthens throughout it.}
\label{fig:symmetry}
\end{subfigure}
\caption{\textbf{Analysis of directional asymmetry in word-level translation.} We observe that translation performance is directional, favoring English, and, to a lesser degree, Chinese and Japanese as target languages throughout the developmental trajectory of pretraining.}
\end{figure}

\begin{figure}[t]
\includegraphics[width=\linewidth]{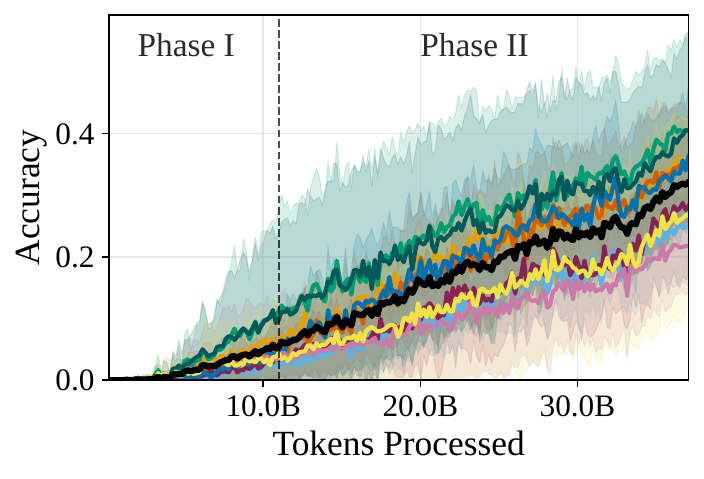}
\caption{\wlt accuracy throughout training, grouped by source language.}
\label{fig:translation_accuracy_from}
\end{figure}
\paragraph{Translation accuracy}
\cref{fig:final_translation_accuracies} presents the final translation accuracy for each directed language pair. Surprisingly, accuracy is not monotonic. Out of $7,200$ word pairs, $57.8\%$ are predicted correctly at \emph{some} point in training, but are often not consistently correct after this point. We further observe a pronounced directional asymmetry in performance; for instance, the accuracy of translating \emph{to} English is significantly higher than translating \emph{from} English, as evidenced by the distinct dark column in the heatmap. We hypothesize that this is due to English's role as the dominant language during pretraining. To a lesser extent, accuracy into Japanese and Chinese also exceeds the accuracy of translating from those languages, a phenomenon potentially reflecting higher language extractability afforded by their unique scripts. As shown in \cref{fig:symmetry}, this directional asymmetry emerges early, coinciding with the start of Phase II, and continues to develop throughout the remainder of pretraining. Notably, the asymmetry for English reaches a plateau mid-way through this phase, indicating that reciprocal directions are being learned at a similar rate, while the asymmetry for other language pairs continues to widen.

\begin{figure*}[p]
    \centering
    \begin{subfigure}[t]{.49\linewidth}
        \includegraphics[width=\linewidth]{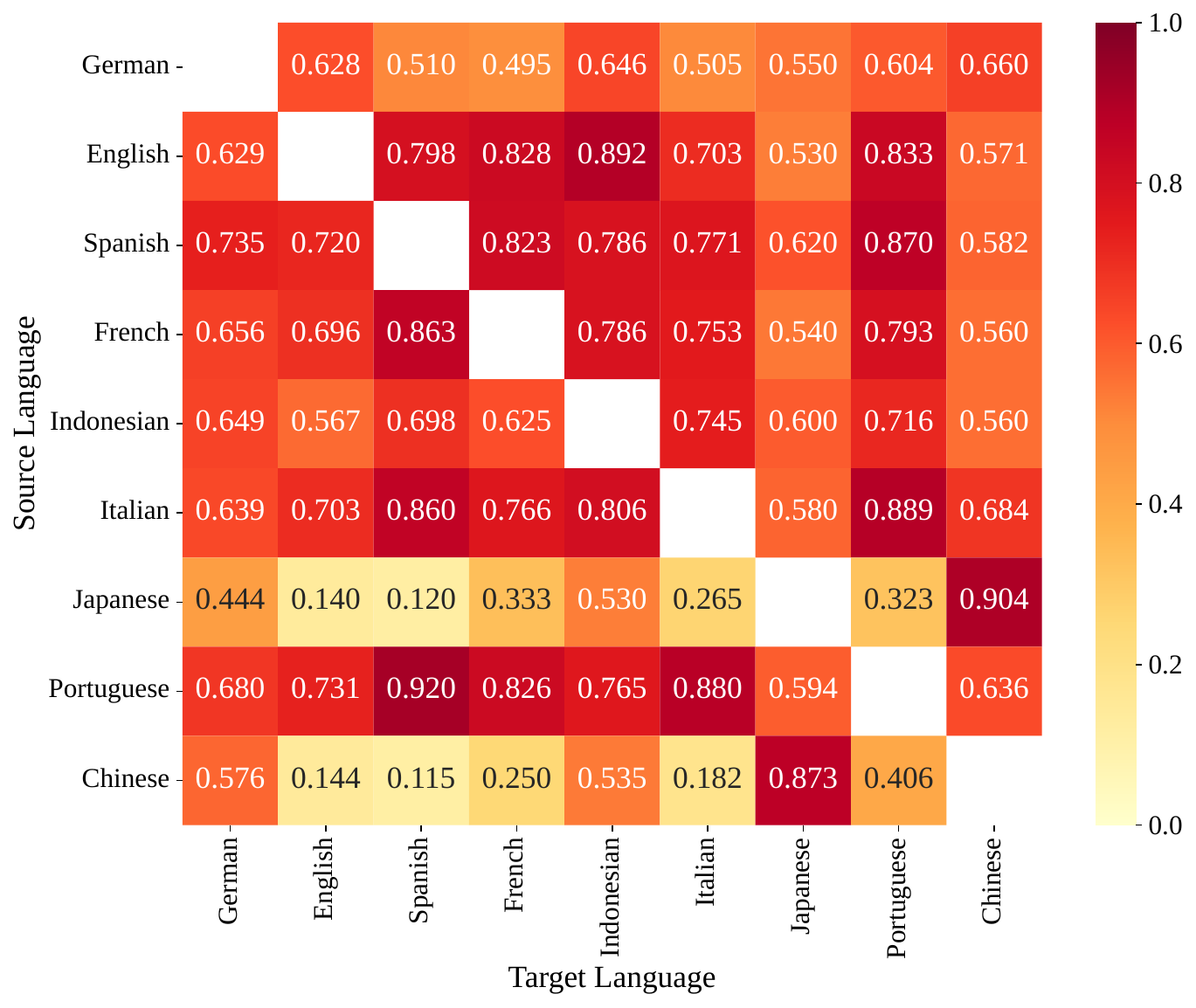} 
        \caption{Maximum during training. }
        \label{fig:app_errorfraction_copying}
    \end{subfigure}\hfill
        \begin{subfigure}[t]{.49\linewidth}
        \includegraphics[width=\linewidth]{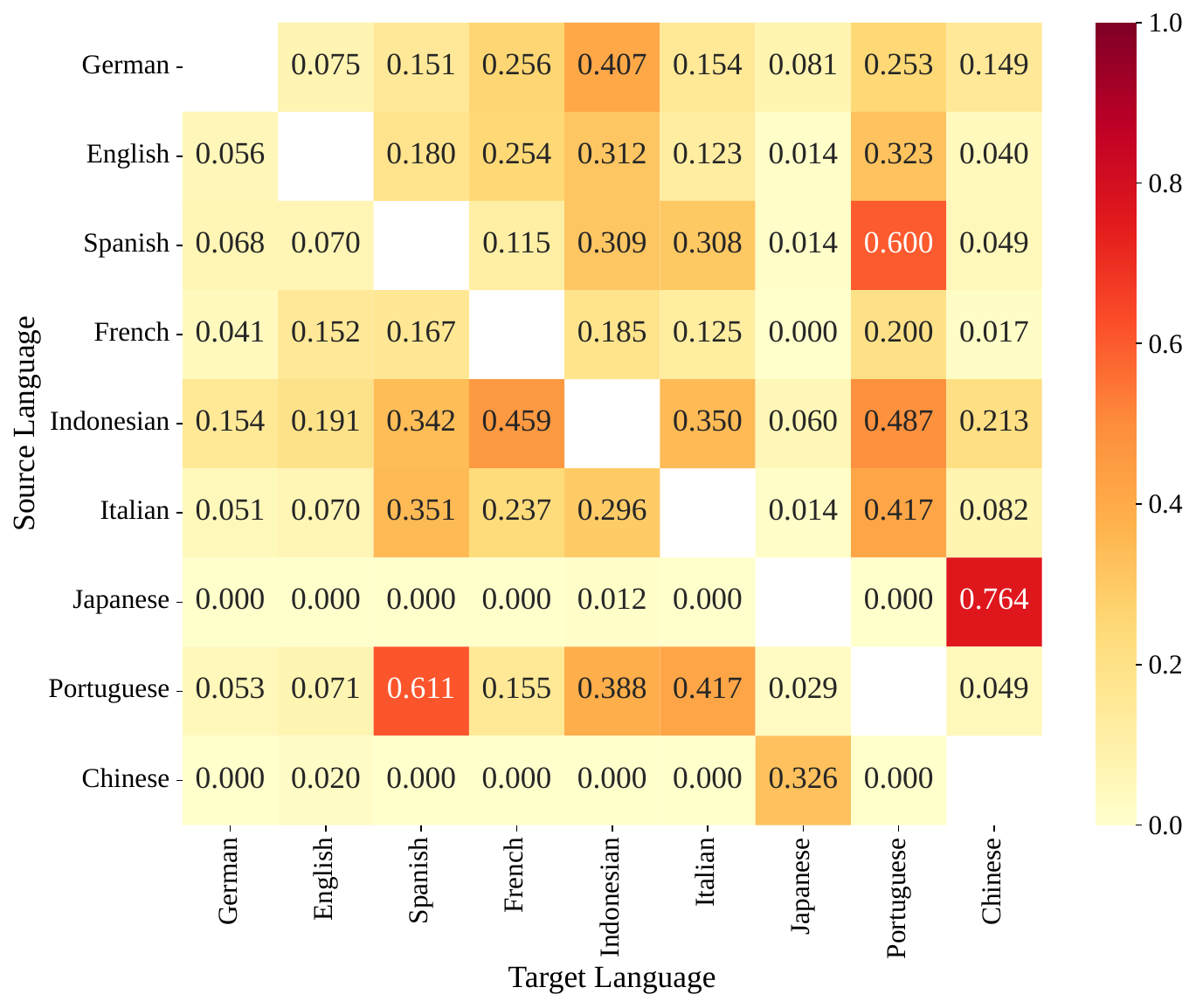} 
        \caption{Final checkpoint. }
        \label{fig:app_errorfraction_copying_final}
    \end{subfigure}
    
    \caption{We report the fraction of incorrect translations attributable to source word copying, measured at the checkpoint where this error fraction is maximal for each language pair (left) and the final checkpoint.}
    \label{fig_app:error_fraction}
\end{figure*}

\begin{figure*}[p]
    \centering
    \begin{subfigure}[t]{.49\linewidth}
    \centering
    \includegraphics[width=\linewidth]{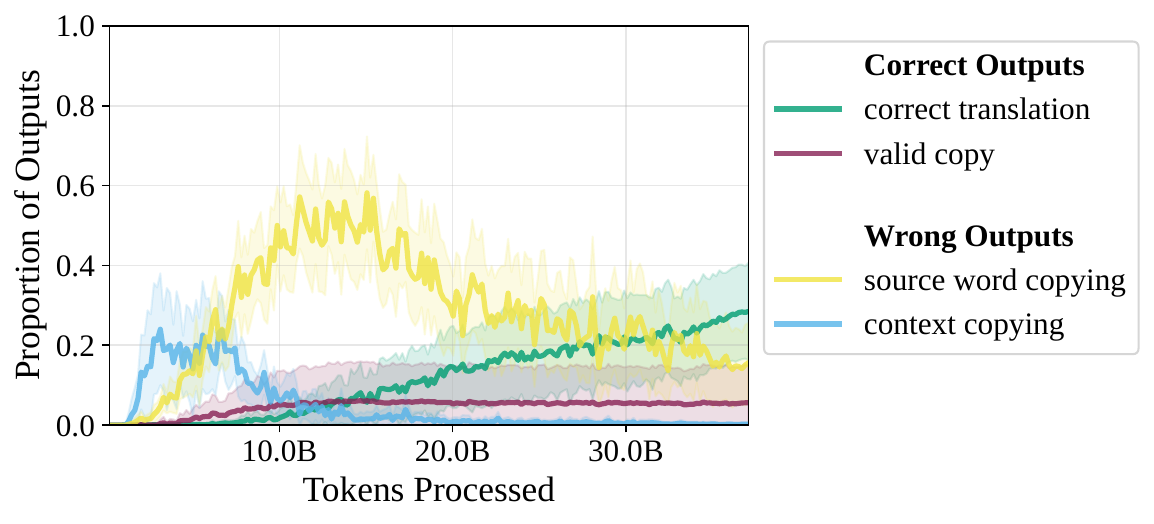}     
    \caption{Word-level translation, for the seven Latin-script languages. }
    \end{subfigure}
   \begin{subfigure}[t]{.49\linewidth}
   \centering
       \includegraphics[width=\linewidth]{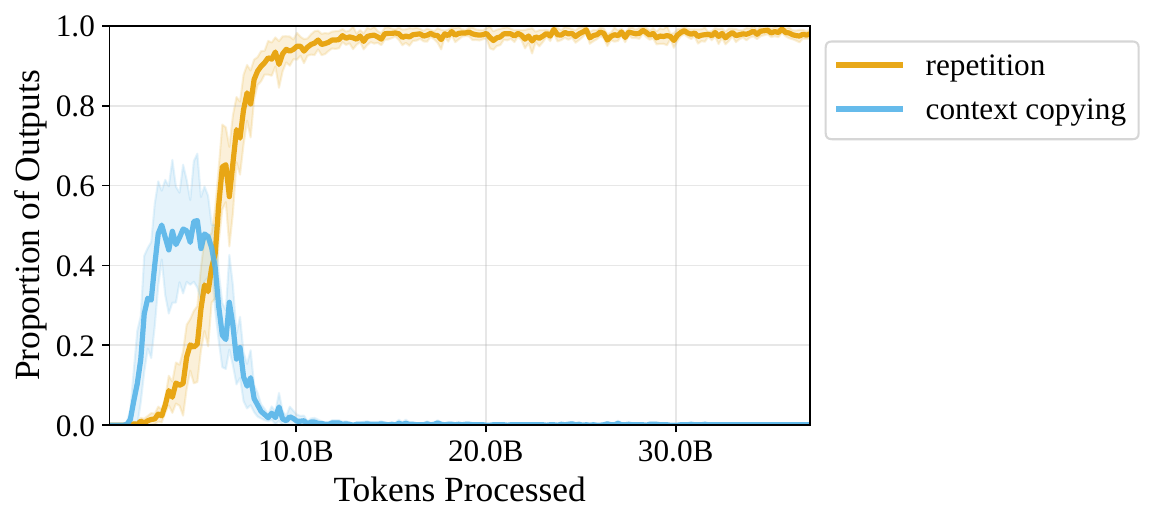}
       \caption{Repetition task.}
   \end{subfigure}
    \caption{\textbf{Decomposition of outputs onto copying types.} Context copying is more prevalent for repetition than for \wlt. Note that the left plot is restricted to the Latin-script languages, as source word copying is less frequent between scripts (see \cref{fig_app:error_fraction}). Error bars show standard deviation of language averages.}
    \label{fig:app_copy_rates_full}
\end{figure*}

\paragraph{Copying as an error mode} \cref{fig:app_copy_rates_full} shows the decomposition of model outputs into specific error modes throughout training. Context copying is a frequent error also for the repetition task, exceeding the respective rate on the word-level-translation task, but decaying faster. 
We further observe that copying behavior is highly language-dependent. The fraction of incorrect translations attributable to source word copying varies significantly across pairs (\cref{fig_app:error_fraction}). This suggests that a script difference discourages source word copying. At the end of training, this error mode is especially present for language pairs with the highest frequency of valid copies in the \wlt data, namely Spanish$\leftrightarrow$Portuguese ($50\%/49\%$), as well as Japanese$\leftrightarrow$Chinese ($27\%/20\%$).

\begin{figure*}[p]
\centering
   \begin{subfigure}[t]{.33\linewidth}
       \centering
   \includegraphics[trim = 0 0 0 0, clip, height=4.2cm]{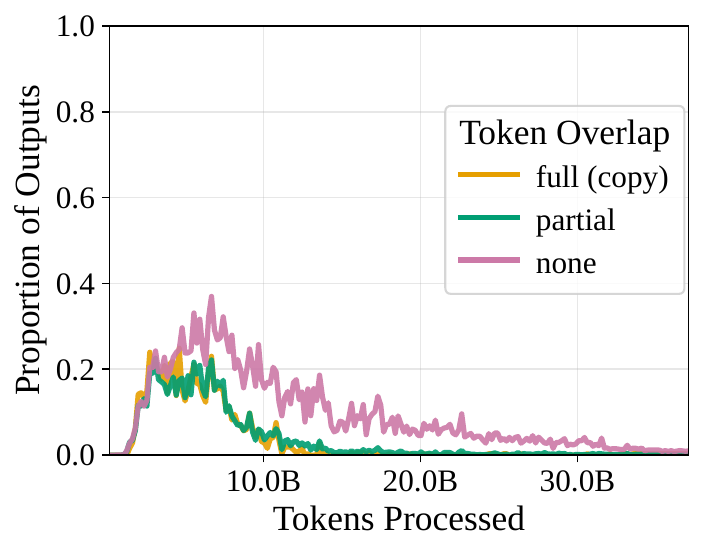}
    \caption{Context copying.}
    \label{fig:app_contextcopying_per_overlap_bin}
      \end{subfigure}
       \begin{subfigure}[t]{.31\linewidth}
       \centering
   \includegraphics[trim= .8cm 0 0 0, clip, height=4.2cm]{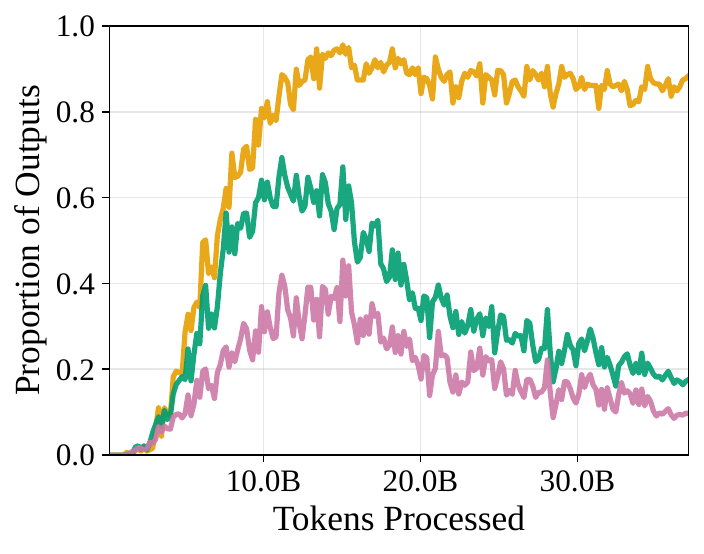}
    \caption{Source word copying.}
    \label{fig:app_sourceword_per_overlap_bin}
      \end{subfigure}
       \begin{subfigure}[t]{.32\linewidth}
       \centering
   \includegraphics[trim= 0 0 0 0, clip, height=4.2cm]{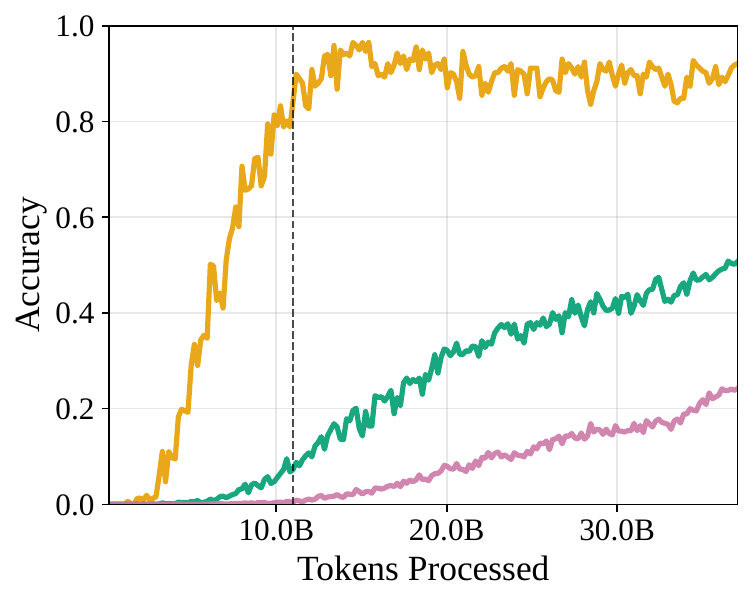}
    \caption{Accuracy.}
    \label{fig:app_accuracy_per_overlap_bin}
      \end{subfigure}
      \caption{\textbf{Breakdown of accuracy and copying variant, per token overlap bucket.} For word pairs with partial token overlap, erroneous source word copying is a more frequent error mode than for pairs with no overlap. For pairs with no token overlap, context copying is more prevalent and decays more slowly throughout training. Translation accuracy is higher for pairs with higher token overlap. }
      \label{token_overlap_analysis}
\end{figure*}

\paragraph{Role of token overlap in inference} We categorize each data sample into one of three buckets based on their token overlap: Of the $7,200$ word pairs, $4.40\%$ are valid copies, $18.5\%$ have partial token overlap, and the majority, $77.1\%$, have no token overlap. As shown in \cref{fig:app_accuracy_per_overlap_bin}, the model achieves roughly $90\%$ accuracy on valid copies, with a slight dip in accuracy in early Phase II. This matches our analysis which shows that during this transition, the model suppresses copying. Differentiating the copying variants with respect to token overlap (\cref{fig:app_contextcopying_per_overlap_bin,fig:app_sourceword_per_overlap_bin}), we observe that partial token overlap increases erroneous source word copying, whereas context copying is more prevalent for words with no token overlap. This suggests token overlap supports source word copying as a more \emph{targeted} copying.

\section{Additional Details on ExPLAIND Experiment Settings and Methodology}
\label{app:details_experiment_settings}

\subsection{Mathematical Details on ExPLAIND}
We refer the interested reader to \citet{eichin2025explaindunifyingmodeldata} for details on the methodology and theoretical considerations regarding ExPLAIND. Since we train our model using AdamW \cite{loshchilov2018decoupled}, we can directly use their formulation. 

Since computing the full ExPLAIND decomposition is infeasible for large models and data, we rely on \citet{eichin2025explaindunifyingmodeldata}'s efficient implementation and consider only every 500th parameter update, as well as a subsample of $20\%$ of the training data for all experiments besides the accuracy sanity check below.

\paragraph{Accuracy of the ExPLAIND decomposition.} We investigate whether the decomposition of the actual training influences through ExPLAIND actually recovers the loss trajectory. I.e., whether summing up all the scores is equal to the actual difference of losses in the respective step. We find that this is indeed the case: Over a subsample (N=200 \wlt predictions) of the data, the average difference of the sum of scores to the actual loss change is less than $1\%$ with a standard deviation of one percentage point over a sample of 50 checkpoints uniformly sampled from those available to us.

\paragraph{Influence of first moment term.} Since we only track a subsample of the checkpoints, tracking the indirect influence of data from previous update steps through the first moment term is only possible for scores, where we accumulate over the data dimension of the \emph{tensor of influences}. I.e., we cannot track the influence of single instances or partitions, only the overall influence. This is because, while we have the summed up first moment term from the optimizer checkpoints, disentangling it to the level of previously seen data requires processing these steps and keeping terms for each sample. As \citet{eichin2025explaindunifyingmodeldata} argue, this is infeasible for large models like ours. In the current study, we present two main experiments using ExPLAIND:
(1) Influence of the actual training data (discussed in \cref{app:explaind_actual}), where we accumulate over the data dimension. Here, we include the influences from the first moment terms; 
(2) the influence of simulated parallel samples (shown in \cref{fig:parallel_on_copy_explaind}), where we cannot quantify the influence of the first moment.

\subsection{Test Partitions for ExPLAIND Analysis}
\label{sec:explaind_test_data}
We study different aspects of our model behavior through the loss decomposition of data batches that exhibit the respective behavior. The details on these \emph{test partitions} (a naming we adopt from \citet{eichin2025explaindunifyingmodeldata}'s terminology to mean \emph{is part of the test set} or \emph{the (unseen) samples we want to explain}) are presented in the following.

\paragraph{\wlt instances $B_{WLT}$.} We sample a set of \wlt instances from the pool of concepts that are predicted correctly for at least half of the translations. Since computational complexity of the decomposition is high, we restrict ourselves to instances where the source and target language are Chinese, German, French, or English. The loss we decompose for these instances is computed solely based on the tokens that constitute the target word in the sequences; remaining tokens are masked. This way, we isolate the prediction of the \wlt target from other factors such as predicting quotations etc. 

\paragraph{Challenging \wlt instances $B_{translate}$.} From the complementary set of \wlt concepts which were predicted correctly for less than half of the translations, we sample a set of ``challenging'' \wlt instances. We again focus on the same languages. Furthermore, we verify that copying is the majority error mode in this set during Phase I, where copying is dominant.

\paragraph{Copy \wlt instances $B_{copy}$.} To decompose copy behavior, we take the challenging \wlt instances and switch their correct \wlt target with the source word. In other words, we create samples where copy is assumed to be the "correct" behavior in the loss function we decompose. This loss increases for instances where non-copy behavior becomes the model's preferred strategy. In other words, by decomposing and contrasting the copy instances with the correct behavior, we can isolate the parts of the model that promote and suppress this behavior over time.

\subsection{Training Partitions for ExPLAIND Analysis}
\label{app:explaind_train_data}
As we describe in \cref{sec:methodology_explaind}, ExPLAIND can be leveraged to study the parameter-wise influence of a training data partition onto the predictions over time. In this work, we rely on two main sources for \textbf{training data partitions} to study our model's training dynamics. Namely:

\paragraph{Actual training data $X_{train}$.} For the checkpoints we consider, we store the complete training batch that was used to update the parameters to \emph{the next} checkpoint. In other words, we store the (unseen) data that is used to compute the next training gradients. Since we are interested in cross-language influences, we partition this data by the language distribution it came from, i.e. the language of the FineWeb-HQ subset the data was taken from. This leads to nine training data partitions, one for each language we include.

\paragraph{Parallel data $X_{parallel}$.} To investigate the mechanisms behind copying, we introduce another training partition, which consists of parallel data, which we describe in the following. We assume that this type of data would introduce learning influences that support beneficial \wlt behavior \cite{kondo-etal-2024-enhancing} and suppress copying as a fallback, especially on instances that are not predicted correctly.

First, two annotators from the team of authors (one is a native English speaker, the other a C1-level speaker per IELTS), were instructed to write short, simple sentences that use each of the concepts in English.
The annotators checked each other's sentences to avoid incorrect or disfluent samples.

We then used an ensemble of machine translation models to generate translations of the sentences in the remaining eight languages, namely \texttt{nllb-200-3.3B} \cite{nllbteam2022languageleftbehindscaling} and \texttt{translategemma-27B-it} \cite{gemmatranslate2026}, which support all of our language pairs, and the language-pair specific models from the \texttt{opus-mt} family of models \cite{tiedemann-thottingal-2020-opus}. As a rough proxy for translation quality, for each translation, we backtranslate using the same model and measure the cosine similarity of sequence embeddings produced by \texttt{Qwen3-Embedding-4B} \cite{zhang2025qwen3embeddingadvancingtext} of the original and the backtranslated sentence. For each sample and language, we keep the highest scoring translation in terms of that cosine similarity.

To simulate the appearance of simple parallel data without much interference from other factors, we concatenate the resulting translations back-to-back, separating them by a single whitespace. Our \wlt test partition is restricted to German, French, English, and Chinese. Accordingly, we only consider the respective parallel training data for these languages. For the other languages, the translations are nevertheless included in the supplements of this work. Note that this training data is \emph{simulated}, in the sense that it was not actually part of the training batch at this checkpoint. ExPLAIND allows for this by simply adding these simulated samples to the actual training batch and computing the decomposition in the same manner.

\subsection{Identifying Copy Parameters using ExPLAIND}
\label{sec:copy_ablation_additional_details}
To identify the parameter groups most relevant to copying, we rank them according to their negative ExPLAIND influence in Phase I and their positive influence in Phase II. To keep our interventions sparse, we choose the affected groups by including only the ones which have at least half of the maximum influence. 

In Phase I, this identifies attention value projections of layers $0$--$2$ as the most relevant to copy learning in that phase.

In Phase II, we identify the attention key and query projections in layer 9, the attention key and value projections in layer 8, and the MLP down projections of layers 6 and 7 as the most relevant layers.

\section{Pretraining Data Analysis}\label{sec:data-analysis}
\subsection{Token Overlap}\label{sec:jsd}
\begin{figure}[t]
    \centering
    \includegraphics[width=\linewidth]{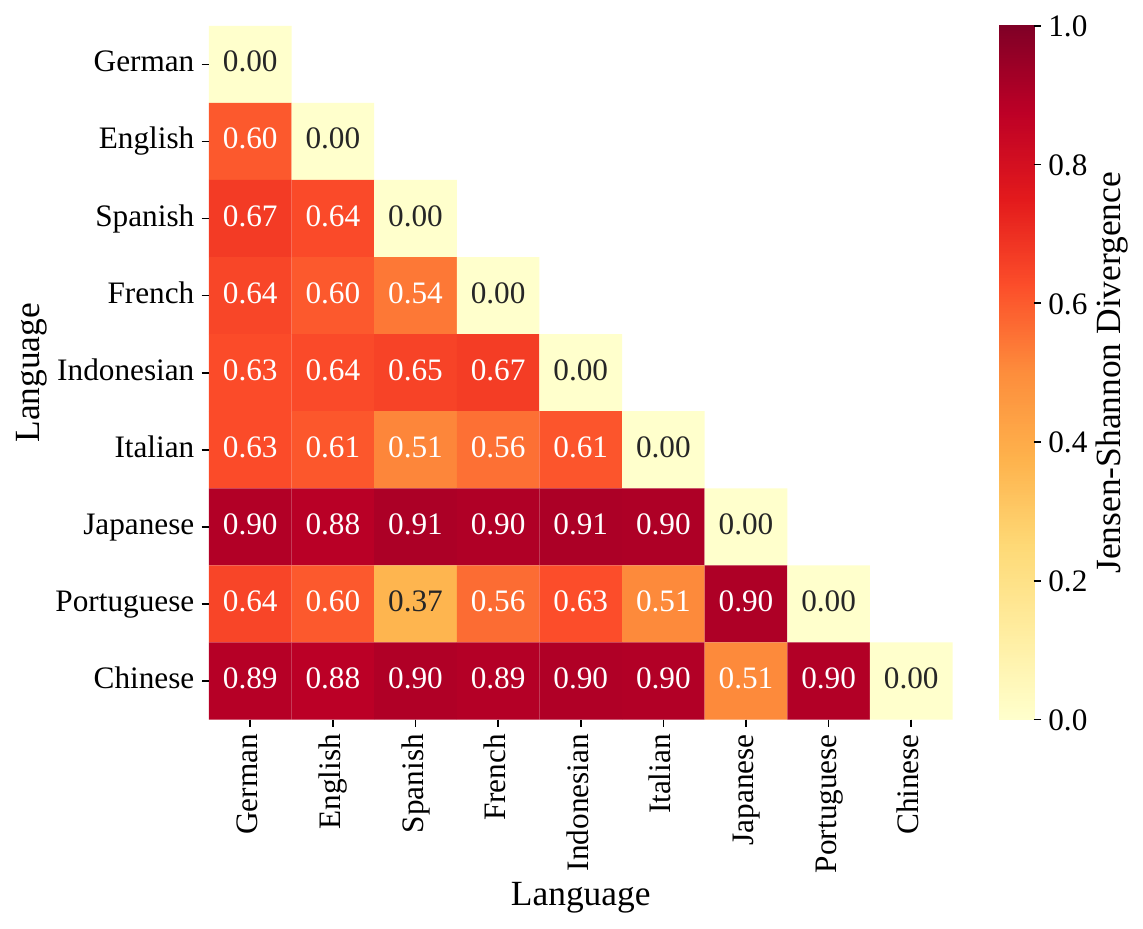}
    \caption{Jensen-Shannon Divergence between language pairs in the pretraining data.}
    \label{fig:jsd}
\end{figure}
Token overlap is considered a key driver in cross-lingual transfer \cite{wu-dredze-2019-beto,patil-etal-2022-overlap,limisiewicz-etal-2023-tokenization,bagheri-nezhad-etal-2025-beyond}. Since we aim to simulate realistic training, we avoid artificial setups as previously studied by \citet{kallini-etal-2025-false}. Instead, we select our nine languages to provide naturally varying amounts of token overlap. 

We confirm this variation in token overlap by calculating Jensen-Shannon Divergence (JSD) \cite{61115} across language pairs in our pretraining data, following \citet{limisiewicz-etal-2023-tokenization} and \citet{hammerl-etal-2025-beyond}. JSD gives a symmetric distance between two token distributions, ranging from 0 to 1. We do not weight JSD to account for the difference in corpus size for English compared to the other languages, as \citet{lu-etal-2020-diverging} argue that weighting misleads results.

As shown in \cref{fig:jsd}, JSD ranges from $0.37-0.91$. Language pairs with differing scripts, such as Japanese and Indonesian, diverge the most; similar language pairs with the same script, such as Portuguese and Spanish, diverge the least.

\subsection{Incidental Parallel Data}\label{sec:parallel}
As described in \cref{sec:pretraining}, we do not intentionally include parallel data, instead, we randomly sample from each language's partition at each step during training. However, \emph{incidental} parallel data may still appear in the training data. To estimate the amount of such incidental parallel data, we mine parallel sentences between each language pair ($36$ total) at 51 out of 200 checkpoints, spaced evenly every 2000 steps between the first checkpoint (step 500) and the last (step $100{,}000$). At each checkpoint, we mine the training sequences seen since the last checkpoint.

We first split each sequence into sentences using an SAT model \cite{frohmann-etal-2024-segment}\footnote{\texttt{sat-3l-sm} via the wtsplit library \cite{minixhofer-etal-2023-wheres}}. We then embed each sentence with LaBSE \cite{feng-etal-2022-language},\!\footnote{\url{https://huggingface.co/sentence-transformers/LaBSE}} a multilingual embedding model. For each source sentence, we retrieve the nearest neighbor in the target language via cosine similarity. Following \citet{artetxe-schwenk-2019-margin}, we use a margin-based criterion: we normalize the pair's cosine similarity by the mean similarity of source and target sentence to their respective $k=4$ nearest neighbors in the other language to compute the margin score. A pair is retained if its margin score is greater than $1.15$ and its raw cosine similarity doesn't exceed $0.99$ (to avoid near-duplicates). We additionally use GlotLID \cite{kargaran-etal-2023-glotlid} to discard any mined pair if either source or target sentence is identified as not belonging to the source or target language, respectively.

Across all language pairs, the mean rate of parallel sentences is low, ranging from $0.24\%$ (Chinese and Indonesian) to $1.61\%$ (English and Spanish), shown in \cref{tab:parallel-data}. Unsurprisingly, since English dominates our pretraining data, pairs including English show higher rates.

We further verify that the rates are relatively constant over training. For each language pair, we compute the Pearson correlation between the mined pair rate and training step across the 51 checkpoints. Correlations are centered at zero across all language pairs (mean $r=0.01$), and none are significant under Bonferroni correction. This indicates that the rate of parallel content consumed between checkpoints does not change significantly, as expected by our random sampling (\cref{sec:pretraining}).

\begin{table}
\centering
\begin{tabular}{lrr}
\toprule
Language pair & Mean rate & CV \\
\midrule
English–Spanish & 1.61\% & 0.083 \\
English–Italian & 1.47\% & 0.066 \\
English–Portuguese & 1.42\% & 0.087 \\
English–French & 1.31\% & 0.063 \\
English–German & 1.04\% & 0.091 \\
English–Japanese & 1.02\% & 0.053 \\
English–Chinese & 0.98\% & 0.066 \\
English–Indonesian & 0.92\% & 0.067 \\
Spanish–Italian & 0.55\% & 0.061 \\
Italian–Portuguese & 0.53\% & 0.069 \\
Spanish–Portuguese & 0.52\% & 0.065 \\
French–Italian & 0.51\% & 0.070 \\
Chinese–Japanese & 0.47\% & 0.052 \\
Spanish–French & 0.46\% & 0.074 \\
German–Italian & 0.43\% & 0.098 \\
French–Portuguese & 0.42\% & 0.087 \\
Italian–Indonesian & 0.39\% & 0.085 \\
German–Spanish & 0.39\% & 0.102 \\
German–French & 0.37\% & 0.109 \\
Chinese–Spanish & 0.36\% & 0.092 \\
Chinese–Italian & 0.36\% & 0.075 \\
Spanish–Indonesian & 0.36\% & 0.099 \\
German–Portuguese & 0.35\% & 0.081 \\
Portuguese–Indonesian & 0.34\% & 0.087 \\
Japanese–Italian & 0.34\% & 0.071 \\
Chinese–French & 0.31\% & 0.061 \\
Spanish–Japanese & 0.31\% & 0.053 \\
Chinese–Portuguese & 0.30\% & 0.085 \\
French–Indonesian & 0.30\% & 0.070 \\
Japanese–French & 0.29\% & 0.058 \\
German–Japanese & 0.27\% & 0.069 \\
Japanese–Portuguese & 0.27\% & 0.058 \\
Japanese–Indonesian & 0.26\% & 0.083 \\
Chinese–German & 0.26\% & 0.065 \\
German–Indonesian & 0.26\% & 0.069 \\
Chinese–Indonesian & 0.24\% & 0.045 \\
\end{tabular}
\caption{Mean rate of mined parallel sentences over sampled checkpoints (51/200) and coefficient of variation (CV), for each of the language pairs. Pairs including English show higher rates, due to its higher proportion in the training data. Rates are consistently low, and the small CV indicates the amount of incidental parallel signal is stable over checkpoints.}\label{tab:parallel-data}
\end{table}

\subsection{How well does training data distribution predict translation learning?}\label{sec:shapley}
To understand how the distribution of our data influences the trajectory of word-pair learning, we use WIMBD \cite{elazar2024whats} to collect cumulative frequency counts of source words, valid translations, and their co-occurrence across training steps. We define co-occurrence as both words appearing within 150 characters of each other in a sequence, using word boundaries for Latin-script languages. For each word pair, we use the respective counts up until that step; i.e., the predictors for a word that is correctly translated at step $10,000$ are the cumulative frequency of the source word, target word, and their co-occurrence up until step $10,000$.

Since these features are inherently correlated (Pearson correlation between $0.31-0.46$), we perform a Shapley value decomposition of $R^2$ \cite{azen2003dominance} to analyze their relative importance in predicting when a word pair is first correctly translated, excluding those that are never correctly translated. We also exclude word pairs where a copy is possible, i.e., any of the valid translations exactly matches the source word. Since our synonym lists vary in length, and are not guaranteed to cover all synonyms, we choose the most frequent count and co-occurrence count across the set of valid translations for our predictors. 

The three predictors explain only $8.7\%$ of the variance in when a word pair is first translated correctly. Of the explained variance, source word frequency accounts for $45.1\%$, co-occurrence for $37.3\%$, and target word frequency for $17.6\%$. Overall, this indicates that simple corpus frequency statistics alone are not a strong predictor of when a word is successfully translated. We leave further investigation to future work.

\section{Additional Results}
We include additional results of our analysis that were omitted from the main paper for brevity.

\subsection{ExPLAIND on Actual Data}\label{app:explaind_actual}
\begin{figure}[t]{}
    \centering
    \includegraphics[trim=0 0 0 73,clip,width=1.0\linewidth]{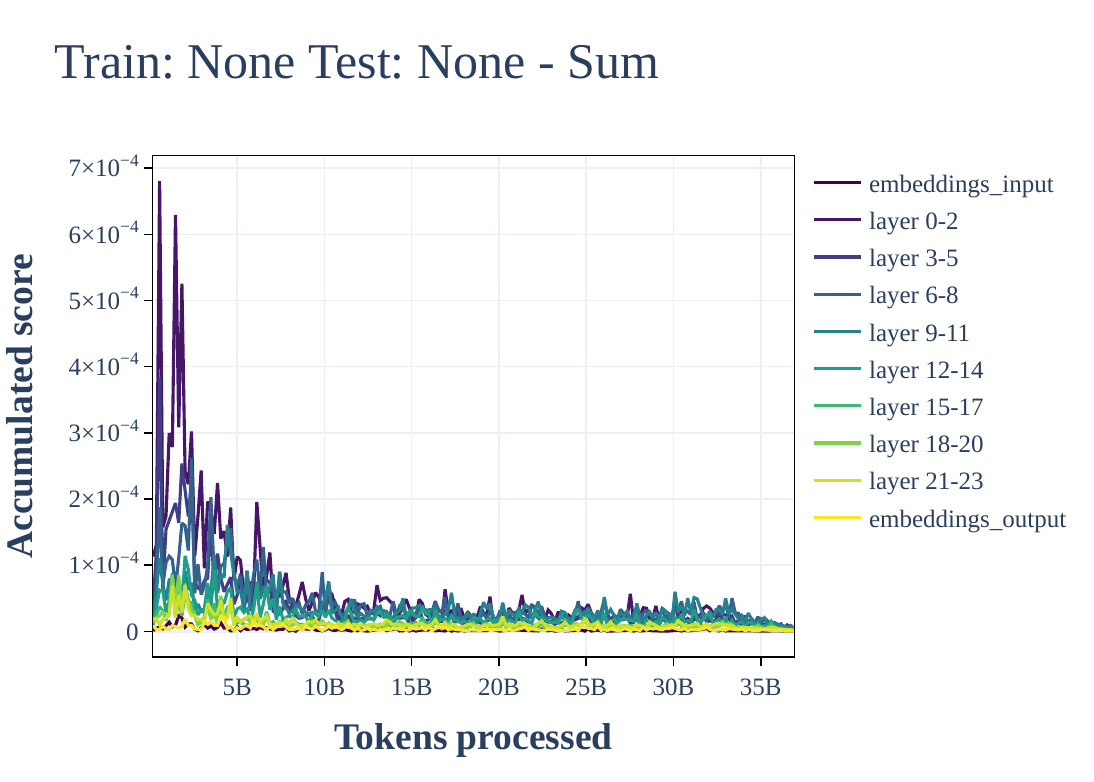}
    \includegraphics[trim=0 0 0 73,clip,width=1.0\linewidth]{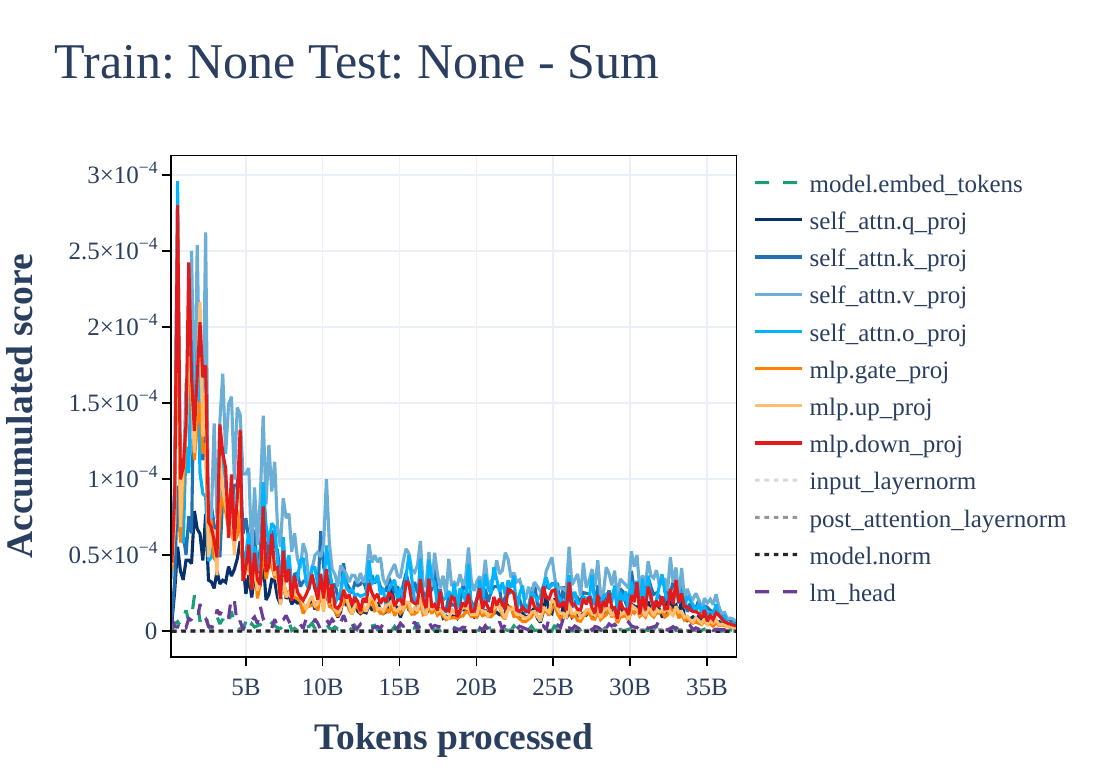}
    \caption{We report the absolute influences $|\Psi(s, \Theta, X_{train}, B_{\wlt})|$ of actual training data $X_{train}$ onto \wlt predictions $B_{\wlt}$ grouped on the parameter level $\Theta$ by subsequent Transformer layers (top) and by layer type (bottom).}
    \label{fig:actual_influences_by_depth}
\end{figure}

In \cref{sec:copy-heuristic}, we use simulated data to identify copy-promoting and copy-suppressing parameters. For comparison, we also study the layer-wise influence of the actual training data $X_{train}$ on a sample of \wlt predictions $B_{\wlt}$, shown in \cref{fig:actual_influences_by_depth}. This shows that Phase I is dominated by changes in the lower layers. The closer a layer is to the input, the more it influences \wlt predictions during this phase. Their strong influence in Phase I is consistent with our finding that lower layers are more relevant for the copy mechanism, which emerges and is dominant in Phase I (\cref{sec:copy-heuristic}).
Further, we find that in the first half of Phase I the decrease in loss is driven equally by changes in the attention and MLP layers. This trend reverses around $5$B tokens processed, where the relative influence of the attention layers on the loss change increases compared to the MLP influences. This may reflect the development of more targeted copying in the latter part of Phase I, where context copying decreases while source word copying increases (\cref{fig:copying_error_mode}). 

\subsection{Logit Lens}
Here we provide more fine-grained versions of the logit-lens analyses shown in the main paper. First, we group candidate words in other languages by their relationship to the translation task (source word, English, languages sharing the target script group, or languages with a different script group), which highlights how concepts develop language-(in)dependently. Second, we group translation candidates by token overlap between the source and target words (no overlap, partial overlap, full). This view isolates how token overlap aids translation.

\begin{figure*}[t]
 \centering
        \begin{subfigure}[b]{0.49\textwidth}
        \centering
        \includegraphics[width=\textwidth]{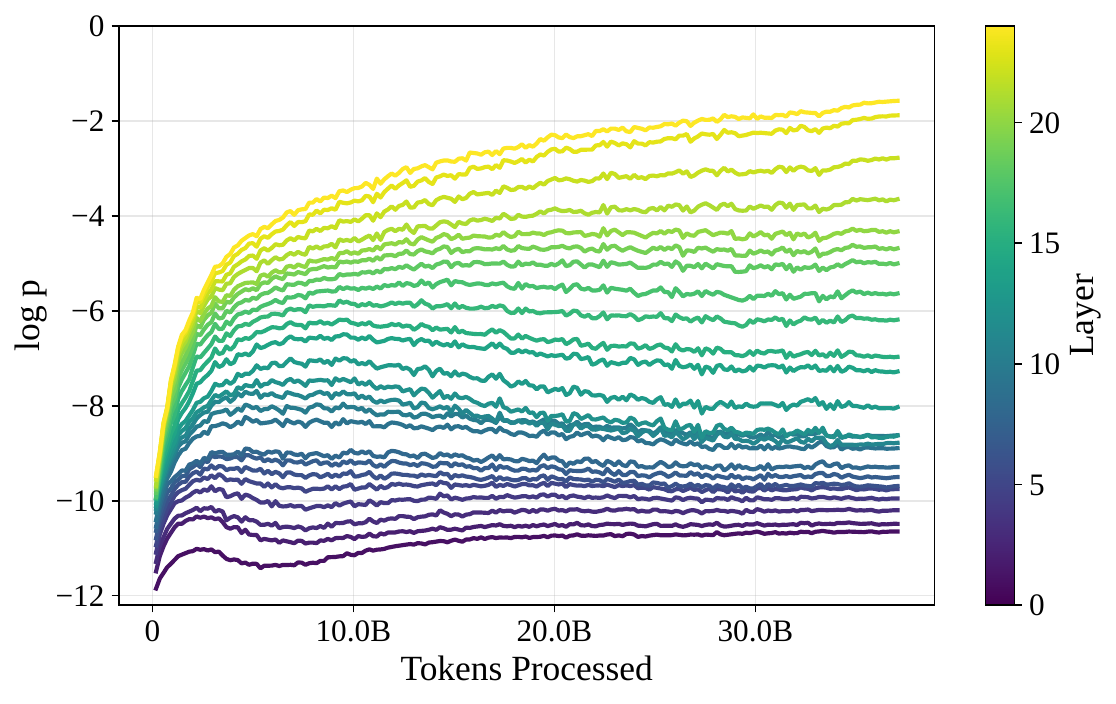}
        \caption{Translation (all).}
        \label{fig:ll-over-time-translation}
    \end{subfigure}
    \hfill
    \begin{subfigure}[b]{0.49\textwidth}
        \centering
        \includegraphics[width=\textwidth]{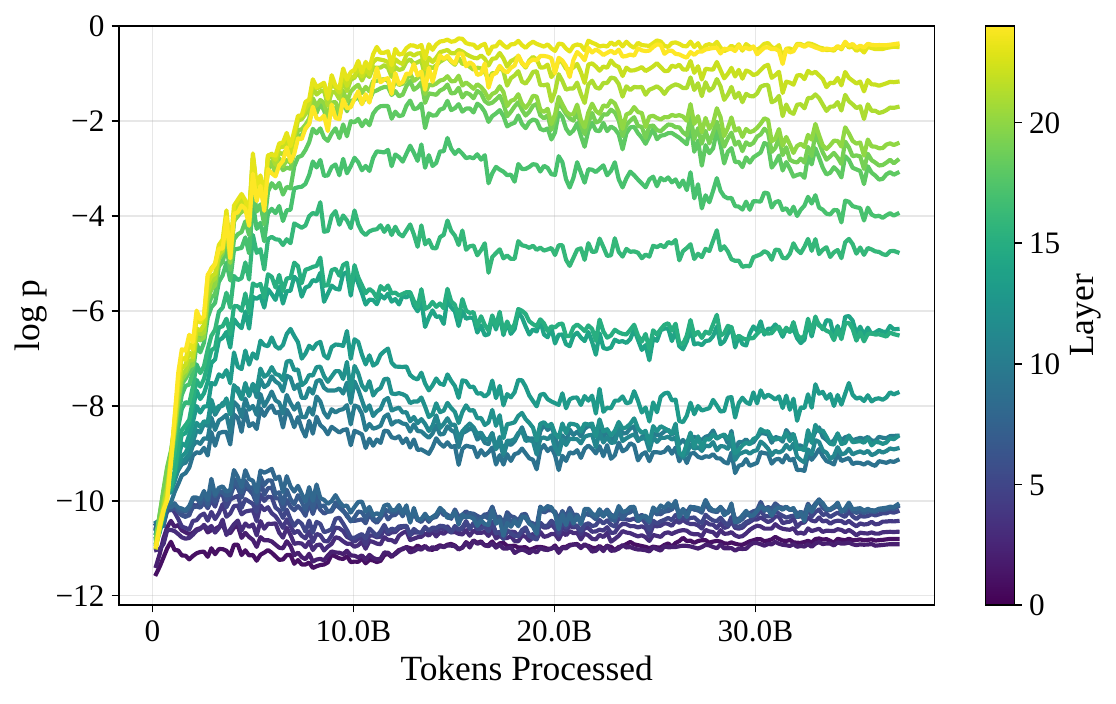}
        \caption{Identical tokens (copying).}
        \label{fig:ll-identical-tokens}
    \end{subfigure}
    \begin{subfigure}[b]{0.49\textwidth}
        \centering
        \includegraphics[width=\textwidth]{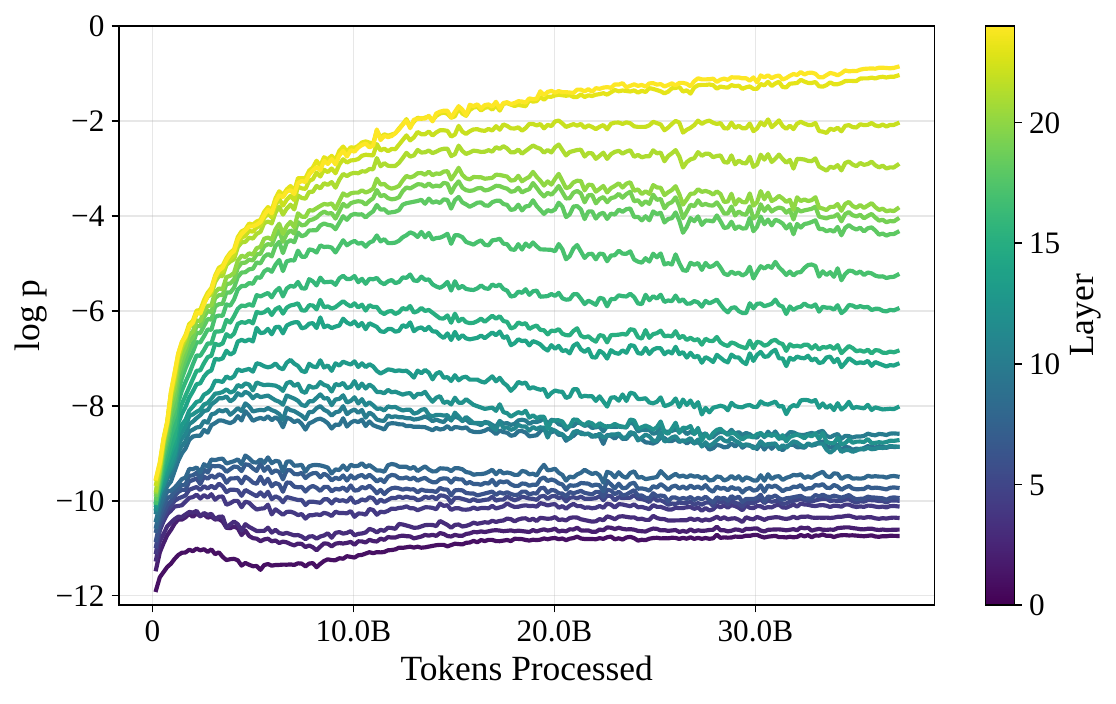}
        \caption{Partial token overlap.}
        \label{fig:ll-over-time-partial}
    \end{subfigure}
    \hfill
    \begin{subfigure}[b]{0.49\textwidth}
        \centering
        \includegraphics[width=\textwidth]{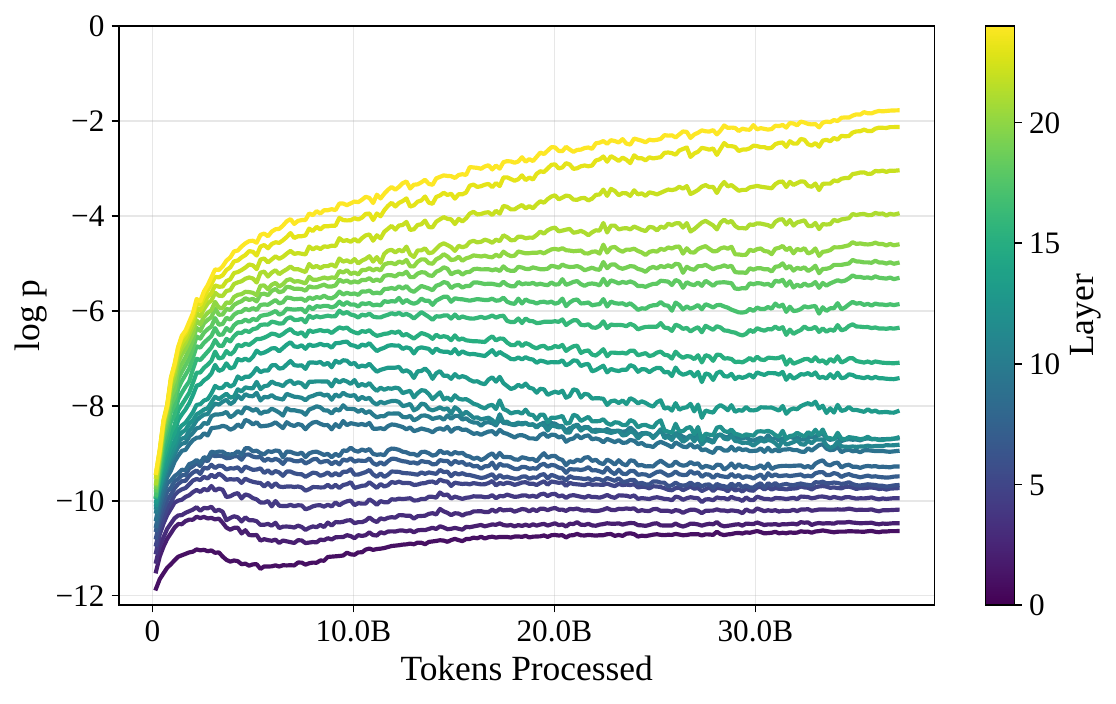}
        \caption{No token overlap.}
        \label{fig:ll-no-token-overlap}
        \end{subfigure}
    \caption{\textbf{Evolution of layer-wise log probabilities of true translation across token-overlap categories.} \cref{fig:ll-over-time-translation} shows all translation tasks averaged across language-pair means. \cref{fig:ll-over-time-partial,fig:ll-no-token-overlap,fig:ll-identical-tokens} separate the same data into token-overlap buckets between the source and target word (no overlap, partial overlap, and identical tokens). \\
    For the full \textit{translation} task, the bottom and intermediate layer blocks remain flat or slightly decrease after an initial increase, while the top layers' log probabilities increase steadily throughout training. Restricting to token overlap, the dynamics in the top block are more nuanced: while the outermost two layers' log probabilities continue to increase over training, for words with token overlap, the inner layers of the top block show a slight decrease. This mirrors the suppression previously observed for the source-word candidate, suggesting competition between copying and translating. In the outermost two layers, this competition is resolved in favor of the true translation (with identical or partial token overlap).  
    }

    \label{fig:app_over_time_token_groups}
\end{figure*}

\begin{figure*}[t]
 \centering
        \begin{subfigure}[b]{0.49\textwidth}
        \centering
        \includegraphics[width=\textwidth]{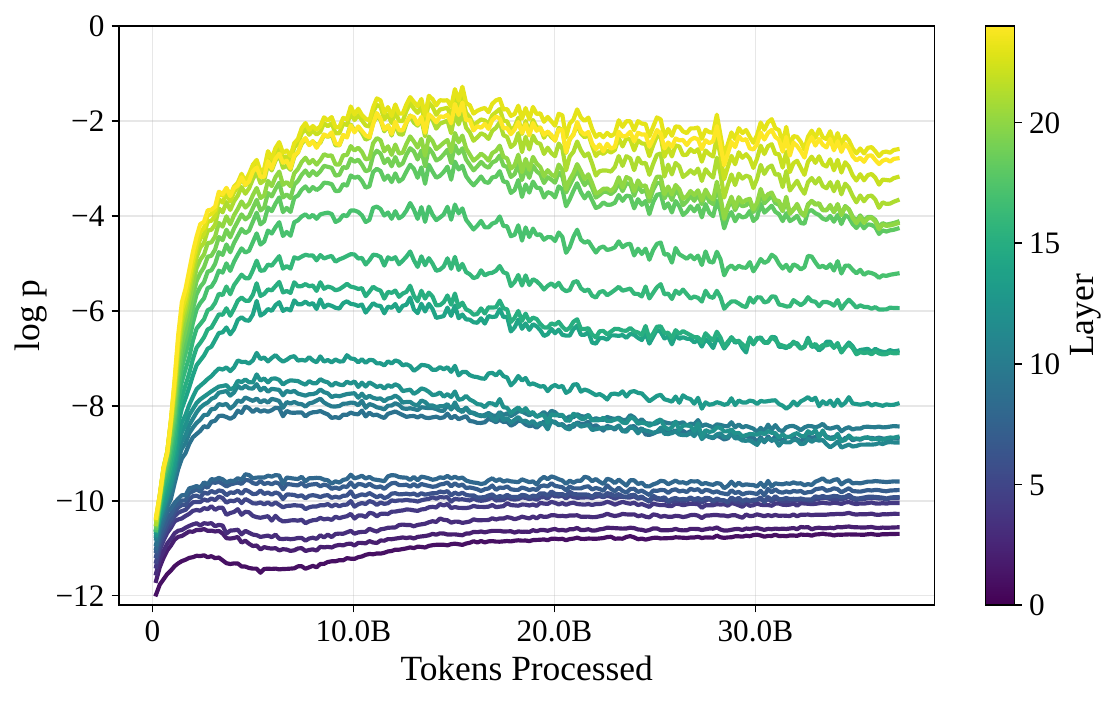}
        \caption{Copying.}
        \label{fig:ll-over-time-copying}
    \end{subfigure}
    \hfill
    \begin{subfigure}[b]{0.49\textwidth}
        \centering
        \includegraphics[width=\textwidth]{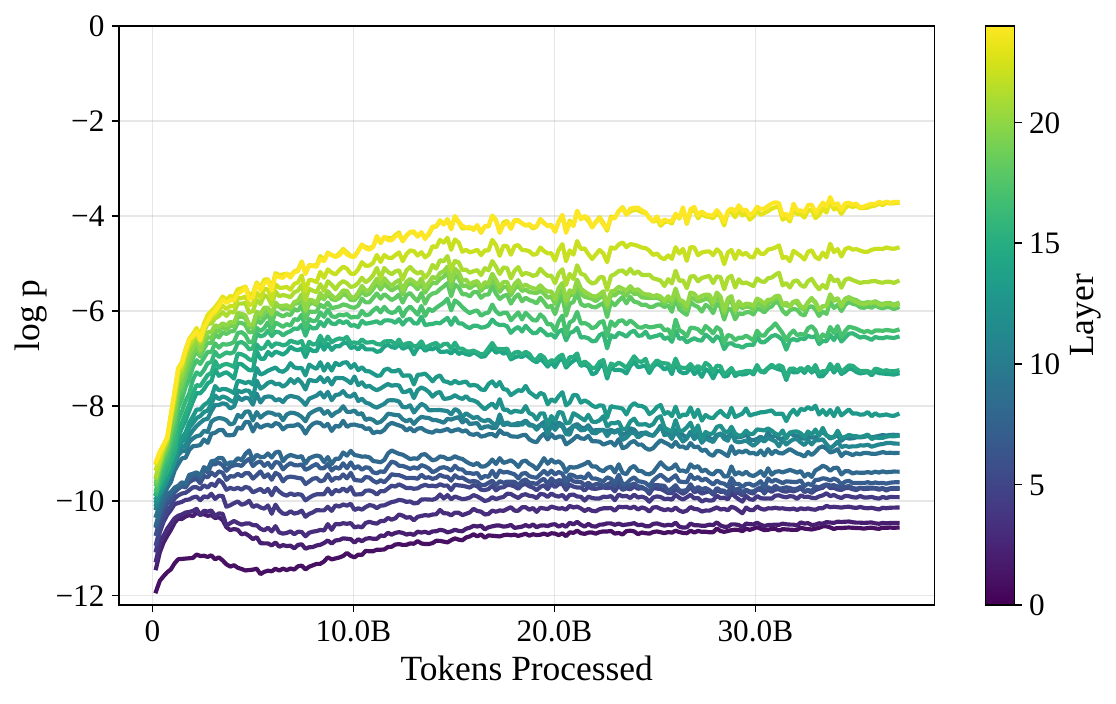}
        \caption{English.}
        \label{fig:ll-over-time-english}
    \end{subfigure}
    \begin{subfigure}[b]{0.49\textwidth}
        \centering
        \includegraphics[width=\textwidth]{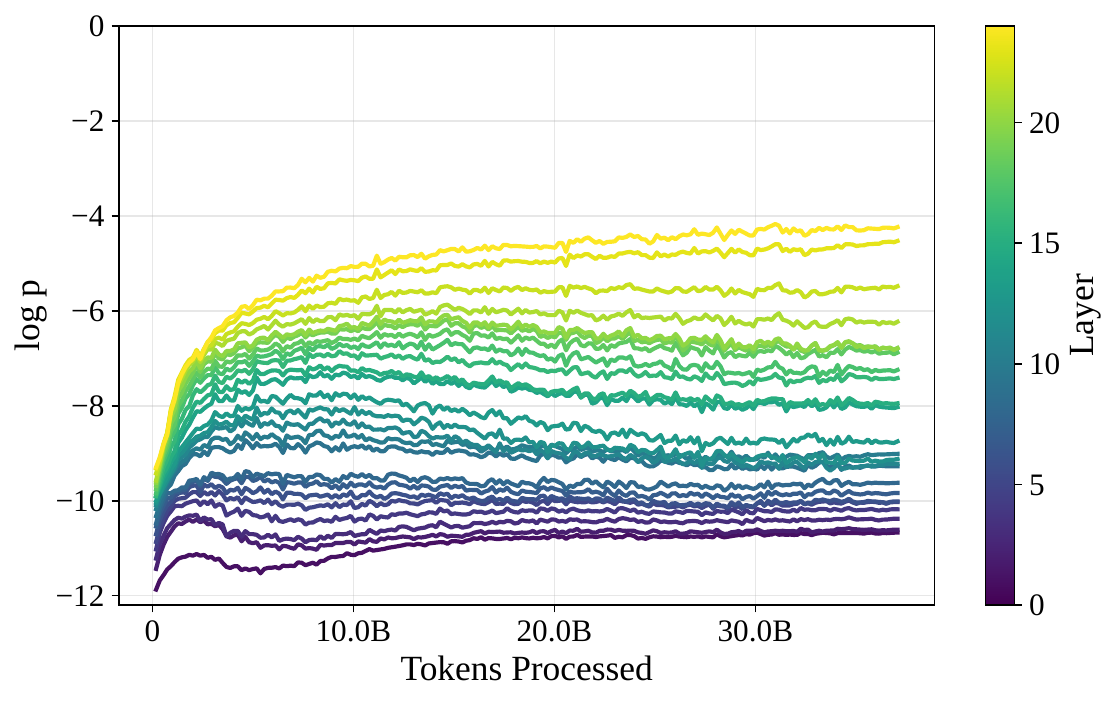}
        \caption{Same script group as the target language.}
        \label{fig:ll-over-time-same-as-target}
    \end{subfigure}
    \hfill
    \begin{subfigure}[b]{0.49\textwidth}
        \centering
        \includegraphics[width=\textwidth]{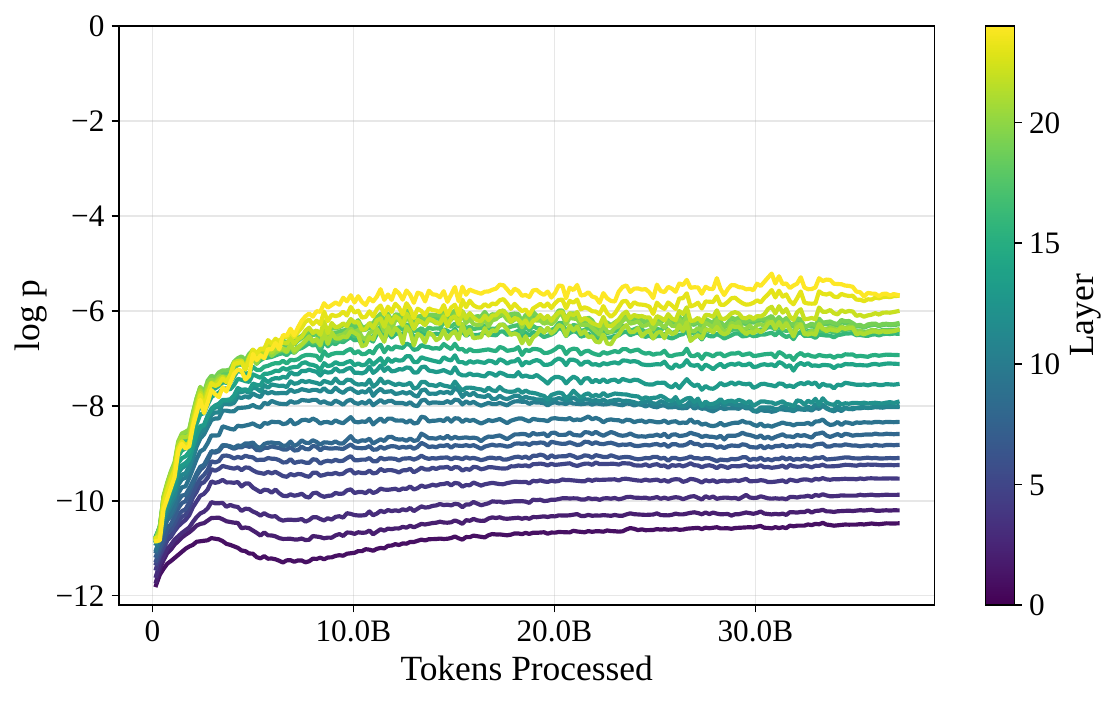}
        \caption{Different script group than the target language.}
        \label{fig:ll-over-time-diff-from-target}
        \end{subfigure}
    \caption{\textbf{Evolution of layer-wise log probabilities for different language categories.} Evolution of log probabilities across training steps for different language categories. For each language pair, we evaluate the model under teacher forcing and track the logits assigned to candidate translations of the same concept in all nine languages. These candidates are grouped according to their relationship to the translation task (source word, English, languages sharing the target script group, or languages with a different script group). We then aggregate the result across all language pairs. Each curve corresponds to a model layer. In this figure we show only language groups corresponding to languages other than the target language; which is shown in detail in  \cref{fig:app_over_time_token_groups}.\\
   For the \textit{source word}, the outer block downranks the copying candidate, evident from a decreasing trend in the second phase. We additionally observe step-wise jumps between some layers (cf. \cref{fig:margin_gradient_ll}). \textit{English} (aggregated only in the case when it is neither the source, nor target language) generally has higher log-probabilities, which reflects its dominance in the training data. Concepts with the \textit{same script group} as the target language show an upward trend of the outer layers log probability, albeit with lower absolute magnitudes than true translations. Languages with a \textit{different script group} yield the lowest final log probability and stay flat throughout the second phase. In contrast to the other groups, some layers close to the output fail to increase the final probability.}
    \label{fig:app_over_time_tf_groups}
\end{figure*}

\begin{figure*}[t]
 \centering
        \begin{subfigure}[b]{0.33\textwidth}
        \centering
        \includegraphics[trim= 0 0 0 0,clip,height=4.3cm]{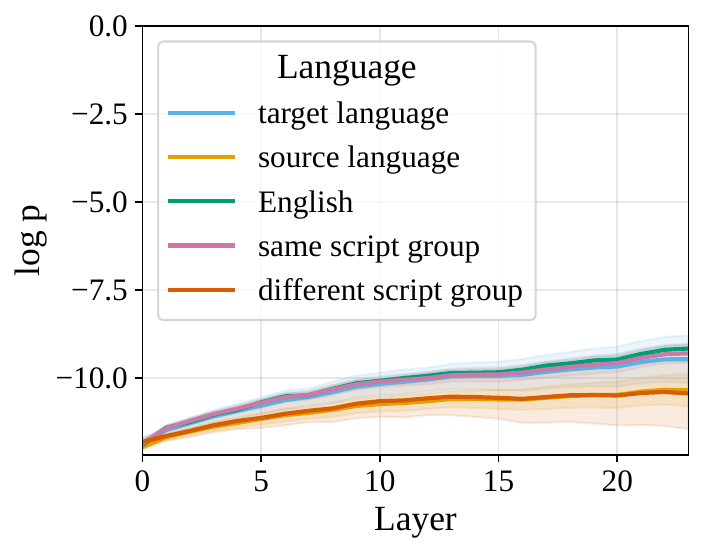}
        \caption{$185$M tokens.} %
        \label{fig:ll-over-layer-step500}
    \end{subfigure}
    \hfill
     \begin{subfigure}[b]{0.31\textwidth}
        \centering
        \includegraphics[trim= 1cm 0 0 0,clip,height=4.3cm]{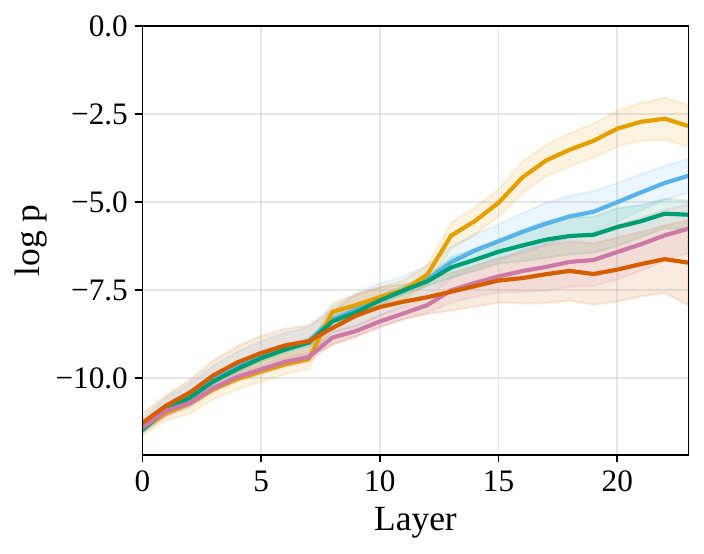}
        \caption{$5.75$B tokens.} %
        \label{fig:ll-over-layer-step6500}
    \end{subfigure}
    \hfill
     \begin{subfigure}[b]{0.31\textwidth}
        \centering
        \includegraphics[trim= 1cm 0 0 0,clip,height=4.3cm]{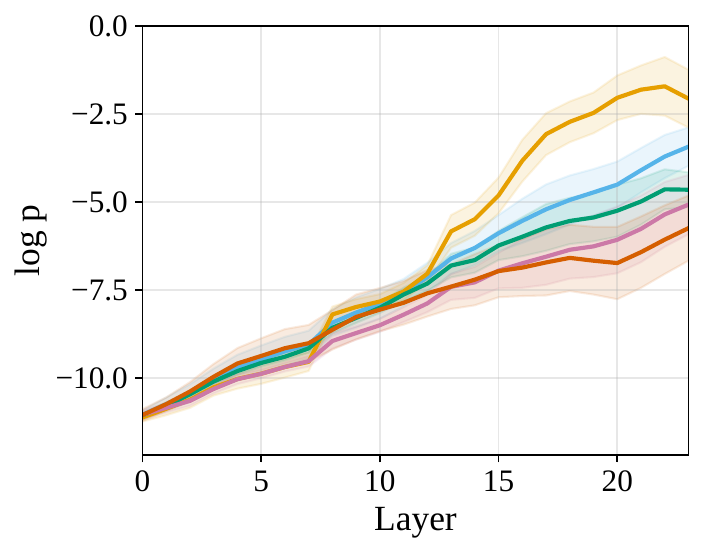}
        \caption{$11.0$B tokens.} %
        \label{fig:ll-over-layer-step295000}
    \end{subfigure}
        \vspace{.1cm}
        \begin{subfigure}[b]{0.33\textwidth}
        \centering
        \includegraphics[trim= 0 0 0 0,clip,height=4.3cm]{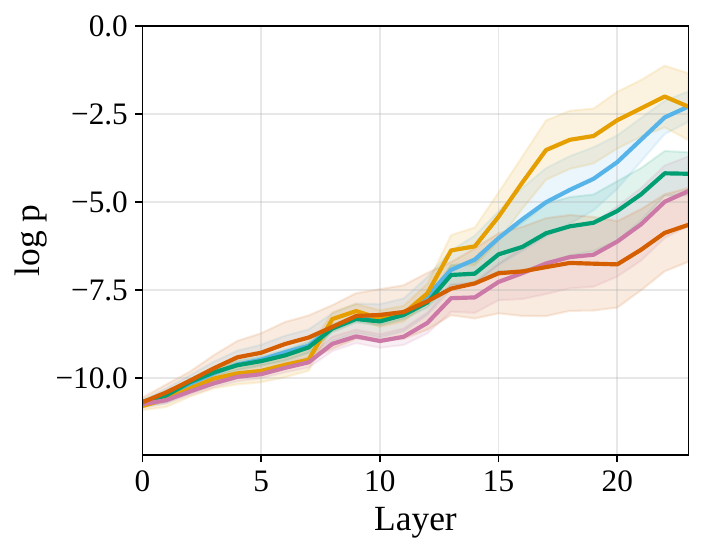}
        \caption{$19.9$B tokens.} %
        \label{fig:ll-over-layer-step53500}
    \end{subfigure}
    \hfill
     \begin{subfigure}[b]{0.31\textwidth}
        \centering
        \includegraphics[trim= 1cm 0 0 0,clip,height=4.3cm]{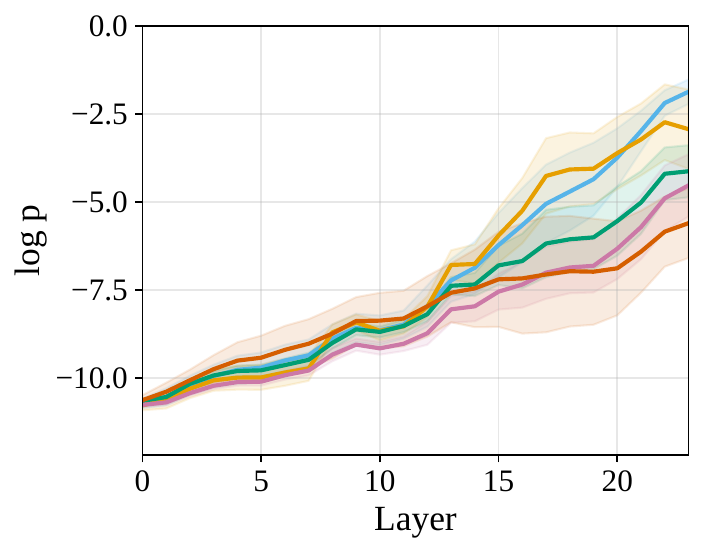}
        \caption{$28.4$B tokens.} %
        \label{fig:ll-over-layer-step76500}
    \end{subfigure}
\hfill
     \begin{subfigure}[b]{0.31\textwidth}
        \centering
        \includegraphics[trim= 1cm 0 0 0,clip,height=4.3cm]{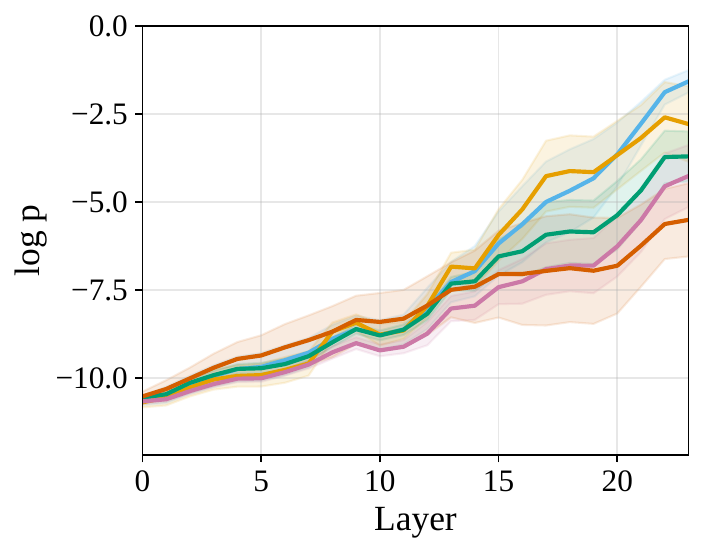}
        \caption{$37.5$B tokens.}
        \label{fig:ll-over-layer-step100000}
    \end{subfigure}
    \hfill
    \caption{\textbf{Evolution of layer-wise log probabilities during training, grouped by language.} Aggregated across all language pairs, we visualize the progression of log probability across model layers, grouped by their relationship to the translation task. In the bottom and intermediate layers, all languages are close to each other. From layer $15$ onward, the source word and, later in training, also the correct translation's log probability increase more rapidly than the other groups. Other languages still show a modest increase, led by English (the dominant language), then languages sharing the same script group as the target, and finally languages with a different script showing the smallest increase. As training progresses, the target translation continues to gain probability mass, eventually overtaking the source word which initially dominates as the primary error mode (\cref{fig:copying_error_mode}). The step-wise increases in log-probability of the copying candidate at specific depths (cf. \cref{fig:margin_gradient_ll}, \cref{fig:app_over_time_tf_groups}), are clearly visible throughout the training. At the final layer, the source word's log probability notably dips. 
    The error bars show standard deviation of language pair averages.}
    \label{fig:app_over_layers_steps}
\end{figure*}

\begin{figure*}[t]
 \centering
        \begin{subfigure}[b]{0.33\textwidth}
        \centering
        \includegraphics[trim= 0 0 0 0,clip,height=4.3cm]{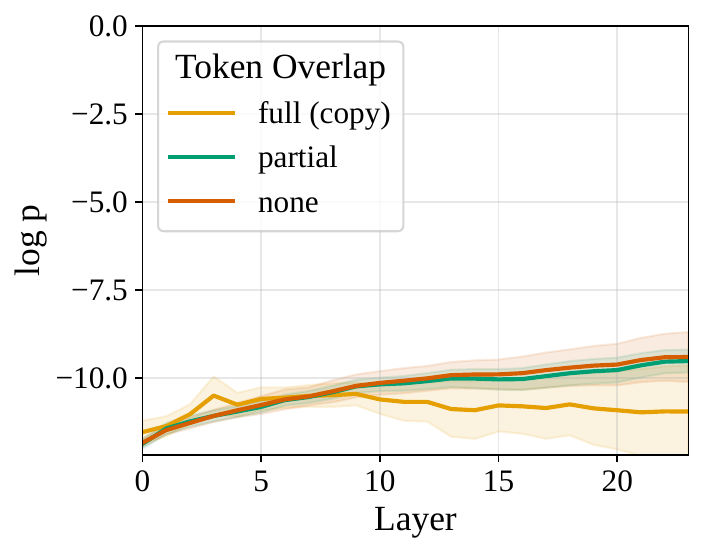}
        \caption{$185$M tokens.} %
        \label{fig:ll-over-layer-jaccard-step500}
    \end{subfigure}
    \hfill
     \begin{subfigure}[b]{0.31\textwidth}
        \centering
        \includegraphics[trim= 1cm 0 0 0,clip,height=4.3cm]{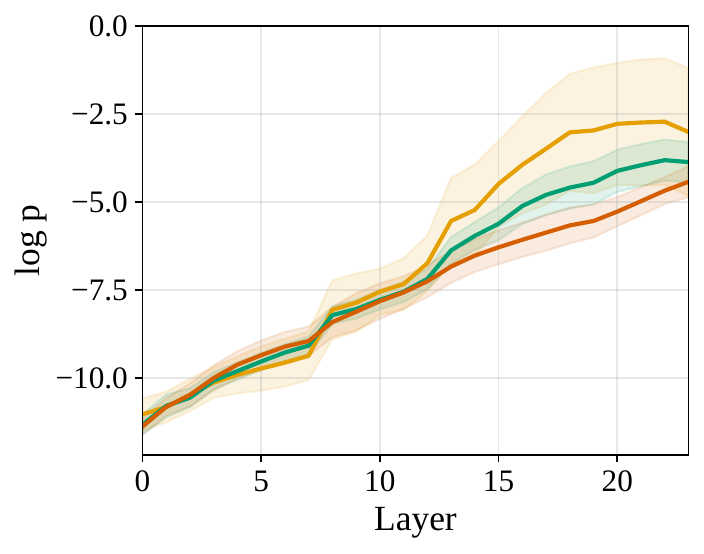}
        \caption{$5.75$B tokens.} %
        \label{fig:ll-over-layer-jaccard-step6500}
    \end{subfigure}
     \begin{subfigure}[b]{0.31\textwidth}
        \centering
        \includegraphics[trim= 1cm 0 0 0,clip,height=4.3cm]{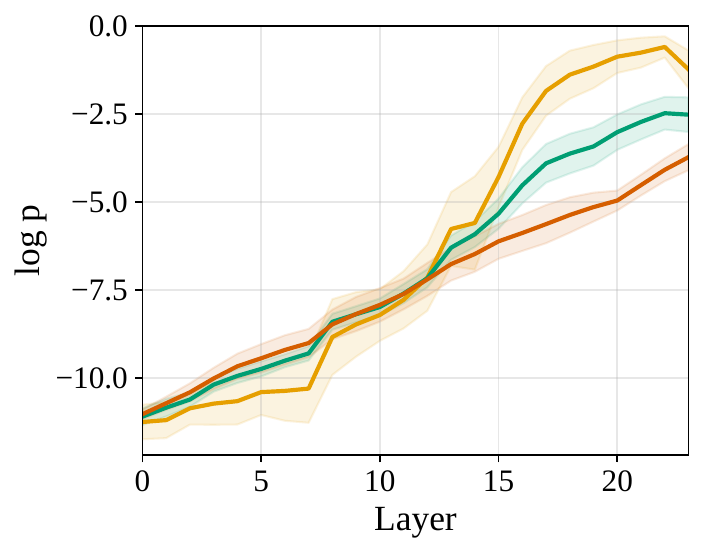}
        \caption{$11.0$B tokens.} %
        \label{fig:ll-over-layer-jaccard-step29500}
    \end{subfigure}
        \vspace{.1cm}
        \begin{subfigure}[b]{0.33\textwidth}
        \centering
        \includegraphics[trim= 0 0 0 0,clip,height=4.3cm]{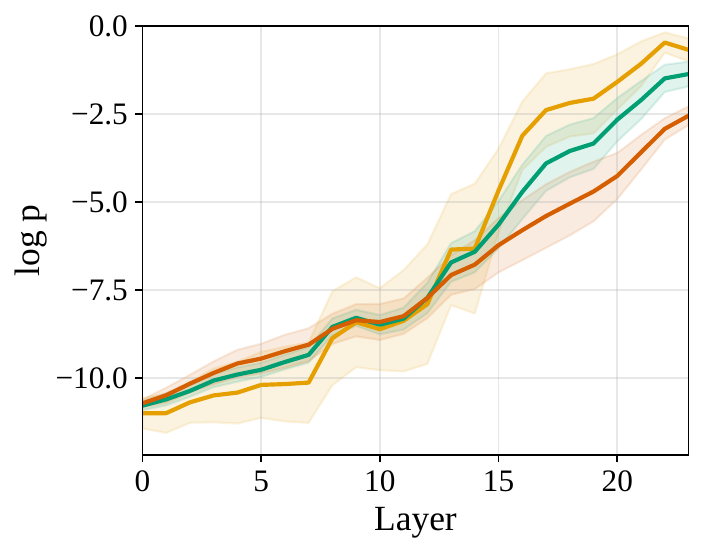}
        \caption{$19.9$B tokens.} %
        \label{fig:ll-over-layer-jaccard-step53500}
    \end{subfigure}
    \hfill
     \begin{subfigure}[b]{0.31\textwidth}
        \centering
        \includegraphics[trim= 1cm 0 0 0,clip,height=4.3cm]{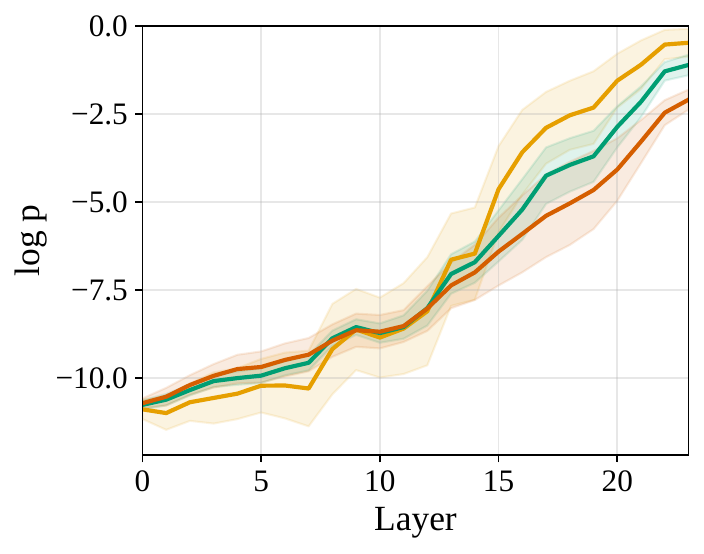}
        \caption{$28.4$B tokens.} %
        \label{fig:ll-over-layer-jaccard-step76500}
    \end{subfigure}
\hfill
     \begin{subfigure}[b]{0.31\textwidth}
        \centering
        \includegraphics[trim= 1cm 0 0 0,clip,height=4.3cm]{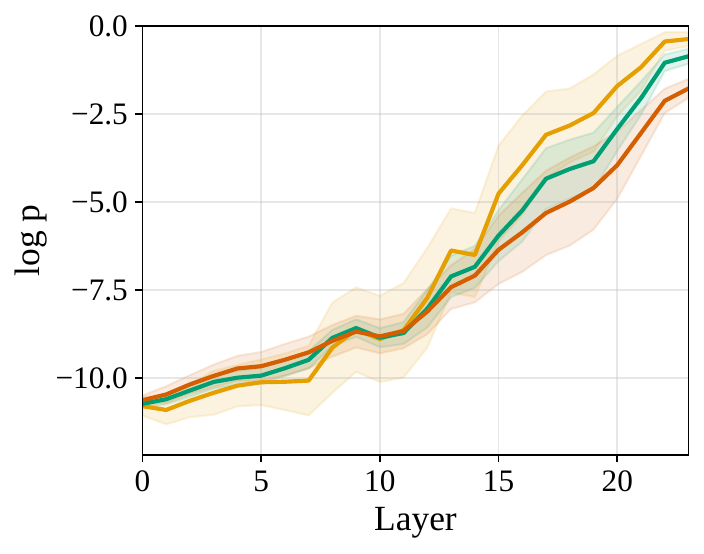}
        \caption{$37.5$B tokens.}
        \label{fig:ll-over-layer-jaccard-step100000}
    \end{subfigure}
    \hfill
    \caption{\textbf{Evolution of layer-wise log probabilities during training, grouped by token overlap.} We analyze the translation task and group the highest-ranking target-language candidate according to its token overlap with the source word. During training, all groups show increasing log probability with layers depths. Around layer $14$, the groups begin to separate: candidates with higher token overlap on average receive higher log probabilities, while candidates with no overlap remain lower but still increase steadily. The exact-copy candidate, which is a valid translation in $4.4\%$ of all translation pairs, initially exhibits a dip in the outermost layer at earlier checkpoints, reminiscent of the dip observed for the source-word candidate (\cref{fig:app_over_layers_steps}). As training progresses, this dip disappears, suggesting that the suppression in the top block observed for copying candidates is reduced when the copied token corresponds to the correct translation. Overall, greater token overlap between source and target appears to support translation, with higher-overlap candidates consistently receiving higher log probabilities in the top layer block. Error bars denote the standard deviation across language-pair averages, counting only language pairs that have at least one concept in the respective groups (e.g., the dataset contains no words with full token overlap (i.e., exact copies) for French–Japanese).}
    \label{fig:app_over_layers_steps_jaccard}
\end{figure*}

\cref{fig:app_over_time_tf_groups,fig:app_over_time_token_groups} provide more fine-grained versions of \cref{fig:main-ll-over-time}, while \cref{fig:app_over_layers_steps,fig:app_over_layers_steps_jaccard} show the corresponding layer-wise views at selected training checkpoints, complementing \cref{fig:main_ll-language_splitting}.

\subsection{Copy Ablation}\label{app:copy_ablations}
\begin{figure*}
    \centering
    \begin{subfigure}[t]{1.0\textwidth}
    \centering
        \includegraphics[trim=0 0 0 70,clip,width=0.7\linewidth]{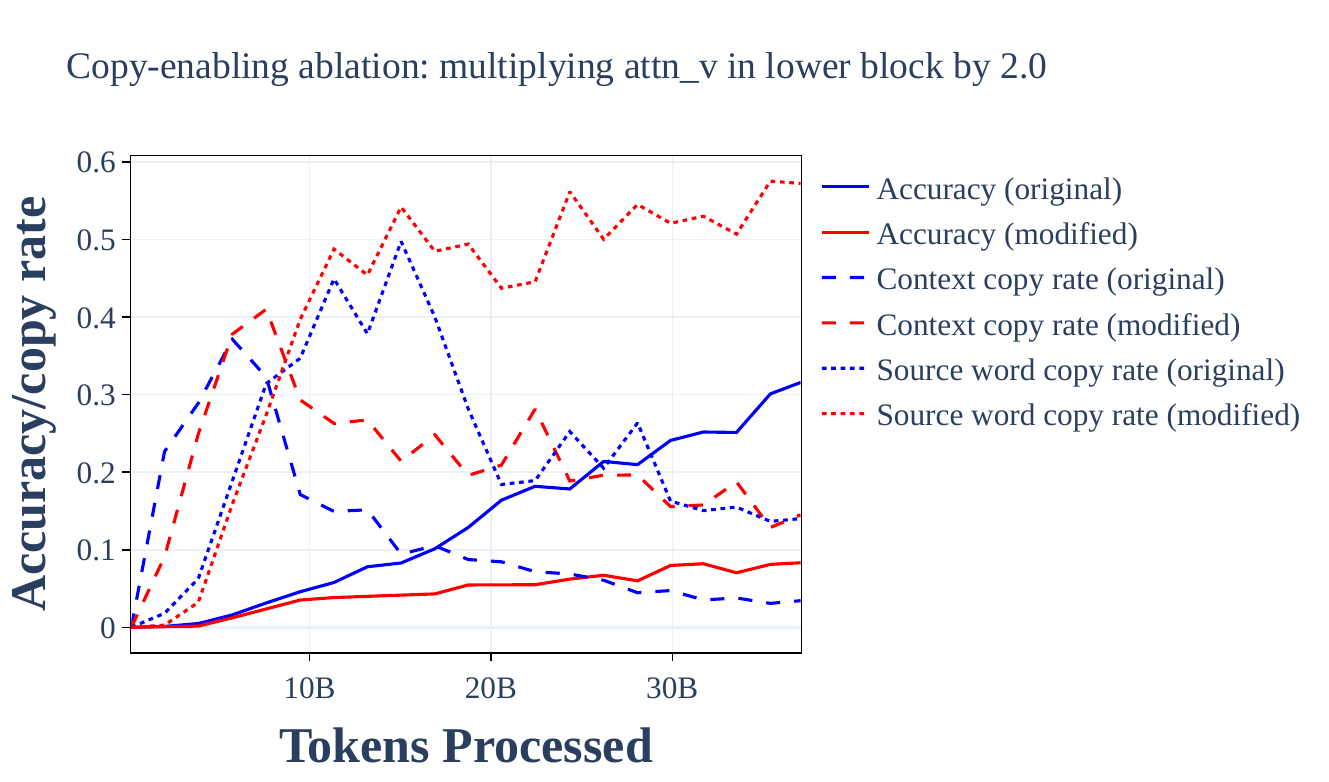}
        \caption{Upscaling by $2.0$.}
        \vspace{20pt}
    \end{subfigure}

    \begin{subfigure}[b]{1.0\textwidth}
    \centering
        \includegraphics[trim=0 0 0 70,clip,width=0.7\linewidth]{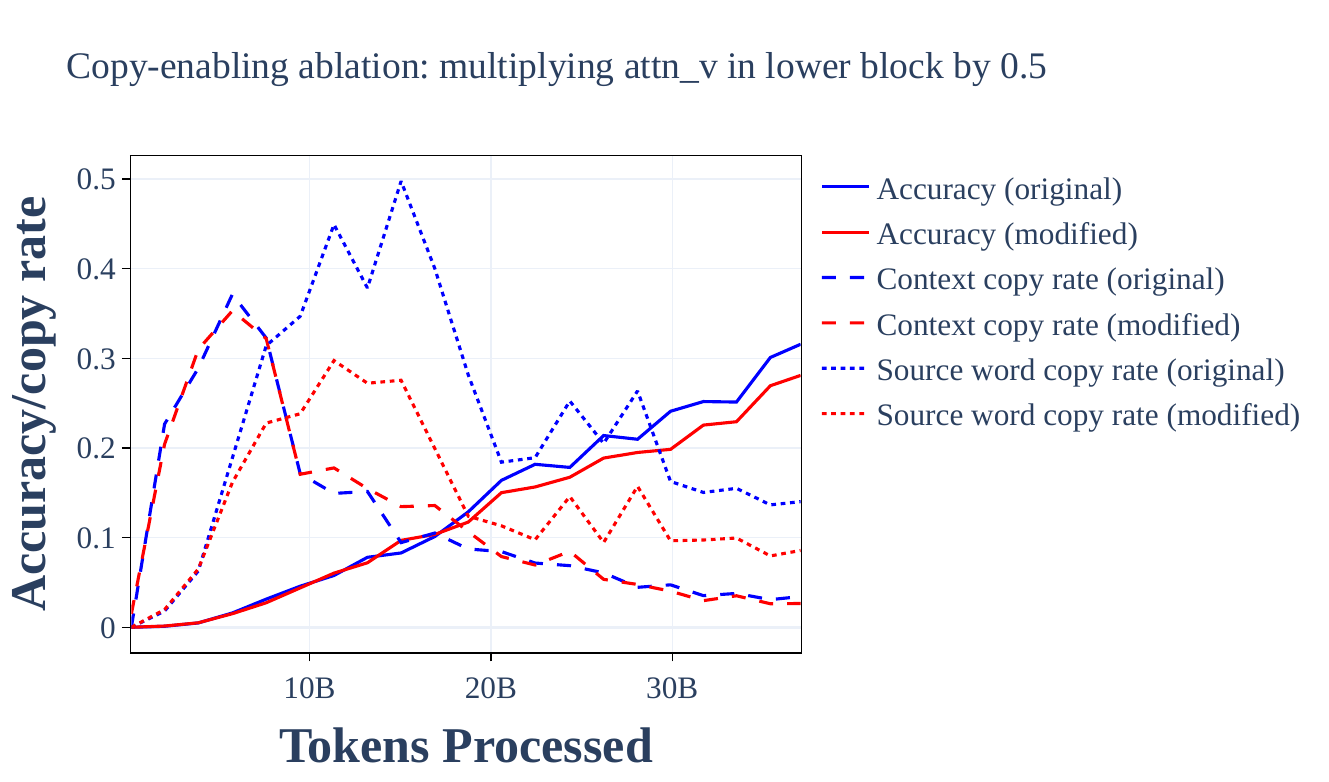}
        \caption{Downscaling by $0.5$.}\
        \vspace{20pt}
    \end{subfigure}
    
    \begin{subfigure}[b]{1.0\textwidth}
    \centering
        \includegraphics[trim=0 0 0 70,clip,width=0.5\linewidth]{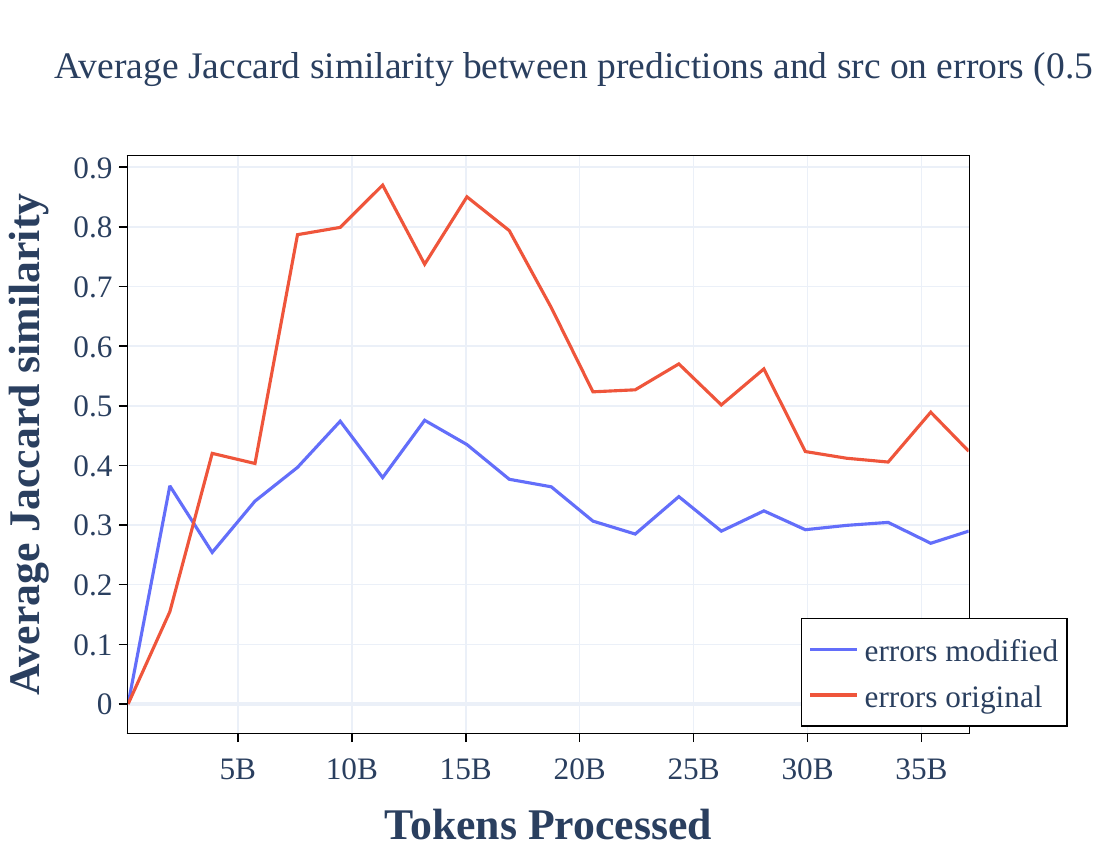}
        \caption{Errors of ablated model with downscaled attention values vs. errors of original model.}
    \end{subfigure}
    \caption{\textbf{Exciting and inhibiting copy behavior in layers $0$--$2$.} We scale the parameters of the copy-promoting parameter groups as identified by ExPLAIND in Phase I and detailed in  \cref{sec:methodology_explaind}. The corresponding parameter groups are the self attention value projections in layers $0$, $1$ and $2$. To investigate their effect on the copy behavior of the model, we perform a naive ablation by doubling and halving their parameters (a and b). Furthermore, we investigate the errors of the ablated model to the ones of the original model in the second scenario: copy-inhibiting. In the final checkpoint, the overall number of exact copies halves from 2203 to 964. For the predictions where at least one of the respective models is correct, we plot the average Jaccard similarity (normalized character overlap), of the predicted word and the source word. After copying in the model is learned successfully (around $8$B), the model with intervention shows as low as half the similarity to the source word, which further supports that our naive approach successfully ablates part of the copy mechanism of the model.}
    \label{fig:copy_ablations_prom}
\end{figure*}

\begin{figure*}
    \centering

    \begin{subfigure}[b]{1.0\textwidth}
    \centering
        \includegraphics[trim=0 0 0 70,clip,width=0.6\linewidth]{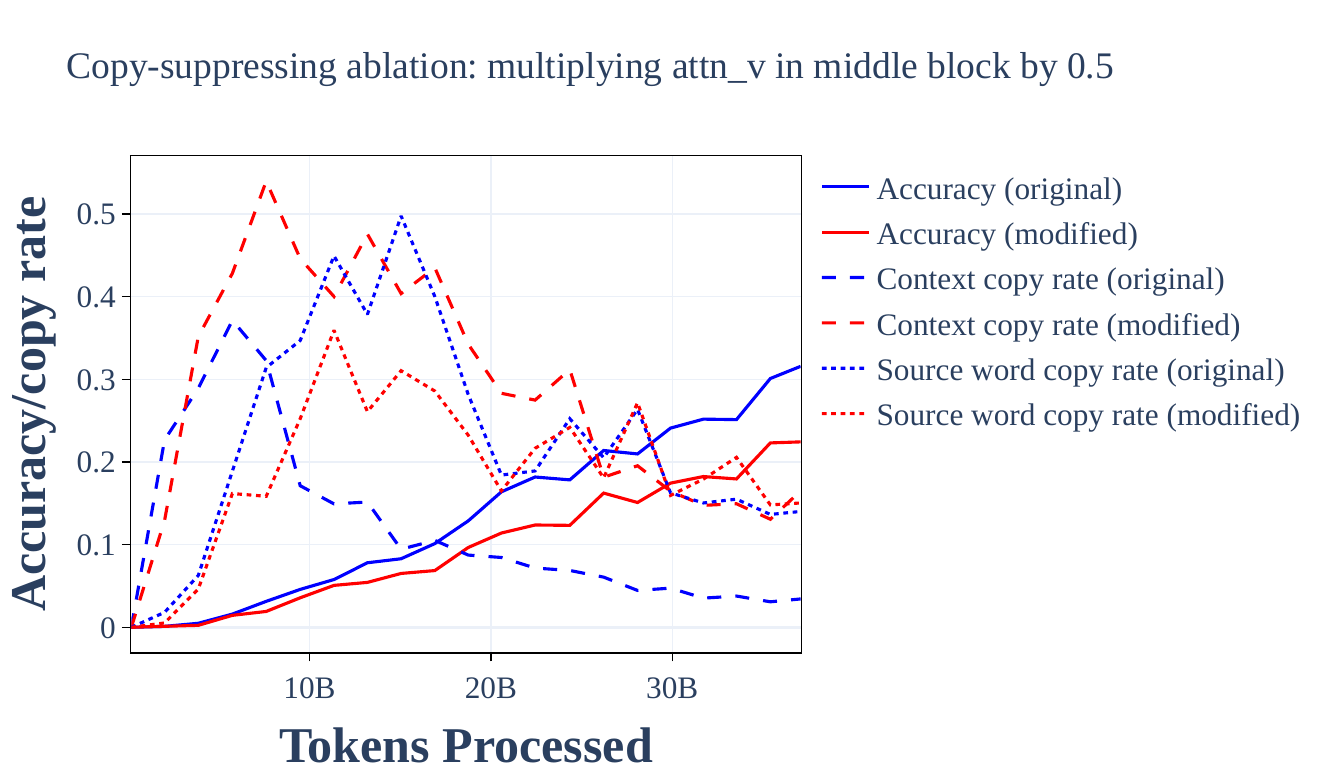}
        \caption{Downscaling by $0.5$.}\
        \vspace{20pt}
    \end{subfigure}

    \begin{subfigure}[b]{1.0\textwidth}
    \centering
        \includegraphics[trim=0 0 0 70,clip,width=0.6\linewidth]{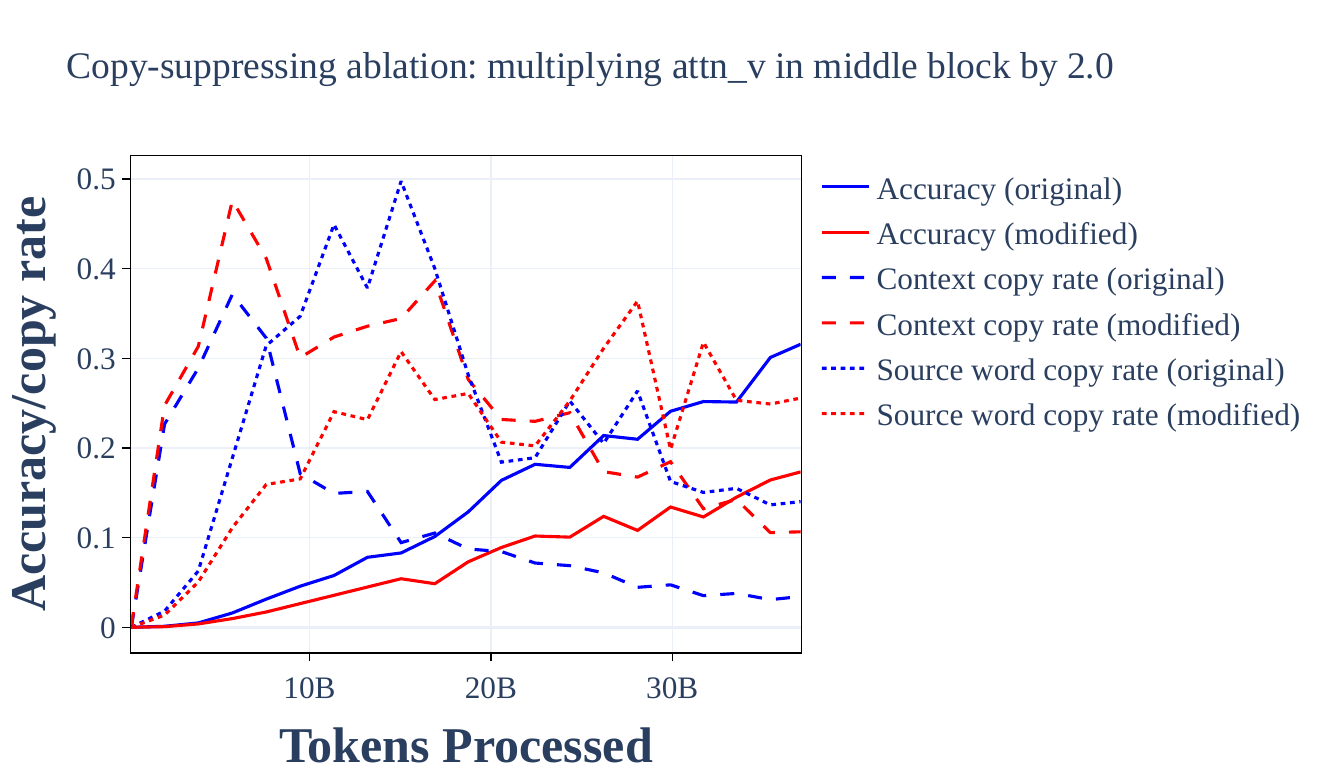}
        \caption{Upscaling by $2.0$.}\
        \vspace{20pt}
    \end{subfigure}
    
    \caption{\textbf{Exciting and inhibiting copy behavior in upper bottom block.} We scale the parameters suppressing copy as identified by ExPLAIND in the upper bottom block in Phase II and detailed in \cref{sec:methodology_explaind}. The corresponding parameter groups are the attention key and query projections in layer 9, the attention key and value projections in layer 8, and the MLP down projections of layers 6 and 7, i.e. a total of six parameter groups accounting for the TOP-6 copy-suppressing influence in Phase II. To investigate their effect on the copy behavior of the model, we perform a naive ablation by doubling and halving their parameters in every 10th checkpoint.}
    \label{fig:copy_ablations_supp}
\end{figure*}

We plot the results of the copy ablations in  \cref{fig:copy_ablations_prom} and \cref{fig:copy_ablations_supp}. The parameters for this ablation are chosen as detailed in \cref{sec:methodology_explaind} and \cref{sec:copy_ablation_additional_details}.

\begin{figure*}
    \centering
    \begin{subfigure}[t]{0.375\textwidth}
    \includegraphics[trim=0 0 250 73,clip,width=1.0\linewidth]{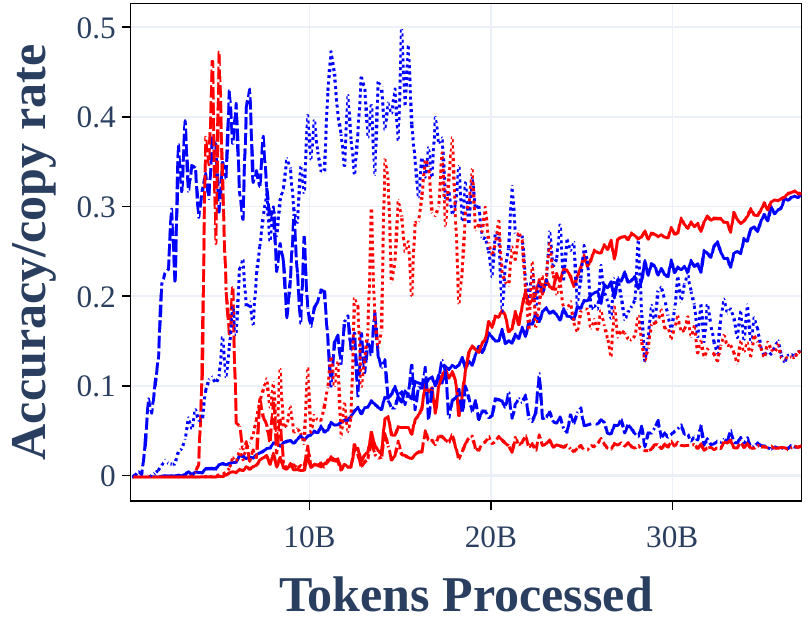}
    \caption{Bottom layers (input embeddings until layer~$9$).}
    \end{subfigure}
    \begin{subfigure}[t]{0.615\textwidth}
    \includegraphics[trim=0 0 0 73,clip,width=1.0\linewidth]{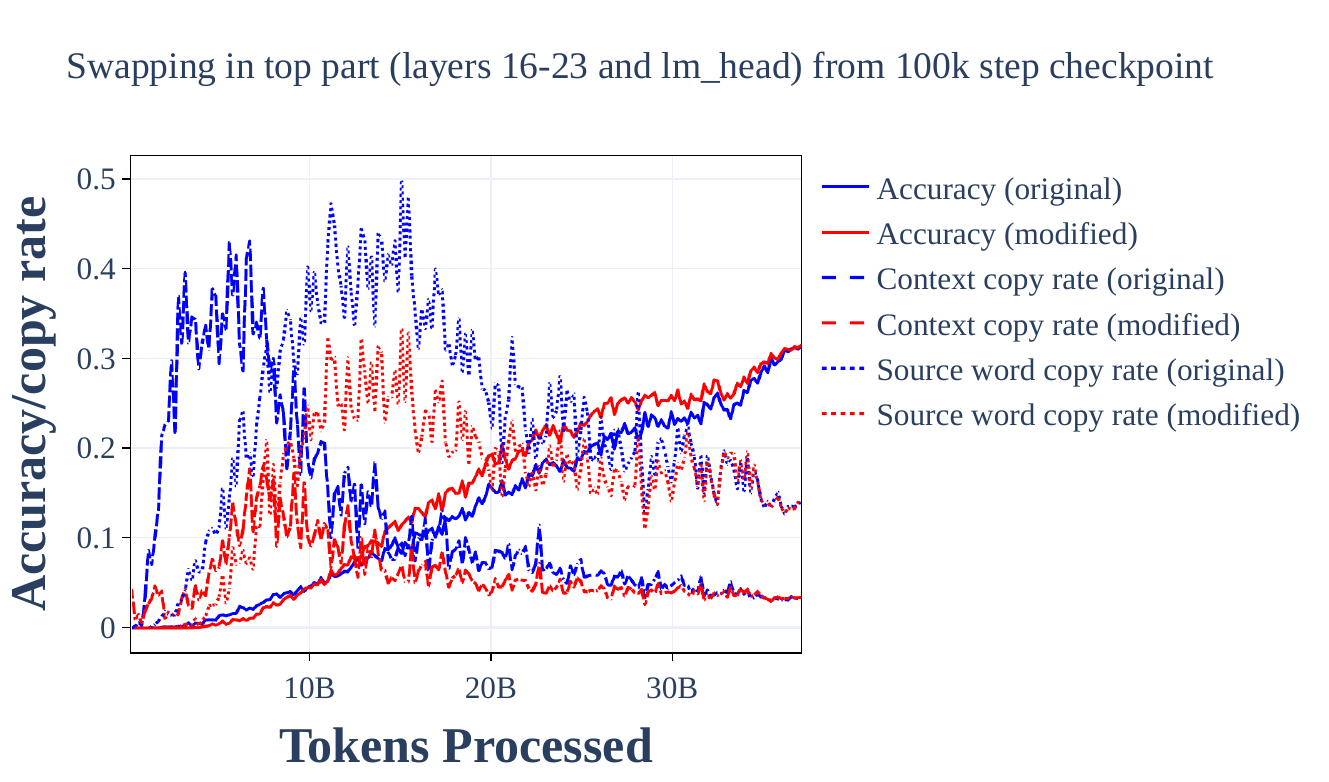}
     \caption{Top layers (layer $16$ until output embeddings).}
     \vspace{20pt}
    \end{subfigure}

    \begin{subfigure}[t]{0.375\textwidth}
    \includegraphics[trim=0 0 250 73,clip,width=1.0\linewidth]{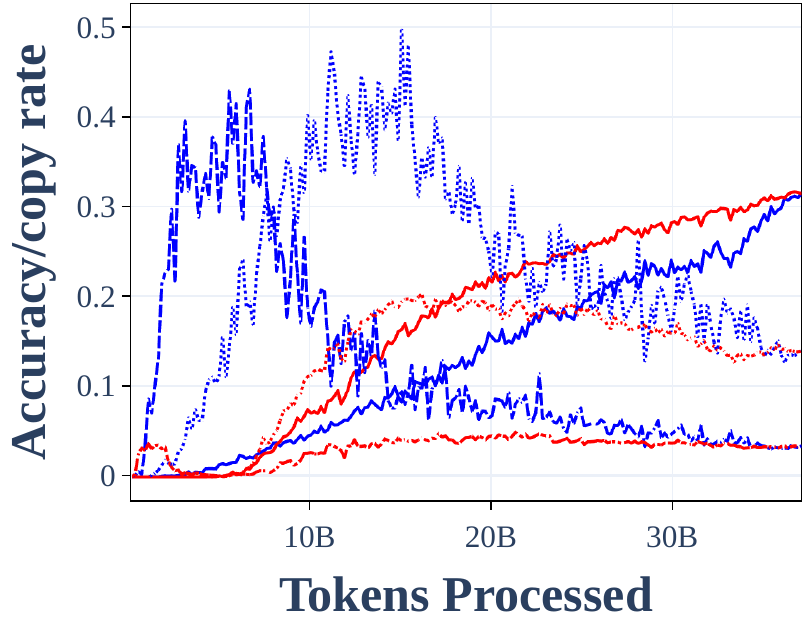}
    \caption{Bottom and intermediate layers (input embeddings until layer $15$).}
    \end{subfigure}
    \begin{subfigure}[t]{0.615\textwidth}
    \includegraphics[trim=0 0 0 73,clip,width=1.0\linewidth]{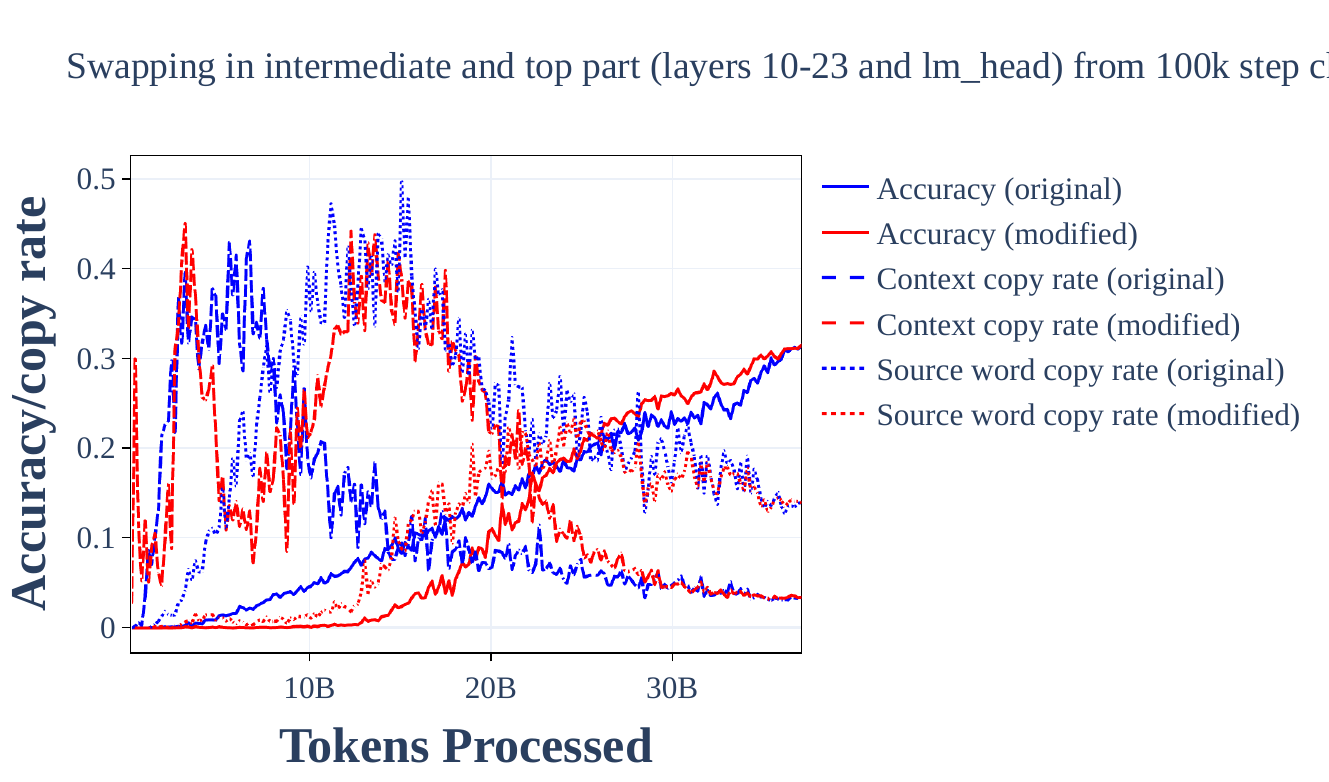}
     \caption{Intermediate and top layers (layer $10$ until output embeddings).}
     \vspace{20pt}
    \end{subfigure}

    \begin{subfigure}[t]{0.375\textwidth}
    \includegraphics[trim=0 0 250 73,clip,width=1.0\linewidth]{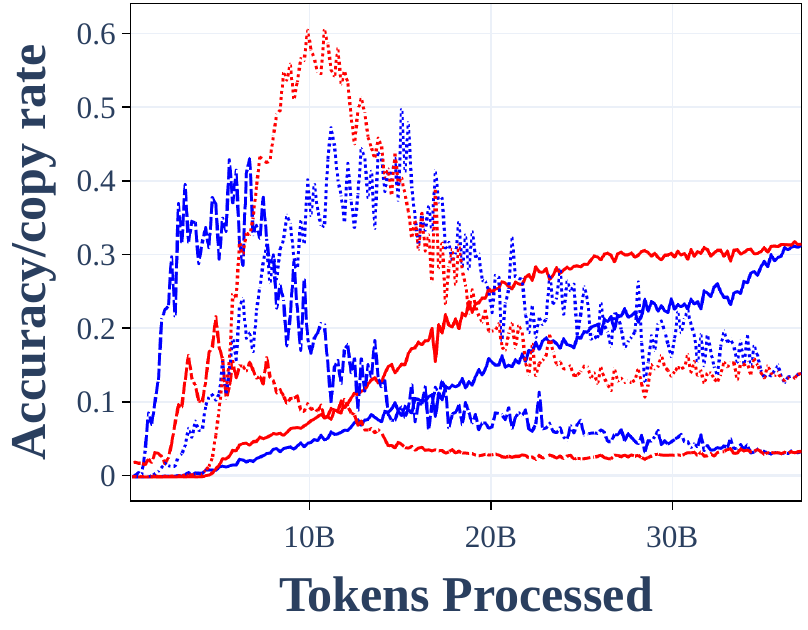}
    \caption{Bottom and top layers.}
    \end{subfigure}
    \begin{subfigure}[t]{0.615\textwidth}
    \includegraphics[trim=0 0 0 73,clip,width=1.0\linewidth]{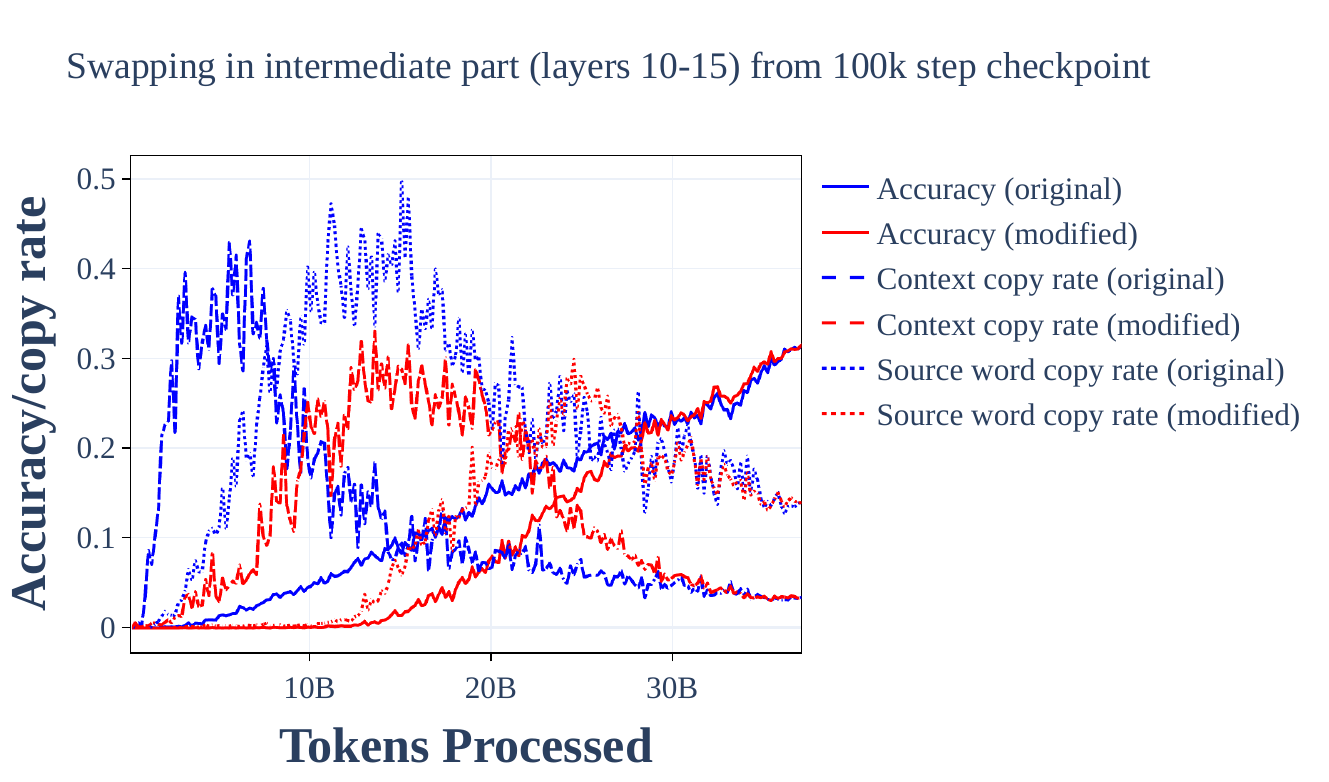}
     \caption{Intermediate layers (layers $10$--$15$).}
    \vspace{20pt}
    \end{subfigure}
    
    \caption{For each of the layer-swapping experiments shown in \cref{fig:appendix_swapping}, we also plot the copy behavior of the respective models over time.}
    \label{fig:layer_swapping_copy}
\end{figure*}

\begin{figure*}
    \centering
    \includegraphics[trim=0 0 0 73,clip,width=0.5\linewidth]{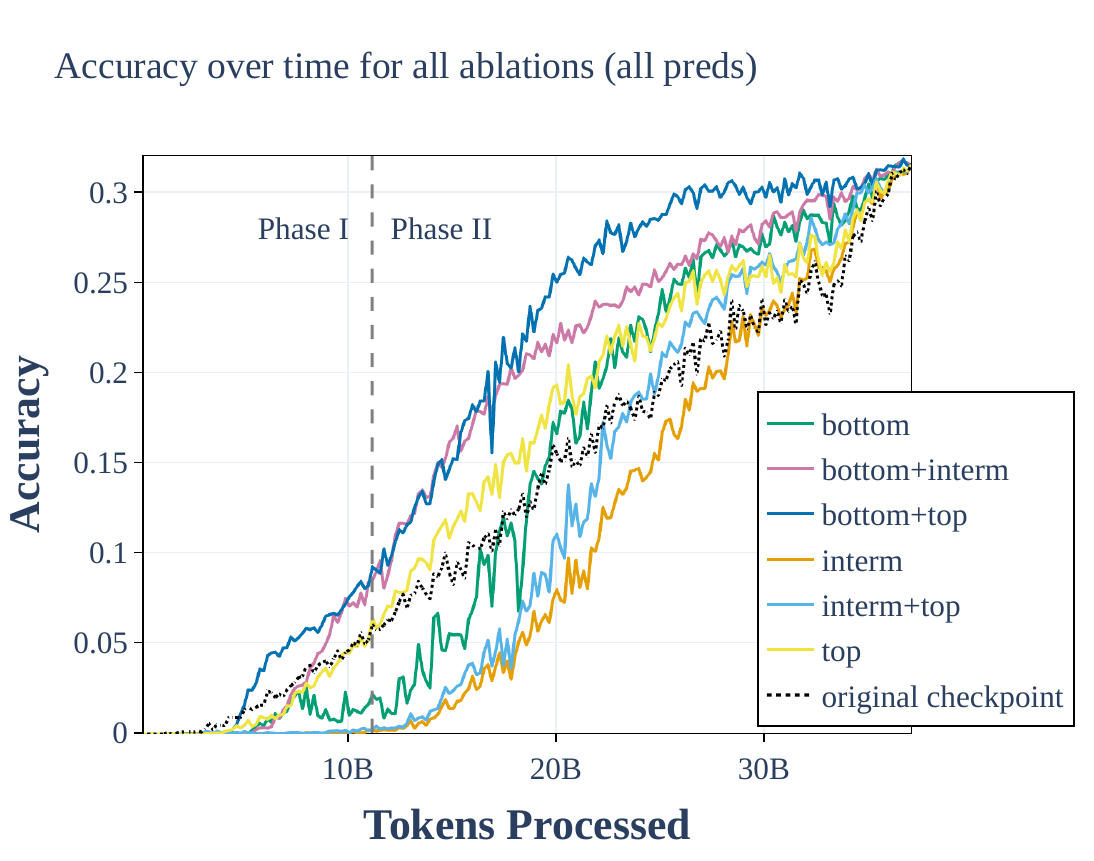}
    \caption{\textbf{Layer swapping over time.} We swap different layer blocks of the final checkpoint into earlier checkpoints and compare \wlt accuracy of the resulting model to the (unchanged) original checkpoint.  %
    }
    \label{fig:appendix_swapping}
\end{figure*}

\begin{figure*}
    \centering
    \begin{subfigure}[t]{0.49\textwidth}
    \includegraphics[trim=0 0 0 73,clip,width=1.0\linewidth]{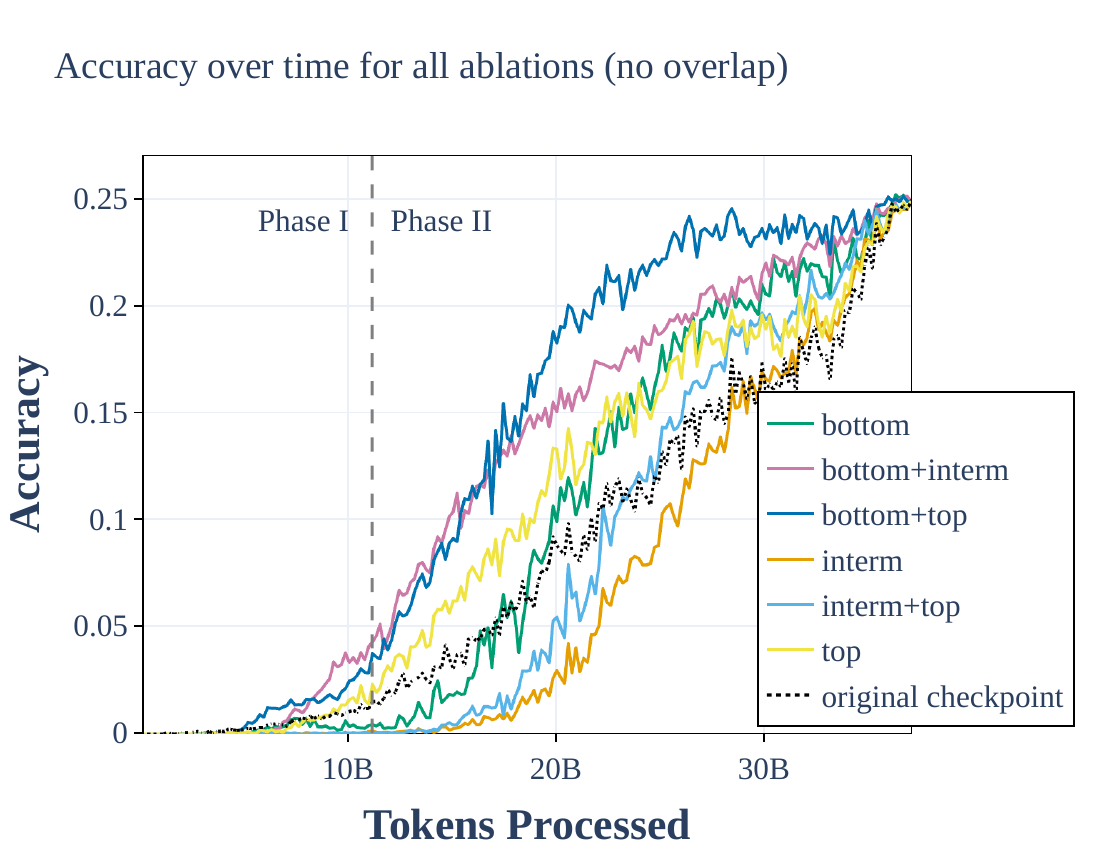}
    \caption{No token overlap.}
    \end{subfigure}
    \begin{subfigure}[t]{0.49\textwidth}
    \includegraphics[trim=0 0 0 73,clip,width=1.0\linewidth]{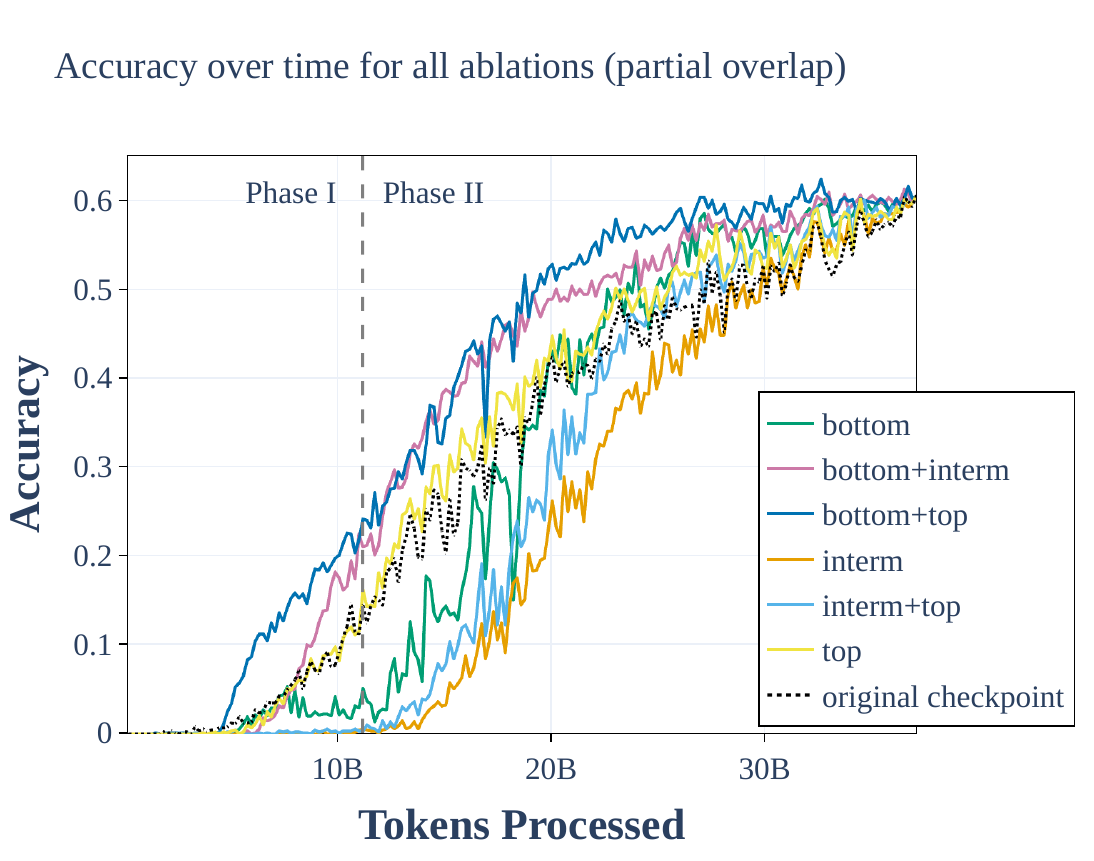}
     \caption{Partial token overlap.}
    \vspace{20pt}
    \end{subfigure}
    \\
    \begin{subfigure}[b]{0.49\textwidth}
    \includegraphics[trim=0 0 0 73,clip,width=1.0\linewidth]{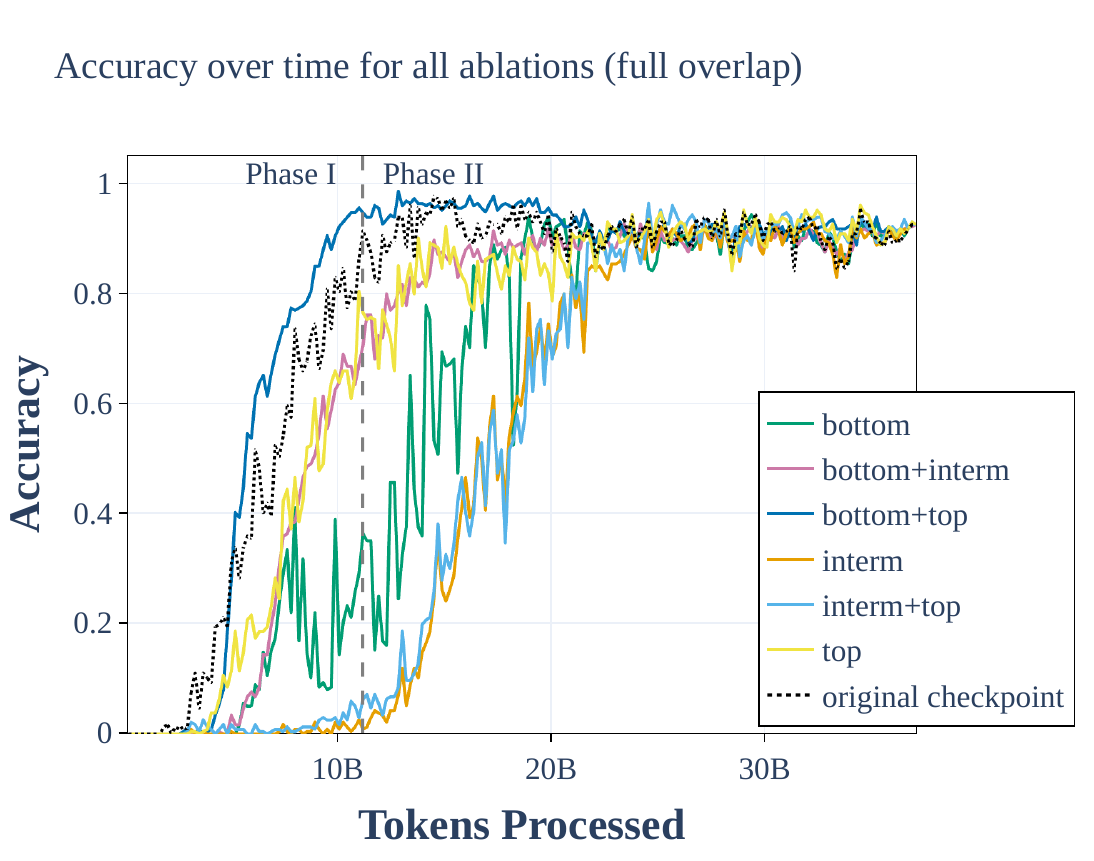}
     \caption{Full token overlap (predicted by copy only).}
    \vspace{20pt}
    \end{subfigure}
    \caption{\textbf{Layer swapping over time.} We swap different layer blocks of the final checkpoint into earlier checkpoints and compare \wlt accuracy of the resulting model to the (unchanged) original checkpoint. Here, we further group the predictions by the level of token overlap between the source word and any of the valid predictions. No overlap refers to predictions where none of the tokens of the correct prediction could be copied from the source word, full overlap refers to predictions which can be predicted by only copying tokens from the source word. In accordance, partial overlap refers to predictions where some, but not all tokens can be copied to translate correctly.
    }
    \label{fig:appendix_overlap_swapping}
\end{figure*}

\end{document}